%% file: main.tex
\documentclass[12pt,reqno]{amsart}

\usepackage[left=1in,right=1in,top=1in,bottom=1in]{geometry}
\usepackage{amsmath, amssymb, graphicx, url, mathtools}
\usepackage[round]{natbib}
\usepackage{xcolor}
\usepackage{wrapfig}
\usepackage{tikz}
\usepackage{mathptmx}

\usepackage{tikz}
\usepackage{multirow}
\usepackage{mathtools} 
\usepackage{wrapfig}
\usepackage{bm}
\usepackage{paralist}
\usepackage{accents}
\usepackage{enumerate}
\usepackage{caption}

\usepackage{mathptmx}

\usepackage{amsmath,amssymb,amsthm}
\usepackage{url,mathtools}
\usepackage{colortbl,framed}
\usepackage{enumitem}

\usepackage{float}
\usepackage[round]{natbib}
\usepackage{placeins}

\renewcommand{\qed}{\hfill{\tiny \ensuremath{\blacksquare} }}%

\newcommand{\BLP}{\mathsf{BLP}}

\newcommand{\ci}{\perp\!\!\!\perp}
\renewcommand{\qed}{\hfill {\tiny {\ensuremath{\blacksquare}}}}

\newtheorem{theorem}{Theorem}[section]
\newtheorem{algorithm}{Algorithm}[section]

\newtheorem{lemma}{Lemma}[section]

\newtheorem{definition}{Definition}[section]

\theoremstyle{definition}

\newtheorem{remark}{Comment}[section]
\numberwithin{remark}{section}
\newtheorem{example}{Example}
\numberwithin{equation}{section}
\numberwithin{theorem}{section}

\usepackage{url}

\usepackage{pagecolor,lipsum}

\newcommand{\Ep}{{\mathrm{E}}}

\renewcommand{\Pr}{{\mathrm{P}}}

\renewcommand{\hat}{\widehat}

\renewcommand{\Pr}{{\mathrm{P}}}

\renewcommand{\hat}{\widehat}
\renewcommand{\leq}{\leqslant}
\renewcommand{\geq}{\geqslant}

\DeclareMathOperator{\Var}{Var}

\usepackage{wallpaper}
\usepackage{graphicx}

\usepackage{xr}
\makeatletter

\newcommand*{\addFileDependency}[1]{
\typeout{(#1)}
\@addtofilelist{#1}
\IfFileExists{#1}{}{\typeout{No file #1.}}
}\makeatother

\newcommand*{\myexternaldocument}[1]{%
\externaldocument{#1}%
\addFileDependency{#1.tex}%
\addFileDependency{#1.aux}%
}
\myexternaldocument{Online_Appendix}

\begin{document}

\begin{center}
\bigskip

\end{center}

\title[Generic ML for Features of Heterogenous Treatment Effects]{Fisher-Schultz Lecture: Generic Machine Learning  Inference on Heterogenous Treatment Effects in Randomized Experiments, with an Application to 
Immunization in India}
\author{Victor  Chernozhukov \and  Mert Demirer \and Esther Duflo  \and Iv\'an Fern\'andez-Val}\thanks{The authors respectively from MIT, MIT, MIT and BU.  This paper was delivered (virtually) by Esther Duflo at the Fisher-Shultz Lecture of the Econometric Society World Congress, 2020. We thank three anonymous referees, Susan Athey, Moshe Buchinsky, Denis Chetverikov, Carlos Cineli, Matt Hong, Stella Hong, Steven Lehrer, Siyi Luo, Max Kasy, Sylvia Klosin, Susan Murphy, Whitney Newey, Patrick Power, Victor Quintas-Martinez, Suhas Vijaykumar, and seminar participants at ASSA 2018, Barcelona GSE Summer Forum 2019, Brazilian Econometric Society Meeting 2019, BU, Lancaster, NBER summer institute 2018, NYU, UCLA, Whitney Newey's Contributions to Econometrics conference, and York for valuable comments. Anirudh Sankar provided us with excellent research assistance. We gratefully acknowledge research support from the National Science Foundation, AFD, USAID, and 3ie. An R package that implements the methods in this paper, GenericML, is available on GitHub at \url{https://github.com/mwelz/GenericML}. }
\date{\today; Initial ArXiv Submission, December 2017.}
\maketitle

\begin{abstract}
We propose strategies to estimate and make inference on key features of heterogeneous effects in randomized experiments. These key features include \textit{best linear predictors  of the effects} using machine learning proxies, \textit{average effects sorted by impact groups}, and \textit{average characteristics of most and least impacted units}. The approach is valid in high dimensional settings, where the effects are proxied (but not necessarily consistently estimated) by predictive and causal machine learning methods.  We post-process these proxies into estimates of the key features.  Our approach is generic, it can be used in conjunction with penalized methods, neural networks,  random forests, boosted trees, and ensemble methods, both predictive and causal. Estimation and inference are based on repeated data splitting to avoid overfitting and achieve validity.  We use quantile aggregation of the results across many potential splits, in particular taking medians of  p-values and medians and other quantiles of confidence intervals. We show that quantile aggregation lowers estimation risks over a single split procedure, and establish its principal inferential properties. Finally, our analysis reveals ways to build  provably better machine learning proxies through causal learning: we can use the objective functions that we develop to construct the best linear predictors of the effects, to obtain better machine learning proxies in the initial step.  We  illustrate the use of both inferential tools and causal learners with a randomized field experiment that evaluates a combination of nudges to stimulate demand for immunization in India. \\



\end{abstract}

\input{Introduction}

\input{Section2}

\input{Section3}

\input{Section4}

\input{Section5}

\input{Section6}

\input{ConclusionExtensions}

\bibliography{mybibVOLUME.bib}

\appendix

\input{Appendix-Proofs.tex}

\input{Appendix-AdditionalTheory.tex}

\input{AdditionalEmpirical.tex}

\end{document}

%% file: Introduction.tex
\section{Introduction}

Randomized  Controlled Trials (RCT) and Machine Learning (ML) are arguably two of the most important developments in data analysis methods for applied researchers. 
RCTs play an important role in the evaluation of social and economic programs, medical treatments and marketing  \citep[e.g.,][]{duflo2007using,imbens2015causal}. 
ML  is a name  attached to a variety of constantly evolving statistical learning methods including Random Forest, Boosted Trees, Neural Networks, Penalized Regression, Ensembles, and Hybrids; see, e.g., \cite{wasserman:ML}, \cite{ESL}, \cite{bishop2006pattern}, \cite{murphy2012machine}, \cite{hastie2015}, \cite{goodfellow2016deep} and \cite{james2021} for prominent textbook treatments. ML has become a key tool for prediction and pattern recognition problems, surpassing classical methods in high dimensional settings.

At first blush, those two sets of methods may seem to have very different applications: in the most basic randomized controlled experiment, there is a sample with a single treatment and a single outcome. Covariates are not necessary and even linear regression is not the best way to analyze the data \citep{freeman2008,imbens2015causal}. In practice, however, applied researchers are often confronted with more complex experiments. For example, there might be accidental imbalances in the sample, which require selecting control variables in a principled way. ML tools, such as the lasso method proposed in \cite{BelloniChernozhukovHansen2011,bcfh17}  or the double machine learning method proposed in \cite{DML}, have proven useful for this purpose. Moreover, some complex RCT designs have so many treatment combinations that ML methods may be useful to select the few treatments that actually work and pool the rest with the control groups for statistical power \citep{banerjee2019improving}. Finally, researchers and policy makers are often interested in features of the impact of the treatment that go beyond the simple average treatment effect. In particular, very often, they want to know whether the treatment effect depends on  covariates, such as gender, age, etc. This heterogeneity is essential  to assess if the impact of the program would generalize to a  population with different characteristics, and, for economists, to better understand the driving mechanism behind the  effects of a particular program. In a review of 189 RCTs published in top economic journals since 2006, we found that 76 (40\%) report at least one subgroup analysis, wherein they report treatment effects in subgroups formed by baseline covariates.\footnote{The papers were published in \emph{Quarterly Journal of of Economics}, \emph{American Economic Review}, \emph{Review of Economics Studies}, \emph{Econometrica} and \emph{Journal of Political Economy}. We thank Karthik Mularidharan, Mauricio Romero and Kaspar W\"{u}thrich for sharing the list of papers they computed for another project.}
 
One issue with reporting treatment effects split by subgroups, however, is that there might be a large number of potential ways to form subgroups. Often researchers collect rich baseline surveys, which give them access to a large number of covariates: choosing subgroups ex-post opens the possibility of overfitting. To solve this problem, medical journals and the FDA require pre-registering the sub-sample of interest in medical trials \emph{in advance}. In economics, this approach has gained some traction with the adoption of pre-analysis plans, which can be filed in the AEA registry for randomized experiments. However, restricting the heterogeneity analysis to pre-registered subgroups amounts to throwing away a large amount of potentially valuable information, especially now that many researchers collect large baseline data sets. It should be possible to use the data to discover \emph{ex post} whether there is any relevant heterogeneity in treatment effect by covariates. 

To do this in a disciplined fashion and avoid the risk of overfitting, scholars have recently proposed using ML tools.
Indeed, ML tools seem ideal for exploring heterogeneity of treatment effects when researchers have access to a potentially large array of baseline variables to form subgroups and few guiding principles on which of those are likely to be relevant. Several recent papers, which we review below, develop methods for detecting heterogeneity in treatment effects. Empirical researchers have taken notice.\footnote{In the recent past, several new empirical papers in economics used ML methods to estimate heterogeneous effects. E.g. \cite{Rigol} showed that villagers outperform the machine learning tools when they predict heterogeneity in returns to capital. \cite{Heller} predicted who benefits the most from summer internship projects. \cite{Deryugina} used the methods developed in the present paper to evaluate the heterogeneity in the effect of air pollution on mortality. \cite{Creponetal} also built on the present paper to develop a methodology to determine if the impact of two different programs can be accounted for by different selection. The methodological papers reviewed later also contain a number of empirical applications.}
 
This paper develops a generic approach to using any of the available ML tools to predict and make inference on heterogeneous treatment or policy effects. A core difficulty of applying ML tools to the estimation of heterogenous causal effects is that, while they are successful in prediction empirically, it is much more difficult to obtain uniformly valid inference, i.e., inference that remains valid under a large class of data generating processes. In fact, in high dimensional settings, absent strong assumptions, generic ML tools may not even produce consistent estimators of the \emph{conditional average treatment effect} (CATE), the difference in the expected potential outcomes between treated and control states conditional on covariates. Previous attempts to solve this problem focused either on specific tools (for example, the method proposed by \cite{athey:trees}, which has become popular with applied researchers, and uses trees), or on situations where those assumptions might be satisfied. Our approach to resolving the fundamental impossibilities in non-parametric inference is different. Motivated by  \cite{wasserman:adaptive}, instead of attempting to get consistent estimation and uniformly valid inference on the CATE itself, we focus on providing valid estimation and inference on \emph{features} of CATE. 

We start by building a ML proxy predictor of CATE, and then target features of the CATE based on this proxy predictor.  In particular, we consider  three objects, which are likely to be of interest to applied researchers and policy makers:  
(1) \textit{Best Linear Predictor} (BLP) of the CATE on the ML proxy predictor; (2)  \textit{Sorted Group Average Treatment Effects} (GATES) or average treatment effect by heterogeneity groups induced by the ML proxy predictor; and  (3) \textit{Classification Analysis}  (CLAN) or the average characteristics of the most and least affected units defined in terms of the ML proxy predictor. Thus, we can find out if there is detectable heterogeneity in the treatment effect based on observables, and if there is any, what the treatment effect is for different bins. And finally we can describe which of the covariates are associated with this heterogeneity. 
  
There is a trade-off between more restrictive assumptions or tools and a more ambitious estimation. We address this trade-off by focusing on coarser objects of the function rather than the function itself, but make as little assumptions as possible. This seems to be a worthwhile sacrifice: the objects for which we have developed inference appear to us at this point to be the most relevant, but in the future, one could easily use the same approach to develop methods to estimate other objects of interest. For example, \cite{Creponetal} used the same technique to construct and estimate a specific form of heterogeneity at the post-processing stage. Even then, as we will see, getting robust and conservative standard errors for heterogeneity requires a larger sample size than just estimating average treatment effects. This reflects a different
trade-off: if we do not assume that we can predict \emph{ex ante} where the heterogeneity might be (in which case we can write it down in a pre-analysis plan), power will be lower, and 
detecting heterogeneity will require a larger sample. This is a consideration that applied researchers will need to keep in mind when designing and powering their experiments, and when writing pre-analysis plans: if a particular dimension of heterogeneity is deemed important, it should be pre-specified. 

Another trade-off between our generic approach and the alternative approaches in the literature is that the parameters of interest are estimated conditional on a particular split of the data, since our estimation and inference methods rely on sample splitting to avoid overfitting and other inferential non-regularities.  Conditional on a single data split into a training and a hold-out sample, statistical inference is conceptually straightforward and appealing.  Indeed,  in this case, statistical inference reduces to the classical inference for linear regression and sample means.  Theoretically, if a researcher can credibly pre-commit to a single data split, this gives one clean solution to the inferential problem.  However, this introduces additional variability in the results, stemming from the random draw of the particular split of the data. Therefore, researchers often consider multiple sample splits, and aggregate the results. This reduces the probability that two researchers working with the same data will arrive at different conclusions. To formalize and support this approach, we propose quantile-aggregated inference -- which aggregates inferential results by taking medians of estimates and medians and other quantiles of upper and lower confidence intervals obtained from different splits.  We show that quantile aggregation formally lowers estimation (reporting) risks over a single-split procedure, and we establish its inferential properties.

The proposed approach is generic in that it can be applied in conjunction with any ML method. To compare and select among ML methods, we develop goodness-of-fit measures for the BLP and GATES. We also take one step backward and use these goodness-of-fit measures to build ML proxies that better target the CATE  through causal learning.  We show that these causal machines produce provably better proxies of the CATE than generic (predictive) ML methods. Moreover, by designing the ML to target CATE directly,  the post-processing methods that we develop can focus on providing valid inference, rather than correcting biases.

We apply our method to a large-scale RCT of nudges to encourage immunization in the state of Haryana,  Northern India. This experiment, an important practical application in its own right,  is designed and discussed in \cite{banerjee2019improving}. Immunization is generally recognized as one of the most effective and cost-effective ways to prevent illness, disability, and diseases. Yet,  worldwide, close to 20 million children every year do not receive critical immunizations \citep*{unicef2019progress}. While early policy efforts have focused mainly on improving the infrastructure for immunization services, a more recent literature suggests that ``nudges'' (such as small incentives, leveraging the social network, SMS reminders, social signalling, etc.) may have large effects on the use of those services.\footnote{See, for example,  \cite{banerjee2010improving,bassani2013financial,wakadha2013feasibility,
johri2015strategies,oyo2016interventions,gibson2017mobile,karing2018social,domek2016sms,uddin2016use,regan2017randomized,alatas2019celebrities,banerjee2019leveraging}.
} This project was a collaboration with the government of Haryana, which was willing to experiment with a combination of nudges, with the goal of choosing the most effective policy and implement it at scale. It built a custom vaccination platform, and ran a large-scale experiment covering seven districts, 140 Primary health centers, 2,360 villages involved in the experiment (including 915 at risk for all the treatments), and 295,038 children in the resulting database. Immunization was very low at baseline: in every single village of the district, the fraction of children whose parents had reported they received the measles vaccine (the last in the sequence) was 39\%, and only 19.4\% had received the vaccine before the age of 15 months, whereas the full sequence is supposed to be completed in one year. The experiment was a village-level, cross randomized design of three main nudges: providing incentives, sending SMS reminders,  and seeding ambassadors. It included several variants for each policy: the level and schedule of the incentives, the number of people receiving reminders, and the mode of selection of the ambassadors, leading to a large number (75) of finely differentiated bundles.

 \cite{banerjee2019improving} developed a methodology to identify the most effective and cost-effective bundle of policies, based on an application of LASSO to a marginal effects specification that imposes some structure on the bundles, and in particular, the idea that policy variants (e.g. level of incentives, or level of coverage of SMS reminders) may be indistinguishable in practice. They found that the most cost-effective policy is to combine ``information hubs'' (people identified by others as good at diffusing information)  and SMS reminders. This is cheap and can be done everywhere. In fact, they showed that this policy is the only one among those tested that would actually save money to the government for each measles shot, while increasing immunization. 
But the most \emph{effective} policy, i.e., the policy that increases immunization the most, is the combination of incentives, immunization ambassadors, and SMS reminders, which is much more expensive. Yet, while this policy increases the cost per immunization, 
the effects are important:  the number of monthly measles shots (the last vaccine in the schedule, and thus a marker for full immunization) delivered increases by 3.26, corresponding to 44\% of the mean vaccination rate in the control group that got neither SMS nor increasing incentives and information hubs. The government was therefore interested in finding out where the program would be most effective, to implement it only in those places even at the higher cost per immunization.

The pre-analysis plan specified to look for heterogeneity by gender and by  ``Village-level baseline/national census variables,  including assets, beliefs, knowledge, and attitudes
towards immunization'' but did not identify one or two specific baseline variables to look at. This reflected genuine uncertainty (as is often the case). Many factors can influence policy impact, from attitudes to implementation capabilities to baseline levels, and we did not have a specific theory of where to look. It is precisely the type of context that requires a principled approach to avoid overfitting, and provide a policy-relevant recommendation.\footnote{This approach of finding the best treatment and then looking at where it works the best gets closer to the idea of ``personalized medicine''. Using the same data, \cite{agarwal2020synthetic}  go one step further and use a ``synthetic intervention'' approach to look for the policy that works the best for each kind of village. }

The rest of the paper is organized as follows. Section \ref{sec:model}  formalizes the framework, describes our approach and compares it with the existing literature. Section \ref{sec:id} presents identification and estimation strategies for the key features of CATE of interest. Section \ref{sec:inference} introduces our inference method that accounts for uncertainty coming from parameter estimation and sample splitting. 
Section \ref{sec:further} presents the construction of causal machines that can learn CATE better than purely predictive approaches or some existing proposals for causal approaches.
Section \ref{sec:empirics} reports the results of the empirical application and provides detailed implementation algorithms. Section  \ref{sec:conclusion} concludes with some remarks. The Online Appendix (OA) gathers proofs of the main theoretical results
and additional technical results.

%% file: Section2.tex
\section{Our Agnostic Approach}\label{sec:model}
This section present our framework and approach. We observe $\mathrm{Data} := (Y_i, Z_i, D_i)_{i=1}^N$, consisting of i.i.d. copies of the random vector $(Y,Z,D)$ having probability law $P$, where $Y$ is the outcome of interest, $D$ is a binary treatment indicator, and $Z$ is a possibly high-dimensional vector of covariates that characterize the observational units. The data is defined on an underlying probability space with measure $\Pr$. The expectation operator is denoted by $\Ep$. When we need to emphasize the dependence of $\Pr$ and $\Ep$ on $P$, we use the notation $\Pr_P$ and $\Ep_P$.

\subsection{Model and Key Causal Functions.} Let $Y(1)$ and $Y(0)$ be the potential outcomes
in the treatment state 1 and the non-treatment state 0 \citep[see ][]{neyman1923applications,rubin74}. 
The main causal functions
are the baseline conditional average (BCA):
\begin{equation}\label{BCA}
b_0(Z):= \Ep [Y(0) \mid Z],
\end{equation}
and the conditional average treatment effect (CATE):
\begin{equation}\label{CATE}
s_0(Z): = \Ep [Y(1) - Y(0) \mid Z] = \Ep [Y(1) \mid Z] -  \Ep [Y(0) \mid Z].
\end{equation}

Suppose $D$ is randomly assigned conditional on $Z$,
with probability of assignment depending on a subvector of stratifying variables $Z_1 \subseteq Z$, namely
\begin{equation}\label{indep}D \ci (Y(1), Y(0)) \mid Z,
\end{equation} and the propensity score is known and is given by
\begin{equation}
p(Z) := \Pr[D =1 \mid Z] = \Pr[D=1 \mid Z_1],
 \end{equation}
which we assume  is bounded away
from zero or one:
\begin{equation}
 p(Z)  \in [p_0, p_1] \subset (0,1) \ \  a.s.
 \end{equation}
This setup is similar to \cite{rosenbaum1983}.
 
 The observed outcome is 
$Y = DY(1) + (1-D) Y(0)$.    Under the stated assumption, the causal functions
are identified by the components of the regression function 
of $Y$ given $D, Z$:
 \begin{equation}\label{eq: HetPL1}
 Y  = b_0 (Z) +  D s_0(Z) + U,   \ \  \Ep[U\mid Z, D]= 0,
\end{equation}
that is, $\ b_0(Z) = \Ep [Y \mid D=0, Z]$, and 
 \begin{equation}\label{CATEid}
s_0(Z) = \Ep [Y \mid  D=1, Z] -  \Ep [Y \mid  D=0, Z].
\end{equation}
This regression underlies the use of predictive ML methods that learn $\Ep[Y \mid D, Z]$ and then estimate CATE using the formula.

Alternatively one can identify CATE using the following two equivalent 
``causal" regressions:
\begin{equation}\label{CATEid2}
s_0(Z)= \Ep[ HY \mid Z] =  \mathrm{Cov}[ H, Y \mid Z];
\end{equation}
where $H$ is the residualized treatment scaled by its variance: 
\begin{equation}\label{eq:Htransform}
H = H(D,Z) := \frac{D-p(Z)}{p(Z)(1-p(Z))},
\end{equation}
also known as the Horvitz-Thompson transform.  We mention these alternative strategies here, because as shown in Section 5,  they can lead to better ways of approximating $s_0(Z)$ than through the predictive regression (\ref{eq: HetPL1}), and our inference tools equally apply to ML methods that try to learn $s_0(Z)$ through either of these relations. In fact, in our empirical analysis, the strategies based on \eqref{CATEid2} measurably outperform the strategies based on \eqref{CATEid}.

\subsection{Estimation and Inference Challenges}
Regardless of the way we try to learn $s_0(Z)$,  estimation and inference are challenging in modern high-dimensional settings, because the target function $z \mapsto s_0(z)$ can live in a very complex class. ML methods effectively explore various forms of sparsity to yield ``good" approximations to $s_0(z)$.  In its simplest form, sparsity reduces the complexity of  $z \mapsto s_0(z)$ by assuming that it can be well-approximated by a function that only depends on a low-dimensional subset of $z$,  making consistent estimation possible.    As a result, these methods can perform much better than classical methods in high-dimensional settings under sparsity. However, sparsity or, more generally, low complexity of the CATE function $s_0$, are untestable assumptions that must be used with caution.
 


Without some form of sparsity,  it is hard, if not impossible, to obtain consistent estimators
of $z \mapsto s_0(z)$.  There are several fundamental reasons as well as large gaps between theory and practice that are responsible for this. One fundamental reason is that ML methods might not even produce consistent estimators of $z \mapsto s_0(z)$ in high dimensional settings. For example, if  $z$ has dimension $d$ and the target function $z \mapsto s_0(z)$ is assumed to have $p$ continuous and bounded derivatives, then the worst case (minimax) lower bound on the rate of learning this function from a random sample of size $N$ cannot be better than $N^{-p/(2p + d)}$  as $N \to \infty$, as shown by \cite{stone82}.  Hence if  $p$ is fixed and $d$ is also small, but slowly increasing with $N$, such as $d \geq \log N$, then there exists no consistent estimator of  $z \mapsto s_0(z)$ generally. Hence, generic ML estimators cannot be regarded as consistent, unless further assumptions are made.
Examples of such assumptions include structured forms of linear and non-linear sparsity and super-smoothness.\footnote{The function $z \mapsto s_0(z)$ is super-smooth if it has continuous and bounded derivatives of all orders.}  The problem of obtaining uniformly valid inference on $z \mapsto s_0(z)$ using generic ML methods is even more difficult.\footnote{While the previous assumptions make consistent adaptive estimation possible \citep[e.g.,][]{BickelRitovTsybakov2009}, confidence sets that adapt to unknown regularity (smoothness or sparsity) do not exist 
even for low-dimensional nonparametric problems \citep{low1997,wasserman:adaptive}.  Let $z \mapsto s_0(z)$ be a target  function  that lives in an infinite-dimensional class with unknown regularity $s$ (e.g., smoothness or degree of sparsity). Adaptive consistent estimation (resp. inference)  for  $z \mapsto s_0(z)$ with respect to $s$ is possible if there exists a consistent estimator (resp. valid confidence set) with a rate of convergence (resp. diameter) that changes with $s$  in a (nearly) rate-optimal way. Construction of adaptive confidence bands then requires making additional untestable  assumptions. See, e.g., \cite{gine:nickl}, where self-similarity conditions are used in low-dimensional nonparametric problems.}

In this paper, we take an agnostic view. We neither rely on any sparsity or low-complexity assumptions to make the ML estimators consistent, nor impose other stronger conditions to make ``traditional" confidence intervals valid. We simply treat ML as providing proxy predictors for the objects of interest.  


\subsection{Our Approach}

To address the previous challenges, we propose strategies for estimation and inference on \textit{ key features} of  $s_0(Z)$ rather than on $s_0(Z)$ itself.
Because of this difference in focus, and by relying on sample splitting,  we can loosen the restrictions about the properties of the ML estimators. 

Let $(M,A)$ denote a random partition of the set of indices $\{1, \ldots, N\}$. The strategies that we consider rely on random splitting of $\mathrm{Data} = (Y_i, D_i, Z_i)_{i=1}^N$  into a main sample, denoted by $\mathrm{Data}_M$ $=$ $(Y_i, D_i, Z_i)_{i \in M}$, and an auxiliary
sample, denoted by $\mathrm{Data}_A = (Y_i, D_i, Z_i)_{i \in A}$.  We will sometimes refer to these samples as $M$ and $A$.  
After splitting the sample, we carry out two stages:

\textbf{Stage 1}: From the auxiliary sample $A$,  we obtain ML estimators of  the baseline functions and treatment effects,
which we call the ML proxy predictors,
$$
z \mapsto B(z)  \text{ and }  z \mapsto S(z).
$$
Here $S(Z)$ is a possibly biased and noisy predictor of  $s_0(z)$ and $B(Z)$ is a possibly biased and noisy predictor of $b_0(Z)$ (or other technical ``baseline" functions, as we discuss in Section \ref{sec:further}). We do not require these predictors to be consistent for the true functions.

\textbf{Stage 2}: We post-process the proxies from Stage 1 to estimate and make inference on  features of the CATE function $z \mapsto s_0(z)$ in the main sample $M$.  
The  key features that we target include: 
\begin{itemize} 
\item[(1)] \textbf{Best Linear Predictor} (BLP) of the CATE $s_0(Z)$  on the ML proxy predictor $S(Z)$;
\medskip
\item[(2)] \textbf{Sorted Group Average Treatment Effects} (GATES):  average of  $s_0(Z)$ (ATE) by heterogeneity groups induced by the ML proxy predictor $S(Z)$;
\medskip
\item[(3)] \textbf{Classification Analysis}  (CLAN): average characteristics of the most and least affected units defined in terms of the ML proxy predictor $S(Z)$.
\end{itemize}
Our approach is \textit{generic} with respect to the ML method
being used, and is \textit{agnostic} about its formal properties. However, it relies on sample splitting, which introduces a specific source of uncertainty. To account for this, we use many data splits into main and auxiliary samples to produce robust estimators, and we employ quantile aggregation of inference to combine results across splits. Specifically, for point estimation, we report the median of the estimated key features over different random splits of the data. We take medians and other quantiles of many random conditional confidence sets for interval estimation. Finally, we construct p-values by taking medians of many random conditional p-values.  We establish the formal inferential properties of this procedure.



\subsection{Relationship to the Literature.} We focus the review strictly on the literatures about estimation and inference on heterogeneous effects and inference using sample splitting.  

This work is related to the literature that uses linear and semiparametric regression methods for estimation and inference on heterogeneous effects. \cite{crump2008nonparametric} developed tests of treatment effect homogeneity for low-dimensional settings based on traditional series estimators of the CATE. A semiparametric inference method for characterizing heterogeneity, called the sorted effects method, was given in \cite{CFL2014}. This  approach does provide a full set of inference tools, including simultaneous bands for percentiles of the CATE, but is strictly limited to the traditional semiparametric estimators of the regression and causal functions.  \cite{hansen:kobzur} proposed a sparsity-based method called ``targeted undersmoothing" to perform inference on heterogeneous effects.   This approach does allow for high-dimensional settings, but imposes sparsity as well as additional assumptions that enable the targeted undersmoothing.   A related approach, which allows for simultaneous inference on many coefficients (for example, inference on the coefficients corresponding to the interaction of the treatment with other variables) is proposed \cite{BCK-LAD} using a Z-estimation framework, where the number of interactions can be very large;  see also \cite{dezeure2016high} for a more recent effort in this direction, focusing on de-biased lasso in mean regression problems. This approach,  however, still relies on a strong form of sparsity assumptions.  \cite{zhao2017selective} proposed a post-selection inference framework within  high-dimensional linear sparse models for the heterogeneous effects. The approach is attractive because it allows for some misspecification of the model.

Another approach is to use tree-based and other methods.
 \cite{Imai:stuff} discussed the use of a heuristic support-vector-machine method with lasso penalization for classification of heterogeneous treatments into positive and negative ones. They used the Horvitz-Thompson transformation of the outcome  \citep[e.g., as in][]{hirano:imbens:ridder,abadieDD} such that the new outcome becomes an unbiased, noisy version of CATE.\footnote{Note that using Horvitz-Thompson (HT) transform of outcome, in this and other references, typically gives very noisy signal. One can improve the approach by either including the HT transform interacted with some baseline covariates as regressors in a regression model, as we do in the present paper, or using residualized outcomes in conjunction with HT, as in \cite{semenova2020}.} \cite{athey:trees} made use of the Horvitz-Thompson transformation of the outcome to inform the process of building causal trees, with the main goal of predicting CATE. They also provided a valid inference result  on average treatment effects for groups defined by the tree leaves, conditional on the data split into two subsamples: one used to build the tree leaves and the one to estimate the predicted values given the leaves. Like our methods, this approach is essentially assumption-free.  Our paper is a complement, in that our approach can be used with any ML method. \cite{wager:athey}  proposed a subsampling-based construction of a causal random forest, providing valid pointwise inference for CATE (see also the review in  \cite{wager:athey}  on prior uses of random forests in causal settings) for the case when covariates are very low-dimensional (and essentially uniformly distributed).\footnote{The dimension $d$ is fixed in \cite{wager:athey}; the analysis relies on the Stone's model with smoothness index $\beta=1$, in which no consistent estimator exists once $d \geq \log n$ in the minimax sense.} 
This condition rules out the typical high-dimensional settings that arise in many empirical problems, especially in current RCTs, where the number of baseline covariates is potentially very large. 
 


Several other studies look at model-based strategies for performing inference on CATE.
\cite{semenova2020}  used ML to perform inference on the ``partial" CATE, $\Ep[ s_0(Z) \mid X]$, where $X$ is a prespecified low-dimensional set of covariates. Specifically, they constructed an estimator of a denoised HT transform of the outcome and projected it using a nonparametric series estimator on the set of low-dimensional prespecified covariates of interest $X$, whose dimension is much lower than the dimension of $Z$.\footnote{Specifically, the denoised HT transform of outcome is $\tilde Y = g(1,Z) - g(0,Z) + H ( Y - g(D,Z) )$, where $H = (D-p(Z))/[p(Z)(1-p(Z)]$ and
$g(D,Z) = \Ep (Y \mid D, Z)$. \cite{semenova2020} used ML to estimate $g(D,Z)$ and the propensity score $p(Z)$, in case the latter is unknown.} The main advantage of this approach is that it delivers familiar nonparametric inference on partial CATE (even though inference on the full CATE remains intractable). 
\cite{fan2022estimation}, \cite{zimmert},  \cite{CNS:AutoLocal}, and \cite{CNS:simple} developed kernel versions of this procedure.  Related ideas but based on partialling-out (using residualized outcomes and treatment) appear in \cite{semenova:panel}, \cite{semenova:ROML},
\cite{nie:20}, \cite{foster2019orthogonal}, and \cite{kennedy2020optimal}.   Relative to the approach taken here, the assumptions made in these papers are more restrictive, but deliver stronger results. For example, in the approach of \cite{semenova2020}, one has to specify the baseline covariates $X$ for the partial CATE analysis, (which is exactly what we are trying to avoid in our approach). Second, the methods critically rely on the consistency of ML to estimate the nuisance components well, which can be restrictive  in high dimensional settings, as discussed above.


The idea of using a ``hold out'' sample to validate the result of a ML procedure to discover heterogeneity was suggested in \cite{Heller}, who used the method proposed in  \cite{wager:athey} and compared their results to the heterogeneity in a holdout sample. Our inference approach is different because it calls for multiple splits. This  procedure itself is also of independent interest and could be applied to many problems, where sample splitting is used to produce ML predictions \citep[e.g.,][]{abadie2013endogenous}.   
Related references include \cite{wasserman2009high}, and \cite{meinshausen2009p}, where the ideas are related, but the details are quite different, as we shall explain below. The premise is the same; however, as in \cite{meinshausen2009p} and \cite{rinaldo2016bootstrapping} -- we should not rely on a single random split of the data and should adjust inference in some way. 
Our construction of p-values builds upon ideas in  \cite{meinshausen2009p}, though what we propose is simpler, and our confidence intervals appear to be new.   Of course, sample splitting ideas are classical, going back to \cite{hartigan1969using,kish1974inference,barnard,cox1975note,mosteller:tukey}, though having been mostly underdeveloped and overlooked for inference, as characterized by \cite{rinaldo2016bootstrapping}. 
Finally, our inference method shares with the literature on post-selection inference in statistics that the target estimands are random functions depending on a ML proxy \citep[e.g.,][]{fithian2014optimal,lee2016exact}. 


%% file: Section3.tex
\section{Main Identification Results and Estimation Strategies}\label{sec:id}



In this section we condition on $\mathrm{Data}_A$ (the auxiliary sample in a random split) and therefore consider the functions $$
z \mapsto B_A(z) := B(z; \mathrm{Data}_A) \text{ and }  z \mapsto S_A(z) := S(z; \mathrm{Data}_A).
$$
as fixed functions. To lighten the notation, we keep the dependence on $\mathrm{Data}_A$ implicit in all the expectations, objects and parameters, that is, e.g., we use $\Ep[\cdot]$ and $S(Z)$ instead of $\Ep[\cdot \mid \mathrm{Data}_A]$ and $S_A(Z)$. We shall make the dependence on $\mathrm{Data}_A$ explicit when we discuss estimation and inference in Section \ref{sec:inference}.

\subsection{Best Linear Predictor of CATE} 
The first inferential target is the best linear predictor of the CATE using the proxy $S(Z)$.
\begin{definition}[BLP]\label{def:blp} The best linear predictor of $s_0(Z)$ by $S(Z)$ is the solution to:
$$
 \min_{b_1, b_2} \Ep [ s_0(Z) - b_1 - b_2 S(Z)]^2,
 $$
 which, if exists, is defined as
 $$
\BLP_{}[s_0(Z) \mid S(Z)] := {\beta_1} + {\beta_2} (S(Z)- \Ep_{} S(Z)),
$$
where $\beta_1 = \Ep_{} s_0(Z)$ and  $\beta_2 =  \mathrm{Cov}_{}[s_0(Z), S(Z)]/\Var_{}[S(Z)]$.
\end{definition}

By construction, $\BLP_{}[s_0(Z) \mid S(Z)]$ is an unbiased predictor of $s_0(Z)$, which improves over $S(Z)$ in the mean-squared error sense, that is
$$
\Ep \{ s_0(Z) - \BLP_{}[s_0(Z) \mid S(Z)]\}^2 \leq \Ep [ s_0(Z) - S(Z)]^2.
$$
Indeed, we can quantify the improvement by
$$
\Ep [ s_0(Z) - S(Z)]^2 - \Ep \{ s_0(Z) - \BLP_{}[s_0(Z) \mid S(Z)]\}^2 = (1-\beta_2)^2 \Var[S(Z)] + [\Ep S(Z) - \Ep s_0(Z)]^2,
$$
which is positive unless $S(Z)$ is an unbiased predictor and, either $\beta_2 = 1$ or  $\Var[S(Z)] = 0$.\footnote{The previous expression follows from the decompositions
$
\Ep [ s_0(Z) - S(Z)]^2 = \Ep [ \{s_0(Z) - \Ep s_0(Z)\} - \{S(Z) - \Ep S(Z)\} + \{\Ep s_0(Z) - \Ep S(Z)\}]^2
$
and
$\Ep \{ s_0(Z) - \BLP_{}[s_0(Z) \mid S(Z)]\}^2 = \Ep \{ [s_0(Z) - \Ep s_0(Z)] - [\BLP_{}[s_0(Z) \mid S(Z)] - \Ep s_0(Z)]\}^2,$
using that $\Ep \BLP_{}[s_0(Z) \mid S(Z)] = \Ep s_0(Z)$ and  $\beta_2 =  \mathrm{Cov}_{}(s_0(Z), S(Z))/\Var_{}(S(Z))$.
}
Accordingly, compared to the ML proxy, the BLP can be seen as a refined predictor of the individual CATE, $s_0(Z)$. If $S(Z)$ is a perfect proxy for $s_0(Z)$, then 
$\beta_2 =1.$ In general,  $\beta_2 \neq 1$,  correcting for 
noise in $S(Z)$. If $S(Z)$ is complete noise, uncorrelated to $s_0(Z)$, then  $\beta_2 =0$.
Furthermore, if there is no heterogeneity, that is, $s_0(Z) = s$, then
$
\beta_2 = 0.
$
Rejecting the hypothesis $\beta_2 = 0$ therefore means both that  there is heterogeneity in $s_0(Z)$ and 
$S(Z)$ is a relevant predictor. 

\begin{remark}[Conditioning on $\mathrm{Data}_A$]\label{remark:stability} As we mention above, all the objects and parameters are conditional on the auxiliary sample $\mathrm{Data}_A$. If we make the dependence explicit, $$
\BLP_{A}[s_0(Z) \mid S_A(Z)] = \beta_{A,1} + \beta_{A,2} (S_A(Z)- \Ep[S_A(Z) \mid \mathrm{Data}_A]).
$$
The $\BLP_A$ can be interpreted as the best linear predictor of $s_0(Z)$ given the proxy $S_A(Z)$ obtained from a random partition of the data. Therefore, $\BLP_A$ is random with respect to $A$, but converges to a fixed number as the size of the auxiliary sample grows under suitable stability conditions on $S_A(Z)$. This is discussed in Section \ref{sec:stability} of the SA. If, in addition, $S_A(z)$ is a consistent estimator of $s_0(z)$, then $\beta_{1,A} \to_P \Ep s_0(Z)$ and $\beta_{2,A} \to_P 1$ as the size of the auxiliary sample grows.
\end{remark}

We provide two strategies for identifying and estimating $\BLP_{}[s_0(Z) \mid S(Z)]$. 
 
\subsection*{Strategy A: Weighted Residual BLP}
Consider the weighted linear projection:
\begin{equation}\label{equation: wreg}
Y = \alpha_0'X_1 + \alpha_1 (D - p(Z))  +  \alpha_2 (D- p(Z)) (S - \Ep_{} S)    + \epsilon,  \ \  \Ep_{}[ w(Z) \epsilon 
X ] = 0, 
\end{equation}
where $S:=S(Z)$, $w(Z) := \{p(Z)(1-p(Z))\}^{-1},$ $X :=(X_1', X_2')',$ 
$$X_1 = [1, B(Z), p(Z), p(Z) S(Z)]', \quad  X_2 :=[D- p(Z),  (D- p(Z))(S - \Ep_{} S)  ]'.$$
The term $B(Z)$ could be replaced by any ``noise-reducing" proxy function. For example, the algorithms of \cite{semenova:panel} and \cite{nie:20}, targeting the delivery of $S(Z)$, also construct $B(Z)$ that are meant to approximate $\Ep[Y \mid Z]$ but not $b_0(Z)$; see also Section \ref{sec:further} for other examples of such algorithms.  We include $X_1$ to reduce finite sample noise of the estimators of the BLP parameter based on this strategy.\footnote{Note that $X_1$ can include other functions of $Z$. In our experiments, the use of  $B(Z)$  strongly improves the precision of estimating BLP (and other quantities such as GATEs and CLAN introduced below).}

Note that $\alpha_1$ and $\alpha_2$ are identified under weak assumptions. Further, we note that the interaction $(D- p(Z)) (S - \Ep_{} S)$ is {orthogonal}
to $D- p(Z)$ under the weight $w(Z)$, and to  all functions of $Z$ such as $X_1$. Consequently, we obtain the  following result that  shows that the linear projection \eqref{equation: wreg} identifies the  BLP.

\begin{theorem}[BLP Identification A]\label{theorem: BLP1}  Consider $z \mapsto S(z)$ and $z \mapsto B(z)$ as fixed maps.  Assume that $Y$ and $X$ have finite second moments, and $\Ep X X'$ is finite and full rank, which requires $\Var_{}(S(Z)) > 0$. Then, $(\alpha_1,\alpha_2)$ defined in \eqref{equation: wreg} identifies the coefficients of the BLP, 
$$\alpha_1 = \beta_1, \quad \alpha_2 = \beta_2.$$ 
\end{theorem}

\begin{remark}[Why not Classical OLS of $Y$ on Proxies?]\label{OLS:CATE} It is tempting  and perhaps more natural to consider the projection equation: 
\begin{equation*}
Y = \underbracket{\tilde \alpha_1 + \tilde \alpha_2 B + \tilde \beta_1 D +  \tilde \beta_2 D (S - \Ep_{} S)}_{\textrm{BLP of CEF}}    + \epsilon,  \quad \Ep[\epsilon \tilde X] = 0,
\end{equation*}
where $\tilde X = [1, B, D, D(S-\Ep S)]'$.  The idea here is the classical one: the ordinary least squares method with $Y$ as the outcome provides the Best Linear Predictor or Approximation to the CEF $\Ep[Y\mid D,Z]$, even if the latter is nonlinear. \cite{AngristBook} discuss the importance and practical relevance of this property. However, this property does not translate into providing the BLP of CATE $s_0(Z)$.
  Indeed, even in pure RCTs, while $\tilde \beta_1 = \beta_1$ is true, we have that $\tilde \beta_2  \neq \beta_2$ in general, and therefore
\begin{equation}\label{eq: Recover}
\tilde \beta_1 +  \tilde \beta_2 (S - \Ep S) \neq \BLP(s_0(Z) \mid S). 
\end{equation}
See Appendix \ref{app:sec3} 
of the OA for further discussion and proof.
\end{remark}

The identification result in Theorem \ref{theorem: BLP1} is constructive. We can base a corresponding estimation strategy
on the empirical analog:
\begin{equation}\label{eq:BLP estimate}
\begin{split}
Y_i &= \hat \alpha_0'X_{1i} + \hat \alpha_1 (D_i - p(Z_i))  + \hat  \alpha_2 (D_i- p(Z_i)) (S_i - \Bbb{E}_{N,M} S_i)    + \hat \epsilon_i, \ i \in M,  \ \\  &\Bbb{E}_{N, M}[ w(Z_i) \hat \epsilon_i 
X_i] = 0,
\end{split}
\end{equation}
where $X_i = [X_{1i}',X_{2i}']'$, $X_{1i} := [1, B(Z_i), p(Z_i), p(Z_i) S(Z_i)]'$, $X_{2i} :=[D_i- p(Z_i),  (D_i- p(Z_i))(S_i - \Bbb{E}_{N,M} S_i)  ]'$, and $\Bbb{E}_{N,M}$ denotes the empirical expectation with respect
to the main sample, i.e. $$\Bbb{E}_{N,M} h(Y_i, D_i, Z_i) := |M|^{-1}\sum_{i \in M} h(Y_i, D_i, Z_i) .$$ 

\begin{figure}
	\centering	
 		\caption{BLP Using ML Proxy vs the ML Proxy} 
   \vspace{-1cm}
 \includegraphics[width=.5\textwidth, height=.5\textwidth]{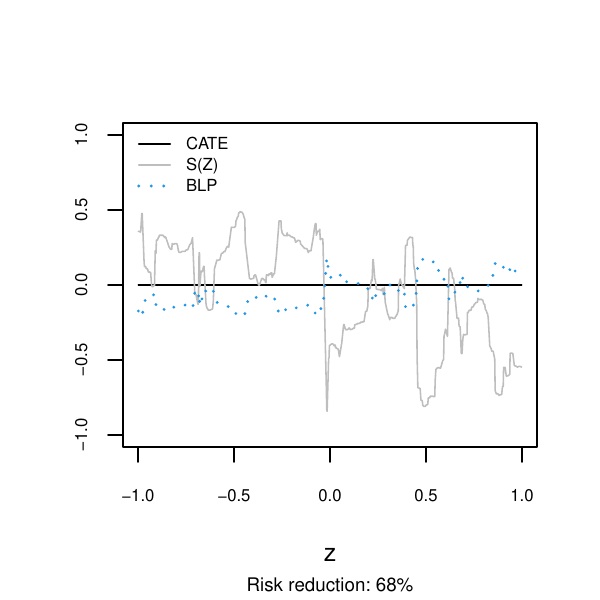}\includegraphics[width=.5\textwidth, height=.5\textwidth]{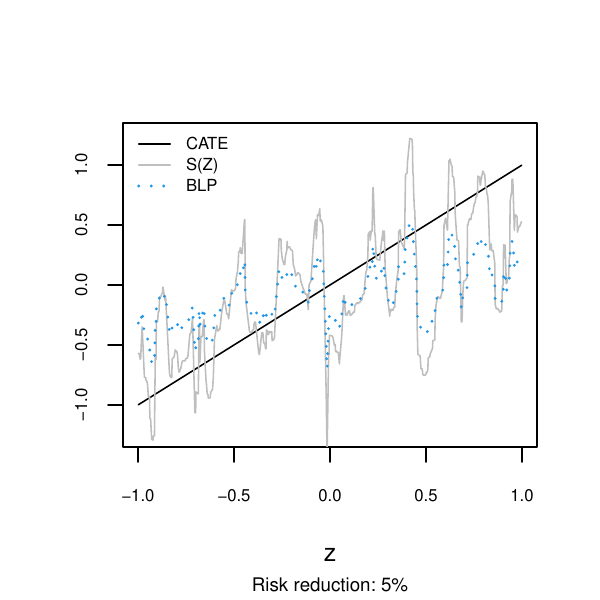}
 \caption*{\footnotesize\textsc{Notes:} The CATE is plotted with the solid black line; the proxy predictor $S(Z)$, produced by Random Forest, is plotted with the solid grey (light) line; and the  BLP is plotted with the dotted blue line. The left panel corresponds to the no heterogeneity example, $s_0(z)=0$ and the right panel to the strong heterogeneity example, $s_0(z)=z$. 
 In both panels, the BLP is less noisy than the ML proxy reducing the RMSE by $68\%$  and $5\%$.}\label{fig:examples}
\end{figure}

Figure \ref{fig:examples} provides two examples. The left panel shows a case  where $s_0(Z) = 0$ with zero effect and zero heterogeneity in the CATE, whereas the right panel shows a case  where $s_0(Z) =Z$ with strong heterogeneity in the CATE.   In both cases, we evenly split 1,000 observations between the auxiliary and main samples, $Z$ follows uniform distribution on $(-1,1)$, $b_0(Z) = 3Z$, $U$ is standard normal, independently of $Z$, and $Y$ is generated by \eqref{eq: HetPL1}.  We obtain the proxy predictor $S(Z)$  by Breiman's random forest, using the ranger implementation in R \citep{ranger17}. 

In the first example, the ML proxy is pure noise by construction, and the BLP post-processor correctly eliminates the noise, producing a CATE prediction that is roughly a 68\% better approximation to the CATE under the RMSE metric.  Furthermore, using our inferential methods of Section 4,  we cannot reject the null hypothesis that the BLP is zero.    In the second case, under strong heterogeneity, the signal in the ML proxy dominates the noise component.  As a result, the BLP does not change the ML proxy drastically, but still gives a meaningful improvement to the CATE under the RMSE metric.  These improvements agree with the theoretical arguments given above.\footnote{To show that the risk reductions are not a fluke, we repeated the calculations in $1,000$ simulations. We found average risk reductions of $65\%$ and $18\%$ in the no heterogeneity and strong heterogeneity examples, respectively.}  

\begin{remark}[Significance for RCTs.] The first example has implications for the empirical analysis of RCTs.  Here we see that one of the best ML algorithms, as per \cite{ESL}, can easily suggest a  heterogeneous CATE when the treatment is, in fact, a placebo.  Placebos (ineffective treatments) are common occurrences in  real-world experiments, and our methodology provides a simple way to confirm that the CATE (and not just ATE) is indeed zero in such cases.
\end{remark}

\subsection*{Strategy B: HT BLP}

This strategy  makes use of the Horvitz-Thompson transform $H$ defined in \eqref{eq:Htransform}.
It is well known that the transformed response $YH$ provides an unbiased
signal about CATE:
\begin{equation*}
\Ep [Y H \mid Z ]  = s_0(Z),
\end{equation*}
and it follows by the properties of the best linear predictor that 
$$ \BLP_{}[s_0(Z) \mid S(Z)] = \BLP_{} [YH \mid S(Z)].$$ 
Note that $\BLP_{}[s_0(Z) \mid S(Z)]$ is a more precise unbiased predictor of  $s_0(Z)$ than $YH$  because, by construction
$$
\Var(\BLP_{}[s_0(Z) \mid S(Z)]) = \Var(\BLP_{}[YH \mid S(Z)])  \leq  \Var(YH).
$$
The R-squared of $\BLP_{} [YH \mid S(Z)]$ quantifies the percent reduction in variance of the BLP relative to $YH$.

The simple linear projection $\BLP_{} [YH \mid S(Z)]$ is completely fine for identification purposes, but can severely underperform 
in estimation and inference due to lack of precision.  We can repair the deficiencies by considering, instead, the linear projection:
\begin{equation}\label{equation_HTreg}
YH = \mu_0' X_1 H + \mu_1  + \mu_2 (S - \Ep_{} S) + \epsilon,  \quad  \Ep \epsilon \tilde X =0,
\end{equation}
where $\tilde X := (X_1' H, \tilde X_2 ')'$,
$ \tilde X_2 := (1, S- \Ep_{} S)'$, and $X_1 := [1, B(Z), p(Z), p(Z) S(Z)]'$ as before. The term $X_1$ could contain other functions of $Z$. We include $X_1H$ in order to \textit{reduce noise}.



The following theorem shows that the linear projection \eqref{equation_HTreg} also identifies the BLP.
\begin{theorem}[BLP Identification B]\label{theorem: BLP2}  Consider $z \mapsto S(z)$ and $z \mapsto B(z)$ as fixed maps.  Assume that $Y$ has finite second moments,  $\tilde X$ is such that $\Ep \tilde X \tilde X'$ is finite
and full rank, which requires $\Var(S(Z)) > 0$. Then,  $(\mu_1, \mu_2)$ defined in \eqref{equation_HTreg} identifies the coefficients of the BLP,
$$\mu_1 = \beta_1, \quad \mu_2 = \beta_2.$$ 
\end{theorem}

Theorem \ref{theorem: BLP2} leads to an  estimator defined through the empirical analog:
\begin{equation}\label{eq:BLP estimate2}
Y_iH_i = \hat \mu_0' X_{1i} H_i + \hat \mu_1  + \hat \mu_2 (S_i -  \Bbb{E}_{N,M} S_i) + \hat \epsilon_i, \ \ i \in M, \ \   \Bbb{E}_{N,M} \hat\epsilon_i \tilde X_i =0,
\end{equation}
where $\tilde X_i := (X_{1i}' H_i, \tilde X_{2i} ')'$,
$ \tilde X_{2i} := (1, S_i- \Bbb{E}_{N,M} S_i)'$, and $X_{1i} := [1, B(Z_i), p(Z_i), p(Z_i) S(Z_i)]'$

\begin{remark}[BLP estimators] The properties of the estimators in \eqref{eq:BLP estimate} and \eqref{eq:BLP estimate2}, conditional on the auxiliary data, are  given in Lemma \ref{lemma: estimation} of the OA. 
\end{remark}

\begin{remark}[Comparison A vs. B] While the identification strategies A and B are natural, one may wonder whether the two corresponding  estimation strategies can be ranked in terms of asymptotic efficiency. We show in Appendix \ref{app:sec3} of the OA that they produce estimators that are first-order equivalent in large main samples. 
\end{remark}

\subsection{Sorted Group Average Treatment Effects}

The second inferential target is the set of group average treatment effects, where the groups are induced by $S(Z)$.

\begin{definition}[GATES] 
 The Sorted Group Average Treatment Effects (GATES) are 
$$
\gamma_k :=  \Ep_{}[ s_0(Z) \mid G_k  ], \quad k=1, \ldots, K.
$$
where
 $
  G_k :=  \{ S \in I_k\},
 $
with $I_k:=[\ell_{k-1}, \ell_{k})$ and
$
-\infty = \ell_0 < \ell_1 < \ldots < \ell_K= + \infty.
$
\end{definition}


\begin{remark}[Choice of groups] We  build the groups to explain as much variation in $s_0(Z)$ as possible. 
 There are many alternatives for creating groups based upon ML tools applied to the auxiliary data.
For example, one can  group or cluster based upon predicted baseline response as in the ``endogenous stratification" analysis \citep{abadie2013endogenous}, or based upon actual
predicted treatment effect $S$.  We focus on the latter approach for defining groups, although our identification and inference ideas immediately apply to other ways of defining groups, and could be helpful in these contexts. The causal tree approach of \cite{athey:trees} can also be viewed as a  GATES analysis, with a specific way of forming groups  via recursive partitioning.\footnote{Another strand of the literature related to the GATES is the learning policy problem, where a ML method is trained to assign units to treatment and control based on their covariates \citep[e.g.,][]{kitagawa2018should,athey2021policy}.  This problem can be seen as a GATES analysis with two groups chosen to maximize some function of the CATEs.}
\end{remark}

\begin{remark}[GATES as Predictors of CATE] The GATES can also be used as nonlinear predictors of the CATE based on the proxy $S$, in a similar fashion to the BLP. Indeed, the GATES provide the BLP of CATE using the group indicators $G_k, k = 1,\ldots, K$.  
\end{remark}



We provide two strategies for identifying and estimating the GATES. 

\subsection*{Strategy A: Weighted Residual GATES}
Consider  the weighted linear projection equation:
\begin{equation}\label{equation: group reg}
Y = \alpha_0'X_1 + \sum_{k=1}^K \alpha_k \cdot [D-p(Z)] \cdot  1(G_k) + \nu, 
\quad \Ep [ w(Z) \nu W] = 0,
\end{equation}
where $ W := (X_1', W_2')'$,  
$X_1$ contains a vector of functions of $Z$, e.g., $X_1 = (B(Z), p(Z)  \{1(G_k)\}_{k=1}^K)'$  and $W_2 := (\{ [D-p(Z)] \cdot \{1(G_k)\}_{k=1}^K )'.$ The presence of $D-p(Z)$ in the interaction $[D- p(Z)] \cdot  1(G_k) $ \textit{orthogonalizes} this regressor relative to all other regressors that are functions of $Z$, such as $X_1$. The controls in $X_1$, as in the BLP, are included to reduce noise in estimation.


Theorem \ref{theorem: GATES} below shows that the  linear projection \eqref{equation: group reg} identifies the GATES. We can therefore base an estimation strategy
on the empirical analog:
\begin{equation}\label{equation: estimate GATES}
Y_i = \hat \alpha_0'X_{1i} + \hat \alpha' W_{2i} + \hat \nu_i, \quad i \in M,  \ \ \  \Bbb{E}_{N, M}[ w(Z_i) \hat \nu_i W_i] = 0,
\end{equation}
where $\hat \alpha = (\hat \alpha_1, \ldots, \hat \alpha_K)'$.




\subsection*{Strategy B: HT GATES}

Here we employ a linear projection on Horvitz-Thompson transformed variables:
\begin{equation}\label{equation: group HT reg}
YH = \mu_0 ' X_1H + \sum_{k=1}^K \mu_k \cdot  1(G_k) + \upsilon, \quad \Ep [\upsilon \tilde W] = 0,
\end{equation}
where $ \tilde W := ( X_1' H, \tilde W_2')'$, $X_1$ includes functions of $Z$, e.g. $X_1$ the same as above, and $\tilde W_2 := [\{1(G_k)\}_{k=1}^K ]'.$ 

Theorem \ref{theorem: GATES} shows that the linear projection \eqref{equation: group HT reg} also identifies the GATES. We can therefore base an estimation strategy
on the empirical analog:
\begin{equation}\label{equation: estimate GATES 2}
Y_i H_i = \hat \mu_0'X_{1i} H_i + \hat \mu' \tilde W_{2i} + \hat \upsilon_i, \quad i \in M,  \ \ \  \Bbb{E}_{N, M}[ \hat \upsilon_i \tilde W_i] = 0,
\end{equation}
where $\hat \mu = (\hat \mu_1, \ldots, \hat \mu_K)'$. 
The resulting estimator has similar performance to the estimator in \eqref{equation: estimate GATES}, and under some conditions their first-order properties coincide.


We now provide a formal statement of the identification results. 

\begin{theorem}[GATES]\label{theorem: GATES} \textit{Consider $z \mapsto S(z)$ and $z \mapsto B(z)$ as fixed maps.  Assume that $Y$ has finite second moments and  $W$  and $\tilde W$  are such that $\Ep WW'$ 
and $\Ep \tilde W \tilde W'$ are finite and have full rank. Consider  $\alpha = (\alpha_k)_{k=1}^K$ defined by the weighted regression equation \eqref{equation: group reg} and $\mu = (\mu_k)_{k=1}^K$ defined by 
the regression equation \eqref{equation: group HT reg}. These coefficients are equal and identify the GATES:
$$
\alpha_k = \mu_k = \gamma_k = \Ep [s_0(Z) \mid G_k],  \quad k = 1, ..., K.
$$}
\end{theorem}

\begin{remark}[GATES estimators] The properties of the estimators in \eqref{equation: estimate GATES} and \eqref{equation: estimate GATES 2}, conditional on the auxiliary data, are  given in Lemma \ref{lemma: estimation} of the OA. 
\end{remark}

\subsection{Classification Analysis} When the BLP and GATES analyses reveal substantial heterogeneity,  it is interesting to know the properties of the subpopulations
that are the most and least affected. Here we focus on the ``least affected group" $G_1$ and ``most affected group" $G_K$,  where the labels ``most" and ``least" can be swapped depending on the context. 

\begin{definition}[CLAN] Let $g(Y,D,Z)$ be a vector of characteristics  of an observational unit. The classification analysis (CLAN) is the comparison of the average
characteristics of the most and least affected groups:
$$
\delta_1 := \Ep [ g(Y,D,Z) \mid G_1  ] \quad \text{ and } \quad \delta_K := \Ep [ g(Y, D, Z) \mid G_K  ].
$$
\end{definition}

The parameters $\delta_1$ and $\delta_K$ are identified because they are averages of variables that are directly observed. The CLAN quantifies the differences between the most and least affected groups and singles out the covariates that are associated with the heterogeneity in the CATE.  The CLAN can be extended to comparisons of features other than averages, such as variances, covariances or distributions.  In the empirical analysis, we estimate the CLAN parameters by taking averages in $M$:
\begin{equation}\label{eq: CLAN estimate}
\hat \delta_1 = \frac{\Bbb{E}_{N,M}  [ g(Y_{i},D_{i},Z_{i}) G_{1,i}]}{\Bbb{E}_{N,M} G_{1,i}} \quad \text{ and } \quad  \hat  \delta_K =  \frac{\Bbb{E}_{N,M} [ g(Y_{i},D_{i}, Z_{i}) G_{K,i}]}{\Bbb{E}_{N,M} G_{K,i}},
\end{equation}
using $G_{k,i} = 1\{S(Z_i) \in I_k\}$, where $I_k = [\ell_{k-1},\ell_{k})$ and $\ell_k$ is the $(k/K)$-quantile of $\{S_{i}\}_{i\in M}$.


\subsection{Goodness of Fit Measures for Fitting CATE} In practical applications it is useful to have goodness-of-fit measures to guide the selection of ML proxies. 


For the analysis based on the BLP of CATE, we propose to use:
\begin{equation}\label{define:lambda}
\Lambda := |\beta_2|^2 \mathrm{Var}(S(Z)) = \mathrm{Corr}( s_0(Z), S(Z))^2  \mathrm{Var}(s_0(Z)).
\end{equation}
Maximizing $\Lambda$ is equivalent to maximizing the correlation
between the ML proxy predictor $S(Z)$ and the CATE $s_0(Z)$, or equivalent to maximizing the $R^2$ in the regression of $s_0(Z)$ on $S(Z)$.  Therefore, an ML method that attains a higher $\Lambda$ is a preferred method. 

Analogously,  for the GATES analysis, we propose to use:
\begin{equation}\label{define:lambda2}
\bar \Lambda = \Ep \left( \sum_{k=1}^K \gamma_k 1(S \in I_k) \right)^2= \sum_{k=1}^K \gamma^2_k  \Pr(S \in I_k).
\end{equation}
 This is the part of variation  of $s_0(z)$, $\Ep s_0(Z)^2$,
  explained by $\bar S(Z) = \sum_{k=1}^K \gamma_k 1(S(Z) \in I_k)$. Hence choosing
  the ML proxy $S(Z)$ to maximize $\bar \Lambda$ is equivalent to maximizing the $R^2$ in the regression
  of $s_0(Z)$ on $\bar S(Z)$ (without a constant).    If the groups
 $G_k = \{S \in I_k\}$ have equal size, namely $\Pr(S(Z) \in I_k) = 1/K$ for each $k=1,..., K$, then
 $$
 \bar \Lambda  =  \frac{1}{K}\sum_{k=1}^K \gamma^2_k.
 $$
Therefore, a ML method that attains a higher $\bar \Lambda$ is a preferred method. The empirical versions of the parameters above are:
\begin{equation}\label{eq: GOOF estimate}
 \hat \Lambda = |\hat \beta_2|^2 \mathbb{E}_{N,M}(S_i- \mathbb{E}_{N,M} S_i )^2, \quad \quad  \widehat{\bar \Lambda}  =  \sum_{k=1}^K \hat \gamma^2_k  \mathbb{E}_{N,M} 1\{S_i \in I_k\}. \end{equation}

The choice of the ML method using goodness-of-fit measures does not pose any additional inferential challenge, when there is clearly an ML method that dominates the others -- so that we select the best method with probability approaching one. This means that the inferential methods of the next section would not need any further adjustment.  When this is not the case, there are several possibilities depending on the scientific reporting objectives. For example, suppose there are two (near) winners and we want to construct a $(1-\alpha)$-confidence set, as in our empirical analysis. Then in the spirit of sensitivity analysis, we can report the union of the $(1-\alpha)$-confidence sets.   This ensures the inferential coverage guarantee of $1-\alpha$ continues to apply, if 
the reader of the empirical report chooses one or the other winner at random. On the other hand, the inferential guarantee needs to be discounted to $1-2\alpha$ via Bonferroni adjustment, if 
the reader of the empirical report chooses one or the other report depending on the empirical results themselves.  We use the first approach because the readers of our empirical analysis are not likely to follow the latter approach.

Finally, if the data sets are big, we could use additional splitting to choose the best-performing ML method, before taking the resulting ML proxies to the main sample.\footnote{
We also refer to Section \ref{sec:further} which discusses other, more exploratory ideas for building the best ML algorithms for targeting CATE already in the first stage. }


%

%% file: Section4.tex
\section{Split-Sample Robust Estimation and Inference Methods}\label{sec:inference}

\subsection{Estimation and Inference: The Generic Targets}

Let $\theta$ denote a generic target parameter or functional.  For example,

\medskip
\begin{itemize}

\item 
$\theta = \beta_2 $ is the BLP slope, the heterogeneity loading parameter;

\item 
$\theta = \BLP_{}[s_0(Z) \mid S(z)] = \beta_1 + \beta_2 (S(z) - \Ep_{} S)$ is the ``personalized" predictor of CATE $s_0(z)$;

\item 
$\theta = \gamma_k$  is the GATES for the group $G_k$;

\item 
$\theta = \gamma_K - \gamma_1$ is the difference in the GATES between the most and least affected groups;

\item 
$\theta = \delta_K-\delta_1$ is the difference in the expectation of the characteristics of the most and least impacted groups in CLAN.

\end{itemize}

Let $(a,m)$ denote a fixed partition of $\{1,\ldots,N\}$.  In this section we make explicit when the probabilities and expectations are conditional on the auxiliary sample $a$, $$\mathrm{Data}_a := \{ (Y_i, D_i, X_i)\}_{i\in a},$$  and also index the  ML proxies $B= B_a$  and $S= S_a$ and estimands $\theta=\theta_a$ by $a$ to make explicit the dependence on $\mathrm{Data}_a$. As we mention in Comment \ref{remark:stability}, this dependence  vanishes as the size of the auxiliary sample becomes large under suitable conditions.


\subsection*{Single Split} We begin the discussion of inference conditional on a single split of data induced by the partition $\{(a,m)\}$ of  $\{1,...,N\}$ into sets of cardinality $(N-n, n)$. All of the examples admit an estimator $\hat \theta_a$  that is approximately Gaussian, conditionally on $\mathrm{Data}_a$, namely as $(N-n, n) \to \infty$ and for any $z$,
\begin{equation}\label{eq:onesplit}
\Pr ( \hat \sigma_a^{-1}(\hat \theta_a - \theta_a) < z\mid   \mathrm{Data}_a) \to_P \Phi (z).\end{equation}
We provide sufficient regularity conditions for (\ref{eq:onesplit}) in Lemma \ref{lemma: estimation}, which may be of independent interest (in single-split inference context).\footnote{The conventional asymptotic normality of the OLS estimator does not imply validity of the single-split inference procedure automatically, because the distribution of the estimator is random. We need to guarantee that this random distribution converges in probability to a normal  (non-random), which requires conditions for splits being "regular" with high probability. We refer to Lemma \ref{lemma: estimation} in the OA for the formal result.}


As a consequence of \eqref{eq:onesplit}, the standard p-values
 $$
p^+_a :=  1-\Phi( \hat \sigma^{-1}_a (\hat \theta_a - \theta_0)), \ \
p^-_a :=  \Phi( \hat \sigma^{-1}_a (\hat \theta_a - \theta_0)),
$$
for testing the null hypothesis $ \theta_a = \theta_0$ against the alternatives $\theta_a>\theta_0$ and  $\theta_a<\theta_0$, respectively, are approximately uniform under the null, namely 
$$
\Pr (p^{\pm}_a < \alpha \mid \mathrm{Data}_a) = \alpha + o_P(1).
$$
As another consequence of \eqref{eq:onesplit}, the  standard confidence interval  (CI)
$$
 [L_a, U_a] :=  [\hat \theta_a \pm  \Phi^{-1}(1-\alpha/2) \hat \sigma_a ] 
$$
covers $\theta_a$ with approximate probability $1-\alpha$ conditional on $\mathrm{Data}_a$:
$$
\Pr [\theta_a \in [L_a, U_a ] \mid  \mathrm{Data}_a] = 1-\alpha -o_P(1).
$$
Thus, we have straightforward inference conditional on a single data split. 

\subsection*{Multiple Splits} In practice, researchers often prefer using multiple splits $(a,m)'s$ to reduce estimation risk and demonstrate that the results are robust to how they split the data.  Therefore, we need a way to aggregate the results across different splits, and propose quantile aggregation methods and analyze their properties.

\begin{definition}[Collection of Splits]  Consider the collection  
$\{(a, m), a \in \mathcal{A}\}$ of partitions of $[N]=\{1,..., N\}$ into auxiliary sets $a$ of cardinality $N-n$ and main sets $m$ of cardinality $n$.  We generate the collection independently of  
$$\mathrm{Data} := (Y_i, D_i,X_i)_{i=1}^N.$$
\end{definition}

Different partitions of $[N]$ yield different estimands and estimators.  To formalize this randomness, we consider $A$ as a uniform random variable taking values $a \in \mathcal{A}$, that is, $ A \sim U(\mathcal{A})$. Therefore, conditional on $\mathrm{Data}$, the estimand $\theta_{A}$ is a random variable.  In what follows, we will mainly target our inference on the median value of $\theta_A$, instead of $\theta_a$, the estimand for a specific partition.  Furthermore, different partitions yield different estimators $\hat \theta_A$ and approximate distributions for these estimators.  Therefore, conditional on $\mathrm{Data}$, estimators $\hat \theta_A$, p-values $p_A$, and intervals $[L_A, U_A]$ are all random variables.  We will use quantile aggregation methods to summarize them.

It is useful to recall some  definitions of quantiles for discrete variables.  For a random variable $X$ with law $\Pr_X$ and $k$ points of support, and index $u \in (0,1)$, the lower and upper $u$-quantiles are $
\mathrm{\underline{Q}}_u(X)  := \inf\{ x \in \Bbb{R}:    \Pr_X(X \leq x) \geq u \},$ and $ \mathrm{\overline{Q}}_u(X) : = \sup\{ x \in \Bbb{R}:   \Pr_X(X \geq x) \geq  1-u)$, respectively.
For $w_u := \lfloor u k \rfloor/(\lfloor u k \rfloor+ \lceil u k \rceil)$, we define $$\mathrm{{Q}}_u(X) : = w_u \mathrm{\underline{Q}}_u(X) + (1-w_u)\mathrm{\overline{Q}}_u(X)$$ as the central quantile.\footnote{For example, the quantile function in \textsf{R} uses this definition \citep{Rcite}.}  If $X$ is continuous, all three definitions coincide.  To define upper,  lower, and central medians, we use $u=1/2$ in the definitions above and $\mathrm{M}$ instead of $\mathrm{Q}_u$.

We now formally define the median-aggregated p-value.

\begin{definition}[Median-Aggregated P-value] The median p-values for testing one-sided alternative hypotheses are
$$
 p^{\pm}={\mathrm{M}} (p^\pm_A  \mid \mathrm{Data}).
$$
The two-sided median p-value is 
$\bar p_{} = 2\min( p^{+}_{}, p^{-}_{}).$
\end{definition}

Aggregation using  lower median p-values was first proposed by \cite{meinshausen2009p} in the context of split-sample hypothesis testing in linear regression with selection. Here we take the central medians, since they are more likely to behave like regular p-values.\footnote{For example, the sample lower median of $\{U,1-U\}$, $U \sim U(0,1)$, obeys $P(\underline{\mathrm{M}} < \alpha) = 2\alpha$ for $\alpha< 1/2$. In contrast, the central median obeys 
$\Pr({\mathrm{M}} < \alpha) < \alpha$ for $\alpha < 1/2$.}

We next define the quantile-aggregated point and interval estimators.

\begin{definition}[Quantile-Aggregated Point and Interval Estimators]  The median point estimator is:
$$
\hat \theta := \mathrm{M}[\hat \theta_A \mid \mathrm{Data}].
$$
The $\beta$-quantile confidence interval is $[L,U]$, where
$$
L :=  {{Q}}_\beta (L_A \mid \mathrm{Data}), \quad
U := {{Q}}_{1-\beta} (U_A \mid \mathrm{Data}), \quad \beta \leq 1/2.
$$
 \end{definition}
We can interpret these definitions as risk-reducing inferential summaries.

\begin{lemma}[Risk Contraction]\label{lemma:risk}
Consider any fixed target value $\theta' \in \Bbb{R}$. Then $\hat \theta$ is more concentrated near $\theta'$ than any single-split generated $\hat \theta_A$:
\begin{equation}\label{eq:risk}
\Ep  | \hat \theta- \theta'|
\leq \Ep | \hat \theta_A- \theta'|.
\end{equation}
Set $\beta=1/2$. Then the confidence set $[L,U]$ has the same concentration property: 
\begin{equation}\label{eq:risk2}
 \Ep | U- \theta'| \vee
  \Ep | L- \theta'| \leq 
  \Ep | U_A- \theta'| \vee
  \Ep | L_A- \theta'|.
\end{equation}
Moreover, for any $\beta \in (0, 1/2]$ the width of $[L,U]$ is weakly smaller than the worst-case width of the sets across splits: 
\begin{equation}\label{eq:risk3}
|U - L|
\leq \sup_{a \in \mathcal{A}}|  U_a - L_a|.
\end{equation}
\end{lemma}
Analogous risk contraction properties hold for the mean aggregation, but we focus on medians for robustness reasons.

In what follows, we study the  formal inferential guarantees of $[L,U]$.   The default choice of $\beta$ is $1/2$, but we obtain useful theoretical guarantees for other choices $\beta< 1/2$ as well.  

\subsection*{Principal Regularity Condition}

As the main regularity condition, we assume approximate normality of the split-sample t-statistics:\footnote{Here and below we use the standard error $\hat \sigma_A$ instead of the theoretical standard deviation $\sigma_A$ in all statements, but we can exchange the two if $\hat \sigma_A/\sigma_A \to_P 1$, which holds under typical conditions, e.g. Lemma \ref{lemma: estimation}.}
\begin{itemize}
    \item [(R1)] There exist a sequence of positive constants  $\gamma'_N \searrow 0$ as $(n,N-n) \to \infty$, such that 
\begin{equation}\label{reg:basic}
\sup_{z \in \Bbb{R} }|\Pr\{ \hat \sigma^{-1}_A (\hat \theta_A - \theta_A) < z \} - \Phi(z)| \leq \gamma'_N.
\end{equation}
\end{itemize}

Suppose that the data $\{(Y_i, Z_i, D_i)\}_{i=1}^N$ are generated as i.i.d. copies of $(Y, Z, D)$.  In this case, for any $a \in \mathcal{A}$:
$$
\Pr\{ \hat \sigma^{-1}_A (\hat \theta_A - \theta_A) < z \} = \Ep \left [\frac{1}{|\mathcal{A}|} \sum_{a \in \mathcal{A}} 1(\hat \sigma^{-1}_a (\hat \theta_a - \theta_a) < z) \right] = \Pr\{ \hat \sigma^{-1}_a (\hat \theta_a - \theta_a) < z \},
$$
because the expression on the right does not depend on $a \in \mathcal{A}$ under the i.i.d. sampling. This observation simplifies the verification of (R1) for least squares type estimators; see Lemma \ref{lemma: estimation}. While the i.i.d. case is our main focus, the main results in this section rely only on the conditions labeled as R, which are likely to hold more generally.

Below we give various theoretical guarantees for our inferential summaries using this condition and adding more conditions to get stronger results.

\subsection{Hypothesis Testing with Multiple Splits}  We start the analysis by  testing homogeneous hypotheses $\theta_A = \theta_0$, which imply that $\theta_a$ does not vary with $a$. Suppose, for example, that we want to test that the slope of the BLP is zero with probability one,
$\beta_{2A} = 0,
$
against the alternative $\beta_{2A}>0$ with positive probability. This problem amounts to both testing the heterogeneity in CATE and the relevance of the ML score $S_A$ as a predictor. Another interesting hypothesis is whether $\beta_{2A} = 1,$ with probability one, that is, whether $S_A$ is well-calibrated and needs no post-processing.

More generally, suppose we are testing the hypothesis 
\begin{equation}\label{eq:H0}
H_0: \theta_A = \theta_0,
\end{equation}
with probability one, against $H^+_1: \theta_A>\theta_0$ with positive probability. Testing using the median p-value $p^+$ will have power against the null when the majority of $\theta_a$'s violate the null, so we can interpret the rejection accordingly. 
Similarly, we can test against $H^-_1: \theta_A<\theta_0$ or $H_1: \theta_A \neq \theta_0$ with positive probability using the median p-values $p^-$ or $\bar p$. 


Below we establish the properties of the median $p$-values under (R1). To get the sharpest results, we can invoke a concentration condition for approximate medians:

\begin{itemize}
\item[(R2)] For all $z = \Phi^{-1} (\alpha)$, where the nominal level
 of interest $\alpha$ is in some closed sub-interval of $(0, 1/4)$, and some sequences of positive constants $\gamma''_N \searrow 0$ and non-negative constants $\varepsilon_N \searrow 0$:
\begin{equation}\label{reg:concentrate median 2}
\begin{array}{lll}
& \Pr \left(  \mathrm{ {Q}}_{.5- \varepsilon_N}(\hat \sigma^{-1}_A  (\theta_A - \hat \theta_A)|\mathrm{Data}) < z \right) & \leq \Pr \left( \hat \sigma^{-1}_A (\theta_A - \hat \theta_A)<z \right)+ \gamma''_N, \\
& \Pr \left(  \mathrm{{Q}}_{.5- \varepsilon_N}(\hat \sigma^{-1}_A  (\hat \theta_A -  \theta_A)|\mathrm{Data}) <  z \right) & \leq \Pr \left( \hat \sigma^{-1}_A (\hat \theta_A -  \theta_A) < z \right)+ \gamma''_N.
\end{array}
\end{equation}
\end{itemize}
This condition states that the approximate median over-the-splits t-statistics tend to concentrate more than any single-split t-statistic. This condition is intuitive, but it is hard to give general primitive conditions for it.\footnote{When the $t$-stats are  independent, then their median concentrates in a fixed interval around 1/2 with probability approaching 1  exponentially fast; see \cite{romano:SS}. Therefore, the approximate median concentration condition holds.  This happens when $m$'s are non-overlapping and some further homogeneity conditions hold. On the other hand, suppose the $p$-values are the same asymptotically, then the inequality in the concentration condition binds, but does not fail. This situation is not common in our context, though.}  We believe \eqref{reg:concentrate median 2} is quite plausible. We verified it for typical values of $\alpha< 1/4$ numerically in various experiments that mimic  empirical applications, and were unable to find any counterexample.

\begin{theorem}[Uniform Validity of Median-Aggregated P-Value]\label{theorem: PV}
Suppose that the null hypothesis $H_0$ in \eqref{eq:H0} holds with probability one. Let $p_{}$ be either of $\{p^+_{}, p^-_{}, \bar p_{}\}$.
(i) Suppose that  approximate normality (R1) holds, then  
$$\Pr (2 p_{} < \alpha) \leq \alpha+ o(1),$$
where the $o(1)$ depends only on $\gamma'_N$. (ii) Suppose in addition that the median concentration condition (R2) holds with $\varepsilon_N =0$, then 
$$
\Pr (p_{}<\alpha) \leq \alpha+ o(1),
$$
where the $o(1)$ depends only on $\gamma'_N$ and $\gamma''_N$.
\end{theorem}

Therefore under the median concentration condition, the median p-values have the standard property.  Without the median concentration condition, the median p-values need to be multiplied by 2.  However, based on our computational experiments, median p-values are conservative even for the nominal level $\alpha$ (once $\alpha< 1/4$), mainly due to the concentration property holding with $\gamma''_N <0$.  We therefore do not recommend multiplying by 2; see also \cite{romano:SS} for a similar point.  The exact form of $o(1)$, given in the proof, allows us to convert the results into those holding uniformly in a set of probability measures $\Pr$.  The proof of the first result partly relies on the idea of  \cite{meinshausen2009p} to use Markov inequality to bound quantiles of an arbitrary collection of marginally uniform random variables.

\subsection{Prediction Intervals with Multiple Splits}  \label{sec:MP}

Outside of the settings with homogeneity, the estimand $\theta_A$ is a random variable, and we might be interested in characterizing its typical values.  Our first approach serves this purpose, and is connected to conformal/permutation inference that is commonly used for predicting unobserved outcomes \citep{hoeffding1952large,vovketal,candes2022}.

Here our goal is to have a prediction interval for $\theta_A$, and the challenge we face is that $\theta_A$ is not observed directly, which places us outside the standard conformal setting. However, for each $a \in \mathcal{A}$, we have a (random) confidence interval $[L_a, U_a]$ that has the covering property:
\begin{equation}\label{eq:brackets}
\Pr (\theta_a < L_a) \leq \alpha/2 + o(1),  \quad \Pr (\theta_a > U_a) \leq \alpha/2 + o(1).
\end{equation}
This condition is implied by the basic regularity condition (R1) in our context.

We take the quantile-aggregated confidence interval $[L,U]$ as our prediction interval for $\theta_A$. 

\begin{theorem}[Properties of the Prediction Interval]\label{theorem:CI2} Suppose that (\ref{eq:brackets}) holds. Then 
$$
\Pr (\theta_A < L) \leq \beta+ \alpha/2 + o(1), \quad
\Pr (\theta_A > U ) \leq \beta+ \alpha/2 + o(1),
$$
where the $o(1)$ terms are the same as in (\ref{eq:brackets}). Therefore, $
\Pr  (\theta_A \in [L,U])
\geq 1- 2\beta-\alpha - 2o(1)$.   
\end{theorem}


We can use the prediction interval $[L,U]$ to characterize the ``majority" of the central values of the random target $\theta_A$ that one could get from sample splitting. For this we set $\beta=.25$ and "small" $\alpha=o(1)$, then $MP=[L,U]$ has the property:
\begin{equation}
\Pr (\theta_A \in MP) \geq .5 - o(1).
\end{equation}
That is, $MP$ contains majority of central values of $\theta_A$. On the other hand, if we set $\beta =1/2$, we get a median aggregated interval $$MI=[L,U].$$ In this setting, we can think of $MI$ as predicting the median of $\theta_A$ with small margin of error, if $\alpha=o(1)$:
\begin{equation}
\Pr(\theta_A < L) \leq 1/2 +  o(1) \text{ and } \Pr(\theta_A > U) \leq 1/2 +  o(1).
\end{equation} The latter gives us a useful interpretation of  median confidence intervals, and notably this interpretation applies under the weakest possible  regularity condition (R1) in our setting.

\subsection{Confidence Intervals for Median Parameter with Multiple Splits}

Instead of predicting the ``majority" of $\theta_A$, we may focus the inference on a single target.

\begin{definition}[Inferential Target] Our inferential target is the median estimand:
$$
\theta^* = \mathrm{M}[\theta_A |\mathrm{Data} ].
$$
\end{definition}
The choice of the target has the intuitive appeal of representing a typical value. Moreover, in many cases, $\theta_A$ will concentrate around its median value $\theta^*$, making it an even more natural target. What follows is the principal regularity condition that covers this concentration scenario.

\begin{itemize} 
\item[(R3)]For some positive sequences of constants $r_N \searrow 0$ and $\gamma'''_N \searrow 0$ as $(n,N-n) \to \infty$, \begin{equation}\label{reg:concentrate}
\Pr \left( \hat \sigma^{-1}_A |\theta^* - \theta_A| > r_N  \right) \leq \gamma'''_N.
\end{equation}
\end{itemize}

The concentration condition above is a convenient property for the interpretability of the inference. Here the rate of concentration of $\theta_A$ around  $\theta^*$ should be faster than the rate $\hat \sigma_A$ of $\hat \theta_A$ estimating $\theta_A$. Thus, the concentration condition implicitly requires the auxiliary set $a$ to be large and the main set $m$ to be small compared to $a$; so that randomness in the inferential target is small compared to the size of the estimation error $\hat \sigma_a$ in the main sample, which typically is proportional to $n^{-1/2}$. Otherwise, if  $n$ is large relative to $N-n$, then each $\theta_a$ would be estimated accurately, but there would be a lot of variation of $\theta_a$ across $a \in \mathcal{A}$.\footnote{As a practical diagnostic, we recommend the researchers to report the variation of $\hat \theta_a$ across sample splits, in addition to the confidence interval for $\theta^*$. We are grateful to Guido Imbens for this suggestion.}
Condition (R3) is high-level; we demonstrate the plausibility of this condition for the BLP parameter in Appendix \ref{sec:stability} of the OA using notions of estimation and algorithmic stability. \footnote{Estimation stability implies that $\theta_A$ concentrates around a fixed value $\theta_\bullet$, in which case the median also concentrates around $\theta_\bullet$. Estimation stability follows from the ML proxy $S_A$ being consistent for some fixed proxy function $s_\bullet$, but not necessarily consistent for the true CATE. This condition can be readily verified using statistical learning theory, as we do in Section \ref{sec:further} for causal learners of CATE. The algorithmic stability condition is strictly weaker than estimation stability, though not as readily available. Our use of these stability criteria is inspired by similar ideas in \cite{wager:PNAS}, \cite{CWZ:JASA}, and \cite{vasilis:stable}, applied to a different context. Appendix \ref{sec:stability} discusses all of this further.}

The following results summarize the properties of the proposed median confidence interval under various conditions.  

\begin{theorem}[Properties of the Confidence Interval for $\theta^*$]\label{theorem:CI}  Let $\beta=1/2$. 
(i) Suppose that (R1) and  (R3) hold. Then, 
$$
\Pr(\theta^* \in [L,U])  \geq 1- 2\alpha- o(1),$$
where $o(1)$ depends only on $\gamma'_N, \gamma'''_N$ and $r_N$. 
(ii) Suppose in addition that (R2) holds with $\epsilon_N =
2\sqrt{\gamma'''_N}$.  Then, 
$$
\Pr(\theta^* \in [L,U])  \geq 1- \alpha - o(1),
$$
where $o(1)$ depends only on $\gamma'_N, \gamma'''_N, \gamma''_N$ and $r_N$.   (iii) In either case, the event $\theta^* \in [L,U]$ implies $|\theta^*- \hat \theta| \leq |U- L|$.
\end{theorem}

Under the strongest assumptions, the target $\theta^*$ is covered with a probability of at least $1-\alpha-o(1)$. Under the minimal set of assumptions, the coverage probability is $1- 2 \alpha - o(1)$. In our numerical results, the confidence intervals tend to be conservative
even under the minimal condition, with coverage exceeding $1- \alpha$. Therefore,  using $1-\alpha$ as the nominal level is our recommended choice based on this evidence.

%% file: Section5.tex
\section{Further Consideration: Causal Machines that Learn CATE Better}\label{sec:further}

Our main proposal so far is to take proxies from any first stage black-box machine and post-process them to better target CATE and perform inference on functionals of CATE, such as the BLP and GATES.  But, can we design the machines to target CATE directly in the first stage?  If we can, then the post-processing methods of the previous section would mostly focus on providing inference, and less on correcting biases of the first stage inputs. Building on \cite{athey:trees}, we propose two types of such causal machines, and connect them to the emerging literature on orthogonal machine learning, such as \cite{nie:20}, \cite{semenova:panel}, and  \cite{foster2019orthogonal}, among others. 
 \subsection{Focusing ML Methods on CATE in Stage 1}\label{sec: Best First Stage}  
 
We propose two options, taking ideas from our stage 2 analysis to stage 1. Specifically, we can train ML proxies in the auxiliary sample based on either:
\begin{itemize}
    \item[(A)] Minimizing $w(Z)$-weighted square prediction errors of $Y$ on  $B$ and $(D-p(Z)) S$;
    \item[(B)] Minimizing square prediction errors of $YH$ on  $BH$ and $S$;
\end{itemize}    
where $B(Z)$ is now a technical ``baseline" function of covariates $Z$, as described below, whose role is to reduce noise in the learning problem. 

\begin{definition}[Causal Learners for Stage 1] We can solve either of:
\begin{align}\label{eq:A}\tag{A}
& (B,S)  \in \arg \min_{b \in \mathcal{B}, s \in \mathcal{S}} \quad \sum_{i \in A} w(Z_i) [Y_i - b(Z_i) -  \{D_i-p(Z_i)\} s(Z_i) ]^2, \\
&(B,S)  \in  \arg \min_{b \in \mathcal{B}, s \in \mathcal{S}} \quad \sum_{i \in A} [Y_i H_i - b(Z_i) H_i - s(Z_i)]^2,\label{eq:B}\tag{B}
\end{align}
where $w(Z) = [p(Z)(1-p(Z))]^{-1}$, and $\mathcal{B}$ and $\mathcal{S}$ are functional parameter spaces.\end{definition}

 We can refer to the first causal learner as the weighted residual (WR) learner, and the second causal learner as the HT  learner. Examples of parameter spaces include spaces of linear functions generated by a set of dictionary transformations of $Z$, reproducing kernels, linear combinations of decision trees, neural networks, and others. In (A) the parameter spaces are meant, but not required, to contain the functions $z \mapsto \tilde b_0(Z) := b_0(z) + p(z) s_0(z)$ and $z \mapsto s_0(Z)$. In (B) the parameter spaces are meant, but not required, to contain the functions $z \mapsto \bar b_0(Z) := b_0(Z) + (1-p(z)) s_0(z)$ and $z \mapsto s_0(Z)$.  

Both (A) and (B) improve over the standard predictive learners that predict $Y$ using the best approximation to $ \Ep[Y \mid D, Z]$  in a given class, but not necessarily the best approximation to the CATE $s_0(Z)$ itself,  and may be of independent interest. Moreover, the loss functions in (A) and (B) are also helpful for validation purposes, and choosing the best or aggregating classes of ML methods for targeting the CATE function. 

The proposal (B) generalizes and refines the strategy of \cite{athey:trees} of predicting $YH$ using (a tree form of) $S$ by introducing denoising by $B$. \cite{semenova2020} developed a related HT strategy that applies series/sieve learners to the denoised HT-transformed outcome, but it explicitly relies on consistent estimation of the regression function, unlike our approach.   We further discuss connections of proposal (A) to the unweighted residual learners of \cite{semenova:panel}, \cite{nie:20}, and others below.\footnote{Both of our proposals appeared to be new around the first circulation of this paper as ArXiv:1712.04802.  See links to the recent literature below, which proposed related, but different ideas. Relative to the initial version of the paper, the current version contributes with several formal learning properties of (A) and (B). We are grateful to the referees for suggesting that we develop these properties.} 

\begin{theorem}[Oracle Properties of the Population Objective Functions]\label{lemma:oracle} Suppose that  $Y$, $b(Z)$, $s(Z)$, and $w(Z)$ are square integrable. (1) Then, the  expectations of the loss functions in \eqref{eq:A} and \eqref{eq:B} are
\begin{eqnarray}
 \Ep w(Z) [Y - b(Z) -  (D-p(Z)) s(Z) ]^2 &=& \Ep[s_0(Z) - s(Z)]^2 + C_{1b} \label{obj2},  \\
 \Ep [Y H - b(Z) H - s(Z)]^2 &=&
\Ep[s_0(Z) - s(Z)]^2 + C_{2b}, \label{obj1}
\end{eqnarray}
where $C_{1b}:=  \Ep[w(Z) (\tilde b_0(Z) - b(Z))^2]+C_1$ and $C_{2b}:= \Ep[w(Z) (\bar b_0(Z) - b(Z))^2]+C_2$ for some constants $C_1$ and $C_2$.  
(2) Therefore, the minimizers, say  $s_\bullet(Z)$, of the left-hand sides of (\ref{obj2}) and (\ref{obj1}) over $s \in \mathcal{S}$, if exist, also minimize the oracle loss function $
\Ep[s_0(Z) - s(Z)]^2$ over the same set. 
\end{theorem}

Theorem \ref{lemma:oracle} shows that the minimizers of the two loss functions provide the best approximation in the mean-square sense to the actual CATE function $s_0(Z)$ in the class $\mathcal{S}$.  This property occurs even though we do not observe $s_0(Z)$, and such performance is usually qualified as ``oracle."  A sufficient condition for the existence of minimizers is that the set $\mathcal{S}$ be convex and closed in the $L^2(P)$ norm.

\begin{figure}
	\centering
 	\caption{Comparison of Predictive Machine Learners vs Causal Learners Based on (A) and (B). }\label{fig:targeting}
\includegraphics[width=.5\textwidth]{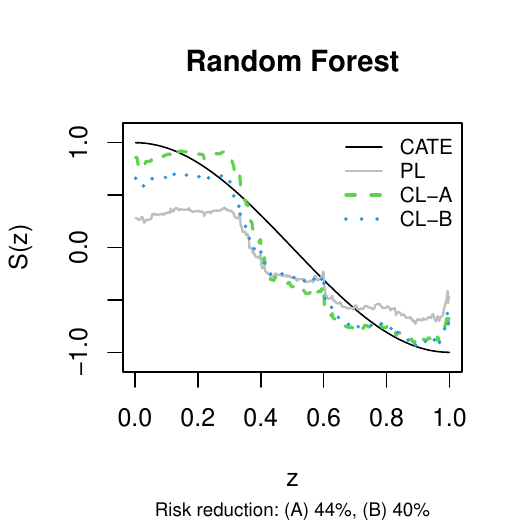}\includegraphics[width=.5\textwidth]{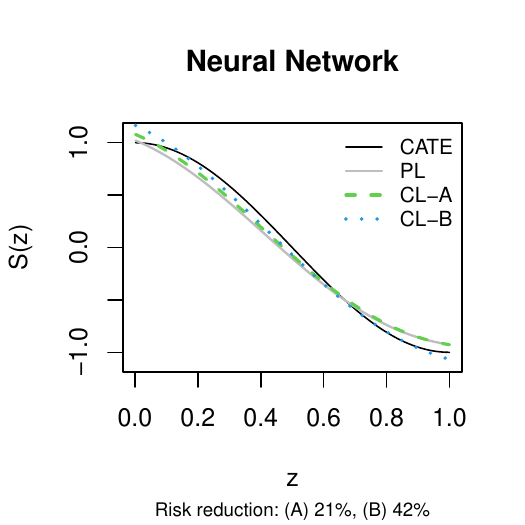}
 \caption*{\footnotesize \textsc{Notes:} The solid black curve is the  CATE function $s_0(Z)$, and the solid grey (light) curve is the predictive learner $S(Z)$ (PL) obtained by random forest and neural network. The dashed green and dotted blue curves are estimators $S(Z)$ produced by the causal learners of CATE based on solving objectives (A) and (B) (CL-A and CL-B) .   The underlying data is generated as $Y_i =b_0(Z_i) + s_0(Z_i)D_i + \xi_i$, where $\xi_i \sim N(0,1/4)$, $Z_i \sim U(0,1)$, $D_i$ is Bernoulli with success probability $p(Z)=.1 \cdot 1\{Z< 1/2\} + .5 \cdot 1\{Z \geq 1/2\}$, $b_0(z)= z$, $s_0(z) = \cos(2 \pi z)$, and $i =1,.., 500$. }
\end{figure}

We illustrate the benefits of using the causal learning objectives (A) and (B) in Figure \ref{fig:targeting}.  In the two panels, we compare the CATE learners derived from the standard predictive Random Forest and Neural Network with the Causal Learners from Random Forests and Neural Network that solve the objective functions \eqref{eq:A} and \eqref{eq:B}.  We find that the causal learners are better at approximating the CATE function, thereby providing better proxies for CATE.  The improvements in the RMSE of approximating the CATE provided by the causal learners range from $21\%$ to $44\%$.

\begin{figure}
	\centering
 \caption{Comparison of OLS and Causal Forest as Predictive Learners vs OLS Causal Learners and Forest Causal Learners Based on (A) and (B). }\label{fig:targeting2}
\includegraphics[width=.5\textwidth]{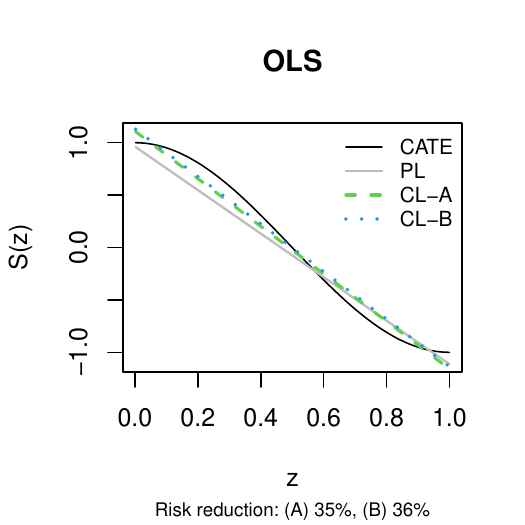}\includegraphics[width=.5\textwidth]{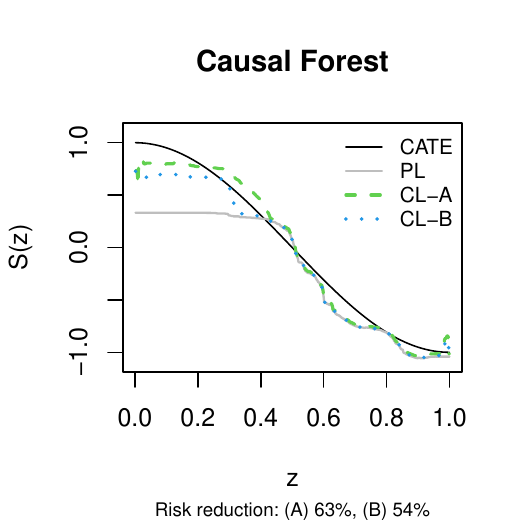}
 \caption*{\footnotesize \textsc{Notes:} The solid black curve is the  CATE function $s_0(Z)$, and the solid grey (light) curve is the predictive CATE learner $S(Z)$ (PL) obtained by  OLS and Causal Random Forest. The dashed green and dotted blue curves are estimators $S(Z)$ produced by the causal learners of CATE based on solving objectives (A) and (B) (CL-A and CL-B) .   The underlying data is generated in the same way as in fig. \ref{fig:targeting}. }
\end{figure}

Likewise, the left panel of Figure  \ref{fig:targeting2} shows that we can improve the standard predictive OLS by the Causal OLS that solves the objective functions (A) and (B).  Here we estimate linear  models in $Z$ for the baseline function and CATE.  The improvement in the RMSE of approximating the CATE provided by causal OLS is about $35\%$.  This finding might be of interest to researchers using OLS in empirical work.

Finally, the right panel of Figure  \ref{fig:targeting2} shows that one can improve the Causal Forest by a causal boosting step that solves the objective functions (A) and (B) by looking for a shallow forest deviation away from the cross-fitted Causal Forest proxy.  The improvements in the RMSE of approximating the true CATE provided by this step are $54-63\%$.  The explanation for this improvement is that the Causal Forest, while explicitly targeting CATE, actually solves a different objective than (A) or (B).\footnote{In our understanding it performs a residual learning approach, but not the weighted residual learning approach, which makes the method under-perform relative to the Forest Causal Learner based upon (A) or (B).}  We also verified that this improvement only applies when the propensity score is not constant, like in this example.

\begin{remark}[Connections to the Literature]\label{remark:wager}
Residual learning like (A), but without weighting by $w(Z)$, appears in the debiased machine learning  of Robinson's partially linear model in \cite{DML}.\footnote{Also called "partialling out", residual, and orthogonal learning, building upon classical econometric ideas due to Frisch-Waugh-Lovell and \cite{robinson}. Note  \cite{DML} presents other strategies as well, with residual learning being just one of them.} In the nonparametric setting, \cite{nie:20} and \cite{semenova:panel} also propose the 
unweighted version of type (A) objection function.   Both of these papers target CATE learning in non-experimental settings.  Their proposal does not use weighting by  $w(Z)$ as ours and therefore does not provide the best approximation property to the CATE in population.  However, it is easy to verify that their proposal provides the best approximation to CATE weighted by $p(Z)(1-p(Z))$:
$$
\min_{s \in \mathcal{S}} \Ep[ p(Z)(1-p(Z)) (s_0(Z) - s(Z))^2]. 
$$
In contrast to our proposal, this weighting gives relatively less attention to units that are either more likely or less likely to be treated than units that are equally likely.  While this property is not appealing in general, in pure RCTs with constant propensity score $p(Z)$, the objective function above reduces to the best approximation to CATE.  Moreover,  weighting by $w(Z)$ in  (A) plays less important role when $s_0$ can be estimated consistently, as in \cite{nie:20},  \cite{semenova:panel}, and \cite{foster2019orthogonal}.  
 We are interested, however, in the high-dimensional settings in which consistent learning of $s_0$ might not be possible. Therefore, below we provide formal estimation results for causal learners (A) and (B) under this ``agnostic" setting. Our inference methods of Section \ref{sec:inference} also apply to the causal learners (A) and (B) used as first-stage proxies.
\end{remark}

\subsection{Learning Guarantees} The learning guarantees of the causal learners \eqref{eq:A} and \eqref{eq:B} follow from the state-of-art statistical learning theory \citep*{rakhlin:offset}, in particular through the use of the expected off-set Rademacher complexity (ORC).

\cite{rakhlin:offset} define the expected ORC of the function class $\mathcal{H}$ as:
$$
\mathcal{R}^o(A, \mathcal{H},c)
:= \Ep \sup_{h \in \mathcal{H}} \frac{1}{|A|} \sum_{i \in A} \left[ e_i h(Z_i) - c h(Z_i)^2 \right],
$$
where $\{e_i\}$ are i.i.d. Rademacher variables taking values $-1$ and $1$ with probability $1/2$, that are generated independently of the data $\{Z_i\}_{i \in A}$, and $c>0$ is a positive constant.  ORC is a statistical measure of complexity that captures the ability of the functional class to fit i.i.d.  Rademacher noise.  The more complex the class is, the higher the ORC.  As shown in \cite{rakhlin:offset},  the expected ORC naturally scales like $ d/|A|$, where $d$ is the effective dimension of the function class and $|A|$ is the sample size. For example, for linear classes, $d$ is the actual dimension of the linear class; and for VC classes, $d$ scales like the VC dimension. 
\cite{rakhlin:offset} show that ORC is the sharpest characterization of complexity: in particular, the ORC is upper bounded by the standard critical radii of function classes defined in terms of local Rademacher complexity  \citep[e.g.,][]{wainwright:book} or in terms of the standard uniform covering entropy \citep{Dudley2000}. This makes the ORC bounds readily available for all function classes used in modern ML.

The following result establishes formal learning guarantees for causal learners in randomized experiments under general settings that do not assume we can learn $s_0$ consistently.

\begin{theorem}[Near-Oracle Guarantees for Causal Learners]\label{lemma:rate}  Suppose that  $Y$, the elements of $\mathcal{B}$ and $\mathcal{S}$, and $w(Z)$ are bounded in absolute values by $K$, and $\mathcal{B}$ and $\mathcal{S}$ are closed, convex, and symmetric sets.   The estimator $S$ obtained as a solution of either \eqref{eq:A} or \eqref{eq:B} is as good as using  the best in class approximation, say $s_\bullet(Z)$, up to an error expressed in terms of ORC:
\begin{equation}
0 \leq \Ep[s_\bullet(Z) - S(Z)]^2 
 \leq \underbracket{\Ep[s_0(Z) - S(Z)]^2 - 
\overbracket{\Ep[s_0(Z) - s_\bullet(Z)]^2}^{\mathrm{oracle \ \ risk}}}_{\mathrm{excess \ \ risk}} \leq  C_K \mathcal{R}^o(A, \mathcal{H}, c_K),
\end{equation}
where $C_K$ and $c_K$ are positive constants that only depend on $K$,   $\mathcal{H}:= 4( w(Z)^2\mathcal{B} +H w(Z)\mathcal{S})$ for type (A) loss, and $\mathcal{H}:= 4(H\mathcal{B}+ \mathcal{S})$ for type (B) loss.
\end{theorem}

The result shows that if the ORC of the functional parameter spaces is small, the excess risk of this estimator relative to the oracle approximation to the CATE is small.  Note that the lower bound also bounds the distance of $S$ to the oracle (best-in-class) $s_\bullet$ predictor of the CATE.  Therefore, the bounds on the excess risk and  distance readily follow from the existing characterization of the ORC.  For example, if $\mathcal{H}$ has VC type covering entropy with VC index $d$, then the ORC is of order $d/|A|$.

Since the ``base" functions $B$'s play only a noise-reducing rule, we can always select $\mathcal{B}$ to be no more complex than $\mathcal{S}$.  For example, we can use $\mathcal{B} \subseteq  \mathcal{S}$ or pre-train $B$ using a separate auxiliary sample, in which case $\mathcal{B}$ is a singleton.\footnote{This also applies to cross-fitting, which is a better form of sample-splitting for practice.} In either case, learning the technical baseline function does not affect the rate of learning the oracle prediction $s_\bullet$; and the ORC is determined solely by the complexity of $\mathcal{S}$.   

Further,  we can use  losses \eqref{eq:A} and \eqref{eq:B} for choosing the best ML method. We provide detailed discussion in Appendix \ref{app:sec5} of the OA.

%% file: Section6.tex
\section{Application: Where are nudges for immunization the most effective? }\label{sec:empirics}

We apply our methods to an RCT in India that was conducted to improve immunization and provide detailed implementation algorithms. We first describe the setting and results, and then provide the implementation details. Our main specification reports median intervals (MI) from causal learners via Boosting as described in Algorithm \ref{alg2:implementation}. We also estimate results using predictive ML methods and report them in Appendix \ref{sec:predictive2} of the OA as they perform worse than the causal learners. Finally, inferential results are robust to using prediction intervals for the majority values (MP, reported in Appendix \ref{sec:mp_causal2} of the OA). For the sample splitting, we allocate 1/3 of the sample to the main sample.\footnote{We find similar results using 1/2 splits. These results are available upon request.}

\subsection{Setting}

Immunization is widely believed to be one of the most cost-effective ways to save children's lives. Much progress has been made in increasing immunization coverage since the 1990s. For example, according to the World Health Organization (WHO), global measles deaths have decreased by  73\% from  536,000 estimated deaths in 2000 to 142,000 in 2018. In the last few years, however, global vaccination coverage has remained stuck at around 85\% (until the COVID-19 epidemics, when they plummeted). In 2018, 19.7 million children under the age of one year did not receive basic vaccines. Around 60\% of these children lived in ten countries: Angola, Brazil, the Democratic Republic of the Congo, Ethiopia, India, Indonesia, Nigeria, Pakistan, the Philippines, and Vietnam. The WHO estimates that immunization saves 2-3 million deaths every year and that an additional 1.5 million deaths could be averted every year if global vaccination coverage improves  (this is comparable to 689,000 deaths from COVID-19 between January and August 2020).\footnote{See WHO ``10 facts on immunization'', \url{https://www.who.int/features/factfiles/immunization/facts/en/index1.html}}

While most of the early efforts have been devoted to building an immunization infrastructure and ensuring that immunization is available close to people's homes, there is a growing recognition that it is important to also address the demand for immunization. Part of the low demand reflects deep-seated mistrust, but in many cases, parents seem to be perfectly willing to immunize their children. For example, in our data for Haryana, India,  among the sample's older siblings who should all have completed their immunization course, 99\% had received polio drops, and about 90\% had an immunization card. 90\% of the parents claimed to believe immunization is beneficial, and 3\% claimed to believe it is harmful. However, only 37\% of the older children had completed the course and received the measles vaccine, according to their parents (which is likely to be an overestimate), and only 19.4\% had done so before the fifteen month of life, when it is supposed to be done between the 10th and the 12th month. It seems that parents lose steam over the course of the immunization sequence, and nudges could be helpful to boost demand. Indeed, recent literature cited in the introduction suggests that ``nudges,'' such as small incentives, leveraging the social network, SMS, etc., may have a large effect on those services. 

In 2017, Esther Duflo, one of the authors of this paper, led a team that conducted a large-scale experiment with the government of Haryana in North India to test various strategies to increase the takeup of immunization services. The government health system rolled out an e-health platform designed by a research team and programmed by an MIT group (SANA health), in which nurses collected data on which child was given which shot at each immunization camp. The platform was implemented in over 2,000 villages in seven districts and provides excellent administrative data on immunization coverage.\footnote{\cite{banerjee2019improving} discuss validation data from random checks conducted by independent surveyors.} From the individual data, we constructed the monthly sum of the number of children eligible for the program (i.e., age 12 months or younger at their first vaccines) who received each particular immunization at a program location. These children were aged between 0 and 15 months. This paper focuses on the number of children who received the measles shot, as it is the last vaccine in the sequence and thus a reliable marker for full immunization.  

\begin{table}[h]\caption{Selected Descriptive Statistics of Villages}\label{table:Haryana_desc}
\centering
\small{
\begin{tabular}{lccc}
  \hline\hline
 & All & Treated & Control \\ 
 \hline \\  [-4mm]
\textbf{Outcome Variables}  \emph{(Village-Month Level)} \\   [0.3mm]
Number of children who completed the immunization schedule  & 8.239 & 10.071 & 7.303 \\ [1mm]

\textbf{Baseline Covariates--Demographic Variables}  \emph{(Village Level)}   \\  [0.3mm]

Household financial status (on 1-10 scale) & 3.474 & 3.163 & 3.633 \\

Fraction Scheduled Caste-Scheduled Tribe (SC/ST) & 0.19 & 0.196 & 0.187 \\

Fraction Other Backward Caste (OBC) & 0.229 & 0.216 & 0.236 \\

Fraction Hindu & 0.912 & 0.854 & 0.942 \\

Fraction Muslim & 0.06 & 0.11 & 0.034  \\

Fraction Christian & 0.001 & 0.002 & 0 \\

Fraction Literate & 0.797 & 0.788 & 0.801  \\

Fraction Single & 0.052 & 0.051 & 0.053 \\

Fraction Married (living with spouse) & 0.516 & 0.499 & 0.525 \\

Fraction Married (not living with spouse) & 0.003 & 0.003 & 0.003 \\

Fraction Divorced or Separated & 0.002 & 0.006 & 0 \\

Fraction Widow or Widower & 0.039 & 0.036 & 0.041 \\

Fraction who received Nursery level educ. or less & 0.152 & 0.152 & 0.151 \\

Fraction who received Class 4 level educ. & 0.081 & 0.08 & 0.082 \\

Fraction who received Class 9 educ. & 0.157 & 0.162 & 0.154\\

Fraction who received Class 12 educ. & 0.246 & 0.223 & 0.257\\

Fraction who received Graduate or Other Diploma & 0.085 & 0.078 & 0.088\\

\textbf{Baseline Covariates--Immunization History of Older Cohort}  \emph{(Village Level)} \\  [0.3mm]

Number of vaccines administered to pregnant mother & 2.265 & 2.194 & 2.301  \\

Number of vaccines administered to child since birth & 4.464 & 4.373 & 4.51 \\

Fraction of children who received polio drops & 0.999 & 1 & 0.999 \\

Number of polio drops administered to child & 2.982 & 2.984 & 2.981 \\

Fraction of children who received an immunization card & 0.91 & 0.869 & 0.931 \\

Fraction of kids who received Measles vaccine by 15 months of age & 0.191 & 0.17 & 0.201 \\

Fraction of kids who received Measles vaccine at credible locations & 0.381 & 0.362 & 0.391 \\

\textbf{Number of Observations} \\  [0.3mm]
Villages & 103 & 25 & 78\\
Village-Months & 843 & 204 & 639 \\
   \hline\hline 
\end{tabular}}
\end{table}

Before the launch of the interventions, survey data were collected in 912 of those villages using a sample of 15 households with children aged 1-3 per village. The baseline data covers demographic and socio-economic variables and the immunization history of these children, who were too old to be included in the intervention. In these 912 villages, three different interventions (and their variants) were cross-randomized at the village level:  
\begin{enumerate}
\item  Small incentives for immunization: parents/caregivers receive mobile phone credit upon bringing children for vaccinations.
\item  Immunization ambassador intervention: information about immunization camps was diffused through key members of a social network.
\item  Reminders: a fraction of parents/caregivers who had come at least one time received SMS reminders for pending vaccinations of the children. 
\end{enumerate}
For each of these interventions, there were several possible variants: incentives were either low or high and either flat or increasing with each shot; the immunization ambassadors were either randomly selected or chosen to be information hubs, using the ``gossip'' methodology developed by \cite{banerjee2019leveraging}, a trusted person, or both; and reminders were sent to either 33\% or 66\% of the people concerned. Moreover, each intervention was cross-cut, generating 75 possible treatment combinations. 

\cite{banerjee2019leveraging} developed and implemented a two-step methodology to identify the most cost-effective and the most effective policy to increase the number of children completing the full course of immunization at the village level and estimate its effects (correcting for bias due to the fact that the policy is found to be the best). 
First, they used a specific version of LASSO to determine which policies are irrelevant and which policy variants can be pooled together. 
Second, they obtained consistent estimates of this restricted set of pooled policies using post-LASSO \citep{CHS:PnP}.
They found that the most cost-effective policy (and the only one to reduce the cost of each immunization compared to the status quo) is to combine information hub ambassadors (trusted or not) and
SMS reminders. But the policy that increases immunization the most is the combination of information-hub ambassador, the presence of reminders, and increasing incentives (regardless of levels).   This is also the most expensive package, so the government was interested in prioritizing villages: where should they scale up the full package? This is an excellent application of this methodology because there was no strong prior. 

\subsection{Results}

We compare 25 treated villages where this particular policy bundle was implemented with 78 control villages that received neither sloped incentives and social network intervention nor reminder. Our data constitute an approximately balanced monthly panel of the 103 treated and control villages for 12 months (the duration of the intervention). The outcome variable, $Y$,  is the number of children 15 months or younger in a given month in a given village who receive the measles shot. The treatment variable, $D$,  is an indicator of the household being in a village that receives the policy. The covariates, $Z$, include 36 baseline village-level characteristics such as religion, caste, financial status, marriage and family status, education, and baseline immunization. The propensity score is constant. 

\begin{table}[t] 
\centering
\small
\caption{Comparison of Causal ML Methods: Immunization Incentives}
\label{table:Haryana_Best}
\begin{tabular}{lcccc}
  \hline
    \hline
  \\ [-3mm]
 & Elastic Net & Boosting & Neural Network & Random Forest \\  [1mm]
  \hline \\ [-3mm]
Best BLP ($\Lambda$) &   67.750 & 32.900 & 53.420 & 25.200 \\ 
& [51.491, 82.368] & [23.246, 44.665] & [42.516, 67.647] & [18.328, 34.705] \\
Best GATES ($\bar \Lambda$)  &  8.254 &  5.104 &  6.001 &  4.492 \\ 
 & [7.329, 9.314] & [4.27, 6.079] & [5.087, 6.888] & [3.339, 5.507] \\ 
   \hline\hline
\end{tabular}
\caption*{\footnotesize Notes: Medians over 250 splits. Note that we used Neural Network Causal Boosting for all methods, using Algorithm \ref{alg2:implementation}. The brackets report interquartile ranges for goodness-of-fit statistics.}
\end{table}

Table \ref{table:Haryana_desc} shows sample averages in the control and treated groups for some of the variables used in the analysis weighted by village population, as the rest of the analysis. Treatment and control villages have similar baseline characteristics (in particular, the immunization status of the older cohort was similar). The combined treatment was very effective on average. During the course of the intervention, on average, 7.30 children per month aged 15 months or less got the measles shot that completes the immunization sequence in control villages, and 10.08 did so in treatment villages. This is a raw difference of 2.77 or 38\% of the control mean. Note that while these effects are not insignificant, we are far from reaching full immunization: The baseline survey suggests that about $38\%$ of children aged 1-3 had received the measles shot at baseline, and $19.4\%$ had received it
before they turned 15 months. These estimates imply that the fraction getting their measles shot before 15 months would only go up to $26.7\% (19.4+0.38*19.4).$

\begin{table}[t]
\caption{BLP of Immunization Incentives Using Causal Proxies}\label{table:Haryana_BLP}
\small
\begin{tabular}{lcccc}
  \hline
    \hline
  \\ [-3mm]
   & \multicolumn{2}{c}{Elastic Net}  & \multicolumn{2}{c}{Neural Network} \\  [1mm]
 & ATE ($\beta_{1}$) & HET ($\beta_{2}$) & ATE ($\beta_{1}$) & HET ($\beta_{2}$) \\ 
 \cline{2-5}  \\ [-3mm]
& 2.814 & 1.047 & 2.441 & 0.899 \\ 
& (1.087,4.506) & (0.826,1.262) & (0.846,3.979) & (0.685,1.107) \\  \relax 
& [0.004] & [0.000] & [0.004] & [0.000] \\    [1mm]
   \hline\hline
\end{tabular}
\caption*{\footnotesize Notes: Medians over 250 splits.  Median Confidence Intervals ($\alpha=.05$) in parenthesis. P-values for the hypothesis that the parameter is equal to zero against the two-sided alternative in brackets.}
\end{table}

The implementation details for the heterogeneity analysis follow Algorithm \ref{alg:implementation} below, with three characteristics due to the design: we weight village-level estimations by village population,  include district--time fixed effects, and cluster standard errors at the village level. Table \ref{table:Haryana_Best} compares the four ML methods for producing proxy predictors $S(Z_i)$ using the criteria in \eqref{define:lambda} and \eqref{define:lambda2}. We find that Elastic Net and Neural Network outperform the other methods, with Elastic Net beating Neural Network by a smaller margin than the other methods. Accordingly, we shall focus on these two methods for the rest of the analysis.

Table \ref{table:Haryana_BLP} presents the results of the BLP of CATE on the ML proxies. We report estimates of the coefficients $\beta_1$ and $\beta_2$, which correspond to the ATE and heterogeneity loading (HET) parameters in the BLP. The ATE estimates in Columns 1 and 3 indicate that the package treatment increases the number of immunized children by $2.81$ based on elastic net estimates and by $2.44$ based on neural network estimates. Reassuringly, these estimates are on either side of the raw difference in means ($2.77$). Focusing on the HET estimates, we find strong heterogeneity in treatment effects, as indicated by the statistically significant estimates. Moreover, the estimates are close to 1, suggesting that the ML proxies are good predictors of the CATE.

\begin{figure}[h] 
\caption{GATES of Immunization Incentives}
\vspace{0.5cm}
\includegraphics[scale=0.9]{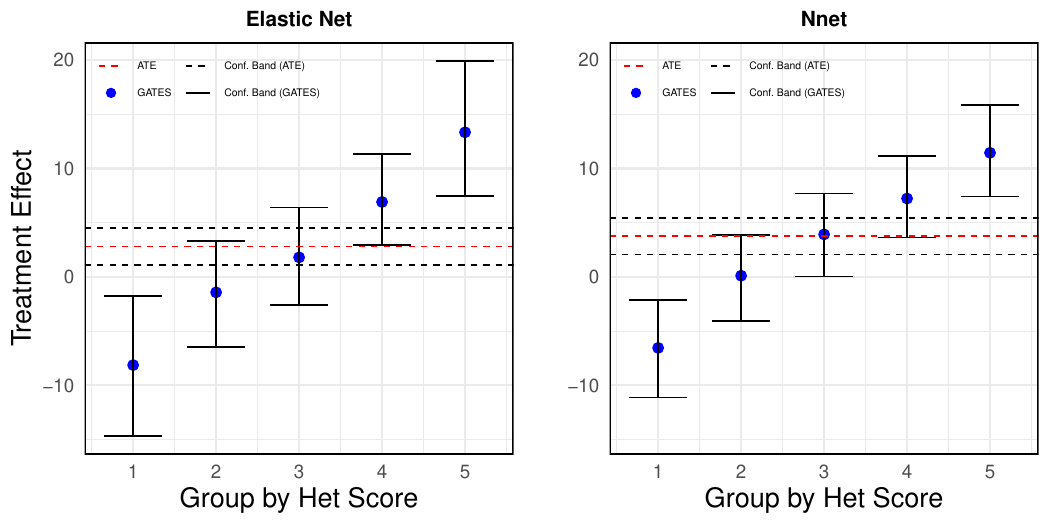}
\caption*{\footnotesize Notes: GATES of Immunization Incentives, based upon Causal Learners. Median point estimates and  Median confidence interval $(\alpha =.05)$ in parenthesis, over 250 splits.}
\label{figure:Haryana_GATES}
\end{figure}

\begin{table}[t]
\vspace{1.5cm}
\centering
\caption{GATES of  20\% Most and Least Affected Groups}\label{table:Haryana_GATES}
\footnotesize
\begin{tabular}{lcccccc}
  \hline
    \hline
  \\ [-3mm]
       & \multicolumn{3}{c}{Elastic Net}  & \multicolumn{3}{c}{Nnet} \\   [1mm]
& 20\% Most  &  20\% Least  & Difference & 20\% Most  & 20\% Least  & Difference  \\  [-1mm]
 & ($G_5$) & ($G_1$)  &  & ($G_5$) & ($G_1$)  &    \\
 \cline{2-7}  \\ [-2mm]
GATE  & 13.230 & -8.000 & 21.60 & 11.210 & -6.551 & 18.13 \\ 
$\gamma_k := $  $ \hat{\Ep} [ s_0(Z) \mid G_k  ]$   & (8.219,18.67) & (-13.41,-2.574) & (13.70,29.74) & (7.721,14.47) & (-10.37,-2.786) & (12.84,23.52) \\  
   & [0.000] & [0.009] & [0.000] & [0.000] & [0.002] & [0.000] \\   [1mm]
Control Mean & 2.431 & 12.580 & -9.916 & 1.058 &  9.576 & -8.544   \\
:= $\hat{\Ep} [ b_0(Z) \mid G_k  ]$  & (1.611,3.216) & (11.71,13.50) & (-11.17,-8.679) & (0.340,1.750) & (8.824,10.31) & (-9.542,-7.589)  \\ 
 & [0.000] & [0.000] & [0.000] & [0.009] & [0.000] & [0.000] \\  [1mm]
   \hline\hline
\end{tabular}
\caption*{\footnotesize Notes: Medians over 250 splits. Median confidence interval $(\alpha =.05)$ in parenthesis. P-values for the hypothesis that the parameter is equal to zero against the two-sided alternative in brackets.}
\end{table}

Next, we estimate the GATES by quintiles of the ML proxies. Figure \ref{figure:Haryana_GATES} presents the estimated GATES coefficients $\gamma_{1}-\gamma_{5}$ along with joint confidence bands and the ATE estimates. In Table \ref{table:Haryana_GATES} we present the result from the hypothesis test that the difference of the ATE for the most and least affected groups is statistically significant. We find that this difference is $21.60$ and $18.13$  based on elastic net and neural network methods, respectively, and is statistically significant. Given that the ATE estimates in the whole population are about $2.5$, these results suggest a large and potentially policy-relevant heterogeneity.

The analysis so far reveals very large heterogeneity, with two striking results. First, the results are very large for the most affected villages. In these villages, an average of $13.23$ extra children eligible for baseline incentives get the measles vaccines every month (starting from a mean of $2.19$ in the elastic net estimation). Second, the impact is \emph{negative} and significant in the least affected villages (an average decline of $8.00$ immunization per month, starting from $12.68$ in the elastic net estimation). It looks like in some contexts, the combined package of small incentives, reminders, and persuasion by members of the social network put people off immunization. 

\begin{table}[t]
\centering
\caption{CLAN of Immunization Incentives }\label{table:Haryana_CLAN_C1}
\fontsize{8}{8}\selectfont
\begin{tabular}{lcccccc}
  \hline
  \hline
  \\ [-1mm]
       & \multicolumn{3}{c}{Elastic Net}  & \multicolumn{3}{c}{Nnet} \\   [1mm]
& 20\% Most  & 20\% Least  & Difference & 20\% Most  & 20\% Least  & Difference\\ [0.5mm] 
 & ($\delta_{5}$) &  ($\delta_1$)  & ($\delta_{5}-\delta_{1}$) &   ($\delta_{5}$) &  ($\delta_1$)  & ($\delta_{5}-\delta_{1}$) \\ [1mm] 
 \cline{2-7}  \\ [-1mm]
  Number of vaccines  & 2.187 & 2.277 & -0.081 & 2.174 & 2.285 & -0.112 \\ 
     to pregnant mother & (2.115,2.259) & (2.212,2.342) & (-0.180,0.015) & (2.111,2.234) & (2.224,2.345) & (-0.202,-0.028) \\ 
   & - & - & [0.190] & - & - & [0.019] \\ 
  Number of vaccines  & 4.077 & 4.639 & -0.562 & 4.264 & 4.734 & -0.490 \\ 
   to child since birth   & (3.858,4.304) & (4.444,4.859) & (-0.863,-0.260) & (4.091,4.434) & (4.549,4.900) & (-0.739,-0.250) \\ 
   & - & - & [0.001] & - & - & [0.000] \\ 
  Fraction of children & 0.998 & 1.000 & -0.002 & 1.000 & 1.000 &  0.000 \\ 
    received polio drops  & (0.995,1.001) & (0.997,1.003) & (-0.006,0.002) & (1.000,1.000) & (1.000,1.000) & (0.000,0.000) \\ 
   & - & - & [0.683] & - & - & [0.943] \\ 
  Number of polio & 2.955 & 2.993 & -0.037 & 2.965 & 2.998 & -0.032 \\ 
     drops to child  & (2.935,2.974) & (2.976,3.010) & (-0.063,-0.010) & (2.953,2.977) & (2.985,3.010) & (-0.049,-0.016) \\ 
   & - & - & [0.013] & - & - & [0.000] \\ 
Fraction of children  & 0.803 & 0.926 & -0.121 & 0.908 & 0.927 & -0.027 \\ 
    received immunization card & (0.754,0.851) & (0.882,0.969) & (-0.187,-0.054) & (0.881,0.932) & (0.898,0.959) & (-0.059,0.007) \\ 
   & - & - & [0.001] & - & - & [0.217] \\ 
  Fraction of children received  & 0.133 & 0.243 & -0.106 & 0.126 & 0.260 & -0.131 \\ 
     Measles vaccine   & (0.097,0.169) & (0.209,0.276) & (-0.153,-0.056) & (0.095,0.159) & (0.228,0.291) & (-0.176,-0.085) \\ 
  by 15 months of age  & - & - & [0.000] & - & - & [0.000] \\ 
  \\
   Fraction of children received & 0.293 & 0.399 & -0.110 & 0.289 & 0.433 & -0.142 \\ 
   Measles vaccine & (0.246,0.338) & (0.358,0.444) & (-0.174,-0.045) & (0.246,0.331) & (0.391,0.475) & (-0.206,-0.084) \\ 
   at credible locations & - & - & [0.002] & - & - & [0.000] \\ 
   \hline \hline
\end{tabular}
\caption*{\footnotesize Notes: Medians over 250 splits.   Median confidence interval $(\alpha =.05)$ in parenthesis.  P-values for the hypothesis that the parameter is equal to zero against the two-sided alternatives in brackets.}
\end{table}

Given these large differences, it is important to determine whether this heterogeneity seems to be associated with pre-existing characteristics. To answer this question, we ask what variables are associated with the heterogeneity detected in BLP and GATES via  CLAN. Table \ref{table:Haryana_CLAN_C1} reports the CLAN estimates for a selected set of covariates and Tables \ref{table:Haryana_CLAN1}--\ref{table:Haryana_CLAN2} in Appendix \ref{app:empirical} of the OA for the rest of covariates.  Regardless of the method used, the estimated differences in the means of most and least affected groups for the number of vaccines to child since birth, number of polio drops to child, the fraction of children receiving measles vaccines by 15 months of age, and fraction of children receiving measles vaccine at credible locations, are negative and statistically significant. Those are various measures of pretreatment immunization levels, all survey-based, that have nothing to do with our measure of impact. These results suggest that the villages with low levels of pretreatment immunization are the most affected by the incentives. These are, in fact, the only variables that consistently pop up from the CLAN. Thus, in this instance, the policy preferred ex-ante by the government (since it is equality-enhancing) also happens to be the most effective. 


While the heterogeneity associated with the baseline immunization rates cannot be causally interpreted (it could always be proxying for other things), it still sheds light on the negative effect we find for the least affected group. Note that this effect is \emph{not} mechanical. Even in the least affected villages, there was a good number of children who did not receive the measles shot, and they were not close to reaching full immunization, where they could not have experienced an increase. It may be that it would have been difficult to vaccinate $13.23$ extra children every month, but there was scope to experience an increase in immunization. We had no prior on that the effect would be larger in the villages with the lowest immunization rate. On the contrary, immunization rates could have been low precisely because parents were more doubtful about immunization. For example, immunization is particularly low in Muslim-majority villages, which is believed to reflect their lack of trust in the health system. There were, therefore, reasons to be genuinely uncertain about where immunization would have had the largest effect. 

One possible interpretation of the negative impact in some villages is that villagers were intrinsically motivated to get immunized. The nudging with small incentives and mild social pressure may have backfired, crowding out intrinsic motivation without providing a strong enough extrinsic motivation to act as in \cite{gneezy2000fine}. A point estimate of $13.23$ extra immunization per month in the most affected group might seem high: a multiplication by $6.0$ of the baseline level  (based on the elastic net specification). This increase in immunizations is not inconsistent with the literature: in a set of villages with a very low immunization rate in Rajasthan, \cite{banerjee2010improving} find that small incentives increase immunization from $18\%$  to $39\%$ (relative to a treatment that just improves infrastructures, and $6\%$ relative to the control group) in a low immunization region (in the entire sample, not in the places where it is most effective), which was also a very large increase. Given the restrictions imposed on the data set (only children 1 year or less at their first immunization were included), the data cover children who were 15 months or younger when getting the measles shot. Among the older cohort in the most affected group, only  $12.7\%$ of children were vaccinated before 15 months. Taking this as a benchmark for the control group, the estimate would still imply that only $64\%$ of the treatment group was immunized before 15 months: a big improvement but not implausible. 

 \begin{table}[t]
\centering
\caption{Cost Effectiveness in GATE quintiles}\label{table:Haryana_CE}
\scalebox{0.75}{
\begin{tabular}{lllllll}
  \hline
& \multicolumn{3}{c}{Elastic Net}   & \multicolumn{3}{c}{Nnet} \\
 & Mean in Treatment  & Mean in Control  & Difference & Mean in Treatment  & Mean in Control & Difference\\ [0.5mm] 
 & ($\hat{\mathrm{E}} [X \mid D =1, G_k  ]) $ & ($\hat{\mathrm{E}} [X \mid D = 0, G_k  ]$) &  & ($\hat{\mathrm{E}} [X \mid D =1, G_k  ]) $ & ($\hat{\mathrm{E}} [X \mid D = 0, G_k  ]$) &\\    \hline
  \hline
  Imm. per dollar ($G_1$) & 0.033 & 0.047 & -0.013 & 0.033 & 0.047 & -0.015 \\ 
   & (0.030,0.037) & (0.045,0.048) & (-0.017,-0.009) & (0.029,0.036) & (0.045,0.049) & (-0.019,-0.010) \\ 
   & - & - & [0.000] & - & - & [0.000] \\ 
  Imm. per dollar ($G_2$)& 0.032 & 0.044 & -0.012 & 0.035 & 0.044 & -0.009 \\ 
   & (0.027,0.037) & (0.042,0.046) & (-0.017,-0.007) & (0.031,0.039) & (0.042,0.046) & (-0.014,-0.005) \\ 
   & - & - & [0.000] & - & - & [0.000] \\ 
  Imm. per dollar ($G_3$) & 0.037 & 0.043 & -0.006 & 0.038 & 0.043 & -0.005 \\ 
   & (0.032,0.041) & (0.040,0.045) & (-0.011,-0.002) & (0.034,0.041) & (0.040,0.045) & (-0.009,-0.001) \\ 
   & - & - & [0.016] & - & - & [0.028] \\ 
  Imm. per dollar ($G_4$)  & 0.039 & 0.039 &  0.000 & 0.038 & 0.041 & -0.003 \\ 
   & (0.036,0.042) & (0.036,0.042) & (-0.005,0.004) & (0.034,0.041) & (0.038,0.044) & (-0.008,0.001) \\ 
   & - & - & [1.000] & - & - & [0.253] \\ 
  Imm. per dollar ($G_5$) & 0.036 & 0.035 & 0.002 & 0.037 & 0.034 & 0.002 \\ 
   & (0.033,0.040) & (0.030,0.039) & (-0.004,0.007) & (0.033,0.040) & (0.031,0.038) & (-0.004,0.007) \\ 
   & - & - & [1.000] & - & - & [1.000] \\ 
   \hline
\hline
\end{tabular}}
\caption*{\footnotesize Notes: Medians over 250 splits.   Median confidence interval $(\alpha =.05)$ in parenthesis.  P-values for the hypothesis that the parameter is equal to zero against two-sided alternative in brackets.}
\end{table}

Our last exercise is to compute the cost-effectiveness of the program in various groups. To do so, we compute in each village the average number of immunizations delivered per dollar spent in a month in each group. The dollar spent is the fixed cost to run an immunization program per month (nurse salaries, administrative overheads, etc.) plus the marginal cost of each vaccine multiplied by the number of vaccines administered (incentives distributed to local health workers, vaccines doses, syringes, etc.) in both treatment and control villages, plus the extra cost of running each particular treatment (the cost of the tablets used for recording in all the treatment villages, the cost of contacting and enrolling the ambassadors, and the cost of the incentives).  
We then estimate the cost-effectiveness in each GATES group as  $\Ep[X(1) - X(0) \mid G_k]$, where $X$ is the immunizations per dollar. $\Ep[X(1) - X(0) \mid G_k] = \Ep[X \mid D = 1 , G_k] - \Ep[X \mid D = 0, G_k]$ by the randomization assumption, and we can estimate each of $\Ep[X \mid D = 1 , G_k]$ and $\Ep[X \mid D = 0 , G_k]$ analogously to CLAN, that is by taking sample averages within treatment groups for each sample split and aggregating over sample splits. 

The results are presented in Table \ref{table:Haryana_CE}. They highlight the crucial importance of treatment effect heterogeneity for policy decisions in this context. Overall, as shown in \cite{banerjee2019leveraging} the treatment is not cost-effective compared to the control (the immunization per dollar spent goes down). This analysis reveals that this is driven (not surprisingly) by negative impacts on cost-effectiveness in the groups where it is least effective. However, in the fourth and fifth quintile of cost-effectiveness, we cannot reject that the immunization per dollar spent is the same in the control group and in the treatment group, despite the added marginal cost of the incentives and the vaccines: this is because the fixed cost of running the program is now spread over a larger number of immunizations.

We performed the following additional analysis. First, since the main result in \cite{banerjee2019leveraging} is that the most cost-effective option on average is the combination of SMS plus Information hubs, an alternative policy question may therefore be whether there are places where it may be more cost-effective to add the incentives to this cheaper treatment. We replicated the heterogeneity analysis comparing these two treatments  and looked at the cost-effectiveness of GATES in these two options. There is also considerable heterogeneity in this comparison (see Figure \ref{figure:Haryana_GATESalt} of the OA). The results for cost-effectiveness are shown in Table \ref{table:Haryana_CEalt} of the OA. There, again, we find that in the two quintiles where adding incentives is most effective, it would be cost-effective, even compared to an alternative status quo of just having SMS and information hubs.

Second, we reproduce the analysis with the first vaccine in the series to be given at birth (Penta 1). The results are very similar to the results for the measles vaccine, with regions with the lowest immunization experiencing the largest effects. Third, still for the penta 1 vaccination, we apply the method to the comparison of villages with or without increasing incentives for vaccination. Once again, we find significant heterogeneity with the most affected regions being those with low immunization to start with (these results are available upon request).

\subsection{Implementation Details} \label{sec:details}
We describe two general algorithms  and provide some specific implementation details for the empirical example.

\begin{algorithm}[\textbf{Inference Algorithm}]\label{alg:implementation} 
The inputs are given by the data $\{(Y_i, D_i, Z_i, p(Z_i)\}$ on units $i \in [N]= \{1,..., N\}$. Fix the number of splits $N_S$ and the significance level $\alpha$, e.g. $N_S=250$ and $\alpha=0.05$. Fix a set of ML or Causal ML methods.

1. Generate $N_S$ random splits of $[N]$  into the main sample, $M$,  and the auxiliary sample, $A$. Over each split apply the following steps:
\begin{enumerate}[label=\textsf{\alph*.}] 
\item Using $A$, train each ML method and output predictions $B$ and $S$ for $M$.

\item Optionally, choose the best or aggregate ML methods using the results  of Section \ref{sec:further}. 

\item Estimate the BLP parameters via WR BLP \eqref{eq:BLP estimate} or HT BLP \eqref{eq:BLP estimate2} in $M$.  
\item Estimate the GATES parameters by WR GATEs \eqref{equation: estimate GATES} or HT GATEs \eqref{equation: estimate GATES 2} in $M$.

\item Estimate the CLAN parameters by taking averages (\ref{eq: CLAN estimate}) in $M$.

\item Compute the goodness of fit measures via (\ref{eq: GOOF estimate}) in $M$.

\end{enumerate}

2. If the winning ML methods were not chosen in Step 1b,  median-aggregate the goodness-of-fit measures and choose the best ML methods.

3. Compute and report the quantile-aggregated point estimates, p-values, and confidence intervals of Section 4. If Step 2 is used, compute and report the union of these statistics for all winners. 


\end{algorithm}

\begin{remark}(Choices) We choose $N_S$ sufficiently large to get enough representative values of the estimates of the target parameter values. In our experience, 250 splits are more than sufficient to obtain stable results in the sense that the point and interval estimates are not sensitive to increasing the number of splits in the empirical application.  We followed a version of the algorithm with Step 2 and without Step 1a. Note that it is also possible to choose the best methods using a hold out sample using either the loss functions of Section 5 or the goodness-of-fit measures of Section 3.
More research is needed to determine better practice for choosing the best ML methods.
\end{remark}

We implemented our causal learners via a  boosting approach that looks for relatively simple deviations from the initial predictive learner to improve CATE predictions.  The reason is that this approach performed better in our simulation experiments than directly solving the objective functions (A) and (B) over large parameter spaces.  We also observed that this boosting implementation performed better in the empirical example.  The performance improvement occurs because the objective functions (A) or (B)  tend to be much noisier than the objective functions in predictive learning and harder to tune.  The following algorithm summarizes the implementation.

\begin{algorithm}[\textbf{Causal Learner via Boosting}]\label{alg2:implementation} 
The inputs are given by data $\{Y_i, D_i, Z_i, p(Z_i)\}$ on units $i \in A \subset \{1,..., N\}$. Fix a predictive learner. Fix  deviation (or boosting) parameter spaces: $\mathcal{B}_\Delta$  and $\mathcal{S}_\Delta$ that contain functions $z \mapsto b_\Delta(z)$ and $z \mapsto s_\Delta(z)$, mapping the support of $Z$ to the real line.

\begin{itemize}
\item[1.] Train the predictive learner on the input, and output the base proxy function $z \mapsto B_A(z)$ and  CATE proxy function $z \mapsto S_A(z)$.
\item[2.]  Solve the objective function (A) or (B) for the parameter sets
 $\mathcal{B}= \{ B_A + b_{\Delta}, b_\Delta \in \mathcal{B}_\Delta\}$ and $\mathcal{S} = \{S_A + s_\Delta, s_\Delta \in \mathcal{S}_\Delta\},$ and  update $z \mapsto S_A(z)$ and $z \mapsto B_A(z)$ to be the solution.
 \item[3.] Optionally,  iterate on step 2 a few times.
 \end{itemize}
\end{algorithm}

Ideally, step 1 of the algorithm should make use of cross-fitting, but we present a simplified version for clarity.

\begin{remark}[Choices for Computational Experiments]  We use simple deviation parameter spaces with low complexity (small ORC). For example, in the computational experiments reported in Figures \ref{fig:targeting} and \ref{fig:targeting2}, we made the following choices: for Causal Neural Network (NN) Learner, we used shallow, regularized  NN as deviation spaces when the predictive learner was NN; for Forest Causal Learner, we used the shallow forest as the deviation space when the predictive learner was Random Forest;  for Causal OLS Learner, we used linear deviation space, when the predictive learner was OLS. Finally, when we used Causal Forest as the predictive learner, we used the shallow forest as the deviation space to obtain the Forest Causal Learner.\footnote{As this may sound confusing, we note that that Forest Causal Learners (FCL) differ from Causal Random Forests (CRF) in stratified experiments as FCL are based on weighted residualization whereas CRF are based on unweighted residualization of \cite{nie:20}. Therefore we use a slightly different name ``Forest Causal  Learner" rather than ``Causal Forest" to distinguish the proposal.}  We have also experimented with hybrid versions, for example, using one type of predictive learner and a different type of booster (for example, RF plus NN as a causal boost or Elastic Net plus NN as a casual boost).
\end{remark}

\begin{remark}[Choices for Empirical Example]  We used simple, regularized neural networks as deviation spaces in all results.  The number of neurons was kept less than or equal to 10 and was chosen based on cross-validating the objective function of type (A) over auxiliary data subsamples.  We used this choice regardless of the predictive ML as the starter.  Causal Learners constructed in this way improved the performance of each predictive ML method, raising the goodness of fit metrics by 5-10\% in relative terms.  However, they did not have any qualitative impact on empirical results (we report the results for predictive learners in Appendix \ref{sec:predictive2} of the OA).
\end{remark}

\begin{remark}[Predictive ML Methods] We considered four ML methods to estimate the proxy predictors: elastic net, boosted trees, neural network with feature extraction, and random forest. The ML methods are implemented in R using the package caret \citep{kuhn2008caret}.  The names of the elastic net, boosted tree, neural network with feature extraction, and random forest methods in caret  are glmnet, gbm, pcaNNet and rf, respectively. For each split of the data, we choose the tuning parameters separately for $B(z)$ and $S(z)$ based on mean squared error estimates of repeated 2-fold cross-validation, except for random forest, for which we use the default tuning parameters to reduce the computational time.\footnote{We have the following tuning parameters for each method: Elastic Net:  alpha (Mixing Percentage), lambda (Regularization Parameter), Boosted trees: n.trees (Number of  Boosting Iterations), interaction.depth (Max Tree Depth), shrinkage (Shrinkage), n.minobsinnode (Min. Terminal Node Size), size (Number of Hidden Units) , decay (Weight Decay),  mtry (Number of Randomly Selected Predictors).}  In tuning and training the ML methods we use only the auxiliary sample. In all the methods we rescale the outcomes and covariates to be between 0 and 1 before training. 
\end{remark}

%% file: ConclusionExtensions.tex
\section{Conclusion and Extensions}\label{sec:conclusion}
We propose to focus inference on key features of heterogeneous effects in randomized experiments, and develop the corresponding methods. These key features include best linear predictors of the effects and average effects sorted by groups, as well as average characteristics of most and least affected units.   Our  approach is valid in high dimensional settings, where the effects are estimated by machine learning methods. The main advantage of our approach is its agnostic nature; it avoids making strong assumptions. Estimation and inference relies on data splitting, where the latter allows us to avoid overfitting and all kinds of non-regularities.  Our inference aggregates the results across many splits, reducing the replication risks, and could be of independent interest. An empirical application illustrates the practical use of the approach. 

Our hope is  that applied researchers use the method to discover whether there is heterogeneity in their data in a disciplined way. 
A researcher might be concerned about the application of our method  due to the possible power loss induced by  sample splitting.  This power loss is the price to pay when the researcher is not certain or willing to fully specify the form of the heterogeneity prior to conducting the experiment. Thus, if the researcher has a well-defined pre-analysis plan that spells out a small number of heterogeneity groups in advance, then there is no need of splitting the sample.\footnote{More generally, the plan needs to specify a parametric form for the heterogeneity as a low dimensional function of pre-specified covariates \citep[e.g.,][]{CFL2014}. In this case, ML tools can still be used to efficiently estimate the CATEs in the presence of control variables but are not required to detect heterogeneity \citep{bcfh17, DML}.}  However, this situation is not common. In general, the researchers might not be able to fully specify the form of the heterogeneity  due to lack of information, economic theory,  or willingness to take a stand at the early stages of the analysis. They might also face data limitations that preclude the availability of the desired  covariates.  Here we recommend the use of our method to avoid overfitting and p-hacking, and impose discipline to the heterogeneity  analysis at the cost of some power loss due to sample splitting.\footnote{ It is not clear whether this loss is real though, as we are not aware of any alternative method that avoids sample splitting and that works at the same  level of agnosticism as ours. In a previous version of the paper we provided a numerical example using a simple parametric model where standard methods without sample splitting are available. We find that the extent of the power loss for not using the parametric form of the heterogeneity roughly corresponds to reducing the sample size by half in a test for the presence of heterogeneity, although the exact comparison depends on features of the data generating process.} If discovering and exploiting heterogeneity in treatment effect is a key goal of the research, the researcher should indeed plan for larger sample sizes (relative to just testing whether the treatment has an effect), but the required sample size remains within the realm of what is feasible in the field. In many applications we are aware of, there was apparent heterogeneity according to some covariates of interest, but the disciplined ML heterogeneity exercise found no systematic difference. This could be because this heterogeneity was a fluke, or because the method does not have the power to detect it in a small sample. In any case,  what this experience suggests is that one should not rely on ex-post heterogenous effects in such cases. 

The application to immunization in India is of substantive interest. 
Our findings suggest that a combination of small incentives, relay by information hub, and SMS reminders can have very large effect on vaccine take up in some villages where immunization was low at baseline, and even be as cost-effective than the status quo, but can also backfire in other places. This suggests that these types of strategy need to be piloted in the relevant context before being rolled out, and that heterogeneity needs to be taken into account.

 Our inference approach generalizes to any problem of the following sort, studied in \cite{semenova2020} using more conventional inference approaches. Suppose we can construct an \textit{unbiased signal} $\tilde Y$ such that $$\Ep [\tilde Y \mid Z]  = s_0(Z),$$ where $s_0(Z)$ is now a generic target function. Let $S(Z)$ denote an ML proxy for $s_0(Z)$. This setting generalizes the CATE inference problem of this paper.\footnote{The unbiased signals arise from multiplying an outcome with a Riesz representer for the effect of interest.  In CATE setting, the representer is the Horwitz-Thomposon transform.} We can then apply our inferential framework to this setting.  The setting covers a variety of causal effects of interest, for example, inference on conditional average causal derivatives, conditional average effects from transporting and distributional shifts of covariates, as we further explain in Appendix \ref{app:sec7} of the OA.


%% file: Appendix-Proofs.tex
\section{
Deferred Discussion and Proofs for Section \ref{sec:id}}\label{app:sec3}

\begin{remark}[Motonicity Restrictions on GATES] Suppose we observe $s_0(Z)$. In this case we can define the ideal GATES as:
$$
\gamma_{0 k}:=\mathrm{E}\left[s_0(Z) \mid G_{0 k}\right], \quad k=1, \ldots, K,
$$
where $G_{0 k}:=\left\{s_0(Z) \in I_{0 k}\right\}$, with $I_{0 k}=\left[\ell_{0, k-1}, \ell_{0, k}\right)$ and $-\infty=\ell_0<\ell_1<\ldots<\ell_K=+\infty$. By construction the ideal GATES obey the monotonicity restriction:
$$
\gamma_{01} \leqslant \ldots \leqslant \gamma_{0 K} .
$$
If $S(Z)$ provides a good approximation to $s_0(Z)$, it is reasonable to expect that the GATES also obey the monotonicity restriction: $\gamma_1 \leqslant \ldots \leqslant \gamma_K$, but there is no guarantee. However, we can always replace $\gamma=\left\{\gamma_k\right\}_{k=1}^K$ by the non-decreasing rearrangement (sorted vector) $\gamma^*=\left\{\gamma_k^*\right\}_{k=1}^K$, such that $\gamma^*$ obeys the monotonicity condition $\gamma_1^* \leqslant \ldots \leqslant \gamma_K^*$. The benefit is that $\gamma^*$ is always closer to $\gamma_0=\left\{\gamma_{0 k}\right\}_{k=1}^K$ than $\gamma$ in the sense that
$$
\left\|\gamma^*-\gamma_0\right\|_{\infty} \leqslant\left\|\gamma-\gamma_0\right\|_{\infty},
$$
where $\|\cdot\|_{\infty}$ is the sup-norm. This follows from the contraction property of the rearrangement (e.g., Chernozhukov et al., 2009). Therefore, we can always use sorting to target the ideal GATES better. Similarly, when performing estimation, we can replace $\hat{\gamma}=\left\{\widehat{\gamma}_k\right\}_{k=1}^K$ by their non-decreasing rearrangement (sorted vector) $\widehat{\gamma}^*=\left\{\widehat{\gamma}_k^*\right\}_{k=1}^K$, which results in an estimator with lower estimation error in the sense that surely:
$$
\left\|\widehat{\gamma}^*-\gamma_0\right\|_{\infty} \leqslant\left\|\hat{\gamma}-\gamma_0\right\|_{\infty} .
$$
\end{remark}

\subsection*{Proof of Theorem \ref{theorem: BLP1}} The subset of  the normal equations, which correspond to  $\alpha := (\alpha_1,\alpha_2)'$, are  $
\Ep [w(Z) ( Y- \alpha_0' X_1  - \alpha' X_2)  X_2] =0.
$
Substituting  $Y = b_0(Z) + s_0(Z) D + U$, and using the definition $X_2 = X_2(Z,D) = [ D-p(Z), (D-p(Z) (S - \Ep_{} S)]'$,  $X_1 = X_1(Z)$,  and the law of iterated expectations, we notice that:
$$
\begin{array}{l}
 \Ep [w(Z) b_0(Z) X_2 ] = \Ep [w(Z) b_0(Z) \underbracket{\Ep[X_2(Z,D)\mid Z]}_{=0} ]  = 0, \\
\Ep [w(Z) U X_2] = \Ep [ w(Z) \underbracket{\Ep [U\mid Z,D]}_{0} X_2 (Z,D)] = 0, \\
\Ep [w(Z) X_1 X_2] = \Ep [w(Z) X_1 (Z) \underbracket{\Ep[X_2(Z,D) \mid Z]}_{=0} ]  = 0. \\
\end{array}
$$
Hence the normal equations simplify to:
$\Ep [w(Z) ( s_0(Z) D -  \alpha' X_2)   X_2] =0.$
Since
$$
\Ep [\{D - p(Z)\} \{D - p(Z)\} \mid Z]  = p(Z) (1- p(Z)) = w^{-1} (Z),
$$
and $S = S(Z)$, the components of $X_2$ are orthogonal by the law of iterated expectations:
$$\Ep [w(Z) (D - p(Z)) (D- p(Z)) (S - \Ep_{} S)] = \Ep (S - \Ep S) = 0. $$  Hence the 
normal equations above further simplify to  $$
\begin{array}{l}
\Ep [w(Z) \{s_0(Z) D -  \alpha_1 (D - p(Z))\}    (D - p(Z))  ] =0, \\
\Ep [w(Z) \{ s_0(Z) D -  \alpha_2 (D- p(Z)) (S - \Ep_{} S)   \}   (D- p(Z)) (S - \Ep_{} S) ] =0.
\end{array}
$$
Solving these equations and using the law of iterated expectations, we obtain
\begin{eqnarray*}
\alpha_1 &=&   \frac{\Ep [w(Z) \{ s_0(Z) D (D - p(Z))] \}}{\Ep [w(Z) (D - p(Z))^2]} = \frac{ \Ep [w(Z) s_0(Z) w^{-1}(Z)] }{\Ep [w(Z) w^{-1}(Z)]} =  \Ep s_0(Z),  \\
\alpha_2 &=&   \frac{\Ep [w(Z) \{ s_0(Z) D (D - p(Z)) (S - \Ep S) \}]}{\Ep [w(Z) (D - p(Z))^2 (S- \Ep S)^2]}\\
&=& \frac{ \Ep [w(Z) s_0(Z) w^{-1}(Z) (S - \Ep S) ]}{\Ep[ w(Z) w^{-1}(Z) (S- \Ep S)^2]}  = \frac{\mathrm{Cov} (s_0(Z), S)}{\mathrm{Var}(S)}.
\end{eqnarray*}
The conclusion follows by noting that  these coefficients also solve the normal equations
$$
\Ep\{[ s_0(Z) - \alpha_1 - \alpha_2(S - \Ep S)][1,  (S - \Ep S)]'\} =0,
$$
which characterize the optimum in the problem of best linear approximation/prediction of $s_0(Z)$ using $S$.  \qed

\subsection*{Proof of Theorem \ref{theorem: BLP2}} The subset of  the normal equations, which correspond to $\mu := (\mu_1, \mu_2)'$, are  $
\Ep [ ( Y H - \mu_0' X_1 H  - \mu' \tilde X_2)  \tilde X_2] =0.
$
Substituting  $Y = b_0(Z) + s_0(Z) D + U$, and using the definition $\tilde X_2 = \tilde X_2(Z) = [1,  (S(Z) - \Ep_{} S(Z))]'$, $X_1 = X_1(Z)$,  and the law of iterated expectations, we notice that:
$$
\begin{array}{l}
 \Ep [ b_0(Z) H \tilde X_2(Z) ] = \Ep [ b_0(Z) \underbracket{\Ep[H(D,Z) \mid Z]}_{=0}  \tilde X_2(Z) ]  = 0, \\
\Ep [ U H \tilde X_2(Z) ] = \Ep [  \underbracket{\Ep [U \mid Z,D]}_{0}  H(D,Z) \tilde X_2(Z) ] = 0, \\
\Ep [ X_1(Z) H \tilde X_2(Z)] = \Ep [ X_1 (Z) \underbracket{\Ep[H(D,Z) \mid Z]}_{=0} \tilde X_2 (Z) ]  = 0.
\end{array}
$$
Hence the normal equations simplify to:
$\Ep  [ (s_0(Z) D H -  \mu' \tilde X_2)   \tilde X_2] =0.$
Since  $1$ and $S- \Ep S$ are orthogonal, the 
normal equations above further simplify to  $$
\begin{array}{l}
\Ep  \{s_0(Z) D H -  \mu_1 \}  =0, \ \ 
\Ep  [\{  s_0(Z) D  H -  \mu_2  (S - \Ep_{} S)   \}  (S - \Ep_{} S) ] =0.
\end{array}
$$
Using that $
\Ep [D H \mid Z] = [p(Z) (1- p(Z))]/ [ p(Z) (1- p(Z))] =1,
$
$S=S(Z)$, and the law of iterated expectations, the equations simplify to
$$\begin{array}{l}
\Ep  \{s_0(Z) -  \mu_1 \}  =0, \ \
\Ep  [\{ s_0(Z) -  \mu_2  (S - \Ep_{} S)   \}  (S - \Ep_{} S)]  =0.
\end{array}$$
These are normal equations that characterize the optimum in the problem of best linear approximation/prediction of $s_0(Z)$ using $S$.   Solving these equations gives the expressions for $\beta_1$ and $\beta_2$
stated in Definition \ref{def:blp}. \qed

\subsection*{Proof of Theorem \ref{theorem: GATES}} The proof is similar to the proof of Theorem  \ref{theorem: BLP1}- \ref{theorem: BLP2}. Moreover, since
the proofs for  the two strategies are similar, we will only demonstrate the proof for the second strategy.

The subset of the normal equations, which correspond to $\mu :=(\mu_k)_{k=1}^K$, are given by $
\Ep [ ( Y H - \mu_0' X_1 H  - \mu' \tilde W_2)  \tilde W_2] =0.
$
Substituting  $Y = b_0(Z) + s_0(Z) D + U$, and using the definition $\tilde W_2 =  \tilde W_2(Z) = [1(G_k)_{k=1}^K]'$,
 $X_1 = X_1(Z)$,  and the law of iterated expectations, we notice that:
$$
\begin{array}{l}
 \Ep [ b_0(Z) H \tilde W_2(Z) ] = \Ep [ b_0(Z) \underbracket{\Ep[H(D,Z) \mid Z]}_{=0}  \tilde W_2(Z) ]  = 0, \\
\Ep [ U H \tilde W_2(Z) ] = \Ep [  \underbracket{\Ep [U \mid Z,D]}_{0}  H(D,Z) \tilde W_2(Z) ] = 0, \\
\Ep [ X_1 H \tilde W_2(Z)] = \Ep [ X_1 (Z) \underbracket{\Ep[H(D,Z) \mid Z]}_{=0} \tilde W_2 (Z) ]  = 0. \\
\end{array}
$$

Hence the normal equations simplify to:
$\Ep  [ \{s_0(Z) D H -  \mu' \tilde W_2\}   \tilde W_2] =0.$
Since components of $\tilde W_2 =  \tilde W_2(Z) = [1(G_k)_{k=1}^K]'$ are orthogonal, the 
normal equations above further simplify to  $
\Ep  [\{  s_0(Z) D  H -  \mu_k  1(G_k)   \}  1(G_k) ] =0.$
Using that
$
\Ep [D H \mid Z]=1,$
$S = S(Z)$, and the law of iterated expectations, the equations simplify to
$$\begin{array}{l}
\Ep [ \{ s_0(Z) -   \mu_k  1(G_k)   \}  1(G_k)  ] =0  \Longleftrightarrow \ \
 \mu_k = \Ep{s_0(Z) 1(G_k)  }/ \Ep[1(G_k)] = \Ep [ s_0(Z) \mid  G_k].  
\end{array} $$ 
The asserted result follows.  \qed

\subsection*{Proof of Comment \ref{OLS:CATE}}\label{sec:proofOLSnotCATEBLP}
We assume that all variables are square integrable. We consider a pure RCT  where $p(Z) = p$. 
Define 
\begin{align*}
\gamma &:=  
\mathrm{Cov}(B,S)/\Var(S)= p^{-1} \mathrm{Cov}(B,DS)/\Var(S), \\
R & := D (S - \Ep S) - p\gamma  (B- \Ep B).
\end{align*}
We can re-parameterize
$$
Y = \tilde \alpha_1 + \tilde \alpha_2 B + \tilde \beta_1 D +  \tilde \beta_2 D (S - \Ep_{} S)    + \epsilon, \quad \Ep[\epsilon \tilde X] = 0,
$$
where $\tilde X = (1, B(Z), D, D(S-\Ep S))$, as follows:
$$
Y = \bar \alpha_1 + \tilde \alpha_2 (B- \Ep B) + 
\tilde \beta_1 (D-p) + \tilde \beta_2 D (S - \Ep S) + \epsilon,
$$
for $\bar \alpha_1 = \tilde \alpha_1 + \tilde \alpha_2 \Ep B + \tilde \beta_1 p$.
Also,
$$
Y = b_0(Z) + s_0(Z) D + U, \quad \Ep [U \mid Z, D] =0,
$$
where  $U$ is independent of any function of $Z$ and $D$. 

Note that $B - \Ep B$, $D-p$, $1$, and $U$ are mutually orthogonal. Also $D(S- \Ep S)$ is orthogonal to $1, D-p,$ and $U$.   Using these notes and Frisch-Waugh-Lovell theorem, it is standard to verify that $\tilde \beta_1 = \beta_1$. Using these notes and the Frisch-Waugh-Lovell theorem, we obtain
\begin{align*}
\tilde \beta_2 & = \Ep Y R/ \Ep R^2  \\
& = \Ep(b_0(Z)+s_0(Z)D +U)R]/ \Ep R^2 \\
& = \Ep[s_0(Z)D +b_0(Z)]R/ \Ep R^2 \\
& = \Ep[s_0(Z)D R]/ \Ep R^2
+ \Ep [b_0(Z)R]/ \Ep R^2 \\
& = \mathrm{Cov}(s_0(Z), D R)/ \Var(R) + \mathrm{Cov}(b_0(Z), R)/ \Var(R).
\end{align*}
This expression does not simplify to $\beta_2 = \mathrm{Cov}(s_0(Z), S(Z))/ \Var(S(Z))$ in general.

Here are some sufficient conditions for the simplification: Suppose $B-\Ep B$ spans $S - \Ep S$, so we can set 
$ B=S$ without loss of generality, then $\gamma =1$, and 
$$
R= (D-p) (S-\Ep S),
$$
in which case 
$$
 \mathrm{Cov}(s_0(Z), D R)/ \Var(R) = \beta_2, \ \ \Ep[b_0(Z) R]/ \Var(R) = 0.  
$$

Another sufficient condition is when $b_0(Z)$ and $s_0(Z)$
are both uncorrelated to $B$ and $S$, in which case
$$
\tilde \beta_2 = \beta_2 =0.
$$

Finally, consider the case where $S$ and $B$ are uncorrelated. In this case 
 $\gamma =0$ so that
$$
R = D (S - \Ep S),
$$
which gives
$$
\tilde \beta_2 = \beta_2 
+ p \Ep b_0(Z) (S- E S)/\Ep R^2, 
$$
which simplifies to $\beta_2$ if  $b_0(Z)$ is also uncorrelated to $S$. \qed

\subsection*{Comparison of the Second Stage Estimation Strategies for BLP of CATE}\label{app:comparison}
We focus on the estimation of the BLP. The analysis can be extended to the GATES using analogous arguments.

Let $X_{2i} = (1, S_i - \Bbb{E}_{N,M} S_i)'$. In the first strategy, we run the weighted linear regression
$$
Y_i =  X_{1i}'\hat \alpha_0 +  (D_i - p(Z_i)) X_{2i}'\hat \alpha    + \hat \epsilon_i, \quad i \in M, \quad \Bbb{E}_{N, M}[ w(Z_i) \hat \epsilon_i 
X_i] = 0,$$ 
where $w(Z) = \{p(Z)(1-p(Z))\}^{-1},$ $X _{i}= [X_{1i}', (D_i - p(Z_i)) X_{2i}']'$, and $\hat \alpha := (\hat \alpha_1, \hat \alpha_2)'.$
Let $\hat \theta := (\hat \alpha_0', \hat \alpha')'$. Then, 
$$
\hat \theta  =\left( \Bbb{E}_{N, M}[ w(Z_i) X_i X_i' ] \right)^{-1} \Bbb{E}_{N, M}[ w(Z_i) X_i Y_i ].
$$
Let $X =  [X_{1}', (D - p(Z)) X_{2}']'$ with $X_2 =  (1, S - \Ep S)'$. By standard properties of the least squares estimator and the central limit theorem
$$
\hat \theta  =\left(\Ep[ w(Z) X X' ] \right)^{-1} \Bbb{E}_{N, M}[ w(Z_i) X_i Y_i ] + o_P(|M|^{-1/2}),
$$
where
$$
\Ep[ w(Z) X X' ] =  \left(\begin{array}{cc} \Ep w(Z)X_1X_1' & 0 \\0 & \Ep X_2 X_2'\end{array}\right).
$$
 In the previous expression we use that $\Ep w(Z) (D - p(Z)) X_{1}  X_{2}' = 0$ and $\Ep w(Z) (D - p(Z))^2 X_{2}  X_{2}' = \Ep  X_2 X_2'$ by iterated expectations.
Then, 
$$
\hat \alpha  =\left(\Ep  X_2 X_2'  \right)^{-1} \Bbb{E}_{N, M}[ w(Z_i) (D_i - p(Z_i)) X_{2i} Y_i ] + o_P(|M|^{-1/2}),
$$
using that  $\Ep[ w(Z) X X' ]$ is block-diagonal between $\hat \alpha_0$ and $\hat \alpha$.

In the second strategy, we run the linear regression
$$
H_i Y_i =  H_i X_{1i}'\hat \mu_0 + X_{2i}' \hat \mu + \hat \varepsilon_i,  \quad  \Bbb{E}_{N,M} \hat \varepsilon_i \tilde X_i =0,
$$
where $H_i = (D_i - p(Z_i)) w(Z_i)$, $\tilde X_i = [H_i X_{1i}', X_{2i}]'$ and $\hat \mu := (\hat \mu_1, \hat \mu_2)'$. Let $\tilde \theta := (\hat \mu_0', \hat \mu')'$. Then,
$$
\tilde \theta  =\left( \Bbb{E}_{N, M}[\tilde X_i \tilde X_i' ] \right)^{-1} \Bbb{E}_{N, M}[ H_i \tilde X_i Y_i ].
$$
Let $\tilde X =  [H X_{1}',  X_{2}']'$ with $X_2 =  (1, S - \Ep S)'$. By standard properties of the least squares estimator and the central limit theorem
$$
\tilde \theta  =\left(\Ep[\tilde X \tilde X' ] \right)^{-1} \Bbb{E}_{N, M}[ H_i \tilde X_i Y_i ] + o_P(|M|^{-1/2}),
$$
where
$$
\Ep[ \tilde X \tilde X' ] =  \left(\begin{array}{cc} \Ep w(Z)X_1X_1' & 0 \\0 & \Ep X_2 X_2'\end{array}\right) = \Ep[ w(Z) X X' ].
$$
In the previous expression we use that $\Ep H X_{1}  X_{2}' = 0$ and $\Ep H^2 X_{1}  X_{1}' = \Ep  w(Z) X_1 X_1'$ by iterated expectations.
Hence,
$$
\hat \mu  =\left(\Ep[  X_2 X_2' ] \right)^{-1} \Bbb{E}_{N, M}[ w(Z_i) (D_i - p(Z_i)) X_{2i} Y_i ] + o_P(|M|^{-1/2}),
$$
where we use that  $\Ep[ \tilde X \tilde X' ] $ is block-diagonal between $\hat \mu_0$ and $\hat \mu$, and  $\Bbb{E}_{N, M}[ H_i X_{2i} Y_i ]  = \Bbb{E}_{N, M}[ w(Z_i) (D_i - p(Z_i)) X_{2i} Y_i ]$.

We conclude that  $\hat \alpha$ and $\hat \mu$ have the same asymptotic distribution because they have the same first order representation.

\section{Deferred Discussion, Results and Proofs for Section \ref{sec:inference}}

\begin{figure}

\begin{center}

		\includegraphics[width=.5\textwidth, height=.5\textwidth]{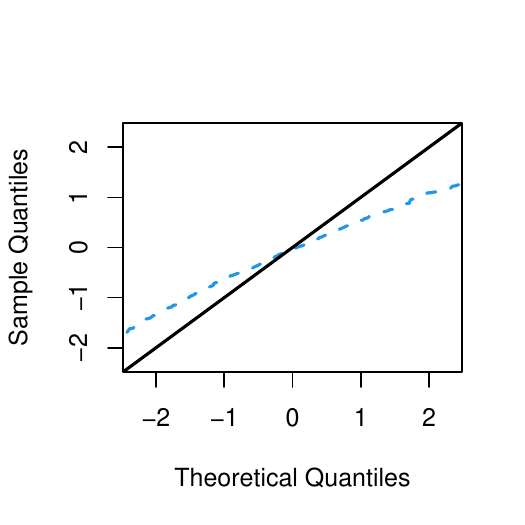}\includegraphics[width=.5\textwidth, height=.5\textwidth]{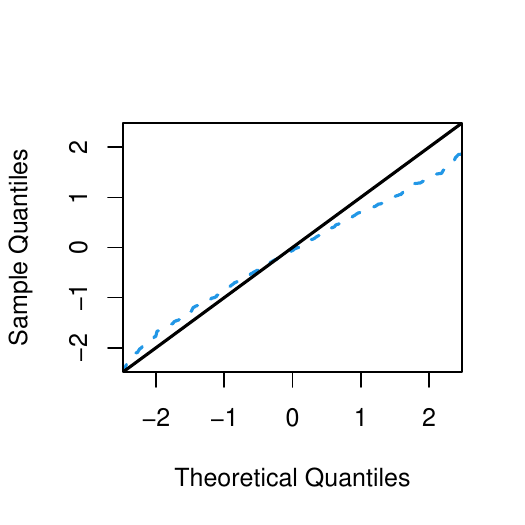}
		\end{center}
		
      \caption{A simple Monte-Carlo experiment illustrating inferential robustness with and without concentration conditions.}
      \caption*{\footnotesize \textsc{Notes.} This example shows that the actual quantiles of the statistic $\mathrm{M}( \hat \sigma_A^{-1}(\hat \theta_A- \theta^*) \mid \mathrm{Data})$ are conservatively bounded by those of  $N(0,1)$ with concentration and without concentration. The estimand $\theta_A$ is generated from $U(0, K)$, with $K= 1/\sqrt{N}$ in the left panel (almost homogeneous case) or with  $K=10$ in the right panel (strong heterogeneous case). The estimator $\hat \theta_A$ is generated as $\theta_A + 1/|M| \sum_{i\in M} \varepsilon_i$, where $\varepsilon_i$'s are i.i.d. exponential random variables centered to have mean zero.
      The main sample indices $M$ are randomly drawn from $\{1,...,N\}$ without replacement, with $N=600$ and the subsample size  $n/N=1/3$.  $\hat \sigma_A$ is the classical standard error for the sample mean. In the left figure we get $99.5 \% $ coverage, and in the right $98.2 \% $ coverage for the nominal level of $95\% $. The results are based on 100 splits, and 1,000 replications.} 
      
       \label{fig:MCcoverage}	

\end{figure}

\begin{remark}[Robustness of the Coverage Property] In our numerical results, the coverage property in Theorem \ref{theorem:CI}  is satisfied even if only (R1) holds. This suggests that it may be possible to establish the coverage property under much weaker conditions. In particular, the coverage property holds without the concentration conditions (R2) and (R3) if \begin{equation}\label{reg:concentrate median-X}
\begin{array}{ll}
\Pr \left(  |\mathrm{M}(\hat \sigma^{-1}_A  (\hat \theta_A - \theta^*)\mid \mathrm{Data})| > z \right) \leq \Pr \left( |\hat \sigma^{-1}_A (\hat \theta_A - \theta_A)| > z \right)+ \gamma'''_N. 
\end{array}
\end{equation}
Perhaps surprisingly, this property does hold in numerical experiments even when $\theta_A$ does not concentrate around $\theta^*$; see, e.g. Figure \ref{fig:MCcoverage}. However, formally demonstrating this property proved difficult and remains an unresolved problem for future research. 
\end{remark}

\subsection*{Other Issues: Stratified Splitting, Small Variation of Proxies} The idea of stratified sample splitting is to balance the proportions of treated and untreated units in both $a$ and $m$ samples so that the proportion of treated units is equal to the experiment's propensity scores across strata.  This balance potentially improves the performance of the inferential algorithms. Stratified sampling formally requires us to replace the i.i.d. assumption with an i.n.i.d. assumption (independent but not identically distributed observations). The inference results continue to apply as long as the conditions (R1), (R2), (R3) hold.  We conjecture that these conditions continue to be plausible under stratified splitting.

Another issue is that the analysis may generate proxy predictors $S$ that have little variation, so we can think of them as ``weak". This causes some target parameters to be weakly identified, e.g., the BLP parameter,  leading to the potential breakdown of the basic normal approximation (\ref{reg:basic}), which our inferential results rely on. To avoid this issue, we can add small noise to the proxies (jittering) so that inference results go through.

\subsection*{Proof of Lemma \ref{lemma:risk}}
To show \eqref{eq:risk} note that,
\begin{equation*}
\Ep  | \hat \theta- \theta'|
= \Ep \Ep [ | \hat \theta- \theta'| \mid \mathrm{Data}]
\leq \Ep \Ep[| \hat \theta_A- \theta'| \mid \mathrm{Data}]
\leq \Ep | \hat \theta_A- \theta'|,
\end{equation*}
where the inequality follows 
from (any) median minizing average
absolute loss and its equivariance property. The equalities hold by the law of iterated expectation. The claim \eqref{eq:risk2} follows in the same way.

To show \eqref{eq:risk3}, let $U^*= \{ U_a^*\}_{a \in \mathcal{A}}$ and $L^*= \{ L_a^*\}_{a \in \mathcal{A}}$ denoted non-decreasing monotone rearrangements of
$\{ U_a\}_{a \in \mathcal{A}}$ and $L= \{ L_a\}_{a \in \mathcal{A}}$. Then 
$$
| U - L| \leq  \| U^* - L^*\|_\infty  \leq \| U - L\|_\infty,
$$
where the second inequality follows from the rearrangement having contractive property in the max distance. \qed

\subsection*{Proof of Theorem \ref{theorem: PV}}
We demonstrate the result for $p^+_{}$. The proofs for other p-values follow similarly. We use $M_{\mathcal{A}}[\cdot]$ as short hand for $\mathrm{M[\cdot | Data]}$, with overlined and underlined versions defined similarly.

To show claim (ii) we note that for $z = \Phi^{-1}(1-\alpha)$ and using that $\Phi(z) = 1- \alpha$:  
\begin{eqnarray*}
\Pr\left ( {M}_{\mathcal{A}}[ 1-\Phi( \hat \sigma^{-1}_A (\hat \theta_A - \theta_0)) ] < \alpha \right)
& = & \Pr\left ( {M}_{\mathcal{A}}[ \Phi( - \hat \sigma^{-1}_A (\hat \theta_A - \theta_0)) ] < \alpha - 1\right) \\
& = &  \Pr \left ( \  {M}_{\mathcal{A}}[  \hat \sigma^{-1}_A (\theta_A - \hat \theta_A) ] < -z \right) \\
& \leq & 
\Pr \left ( \hat \sigma^{-1}_A
(\theta_A - \hat \theta_A) < -z \right) + \gamma''_N \\
& \leq & \Phi(-z) + \gamma'_N + \gamma''_N= \alpha + \gamma'_N+ \gamma''_N,
\end{eqnarray*}
where the first inequality uses the concentration of median assumption, and the last inequality follows from the approximate normality assumption
(\ref{reg:basic}).

To show claim (i), we note that
\begin{eqnarray*}
\Pr\left ( {M}_{\mathcal{A}}[ 1-\Phi( \hat \sigma^{-1}_A (\hat \theta_A - \theta_0)) ] < \alpha \right)
& = &  \Pr \left ( \  {M}_{\mathcal{A}}[  \hat \sigma^{-1}_A (\theta_A - \hat \theta_A) ] < -z \right) \\
& \leq &  \Pr \left (
    \frac{1}{\mathcal{A}} \sum_{a \in \mathcal{A}}
    1 \{ \ \hat \sigma^{-1}_A (\theta_A - \hat \theta_A) ] < -z \} \geq 1/2\right ) \\
& \leq &   2 \Ep \left[ \frac{1}{\mathcal{A}} \sum_{a \in \mathcal{A}} 
    1 \{ \ \hat \sigma^{-1}_A (\theta_A - \hat \theta_A) ] < -z \}\right]\\
& = &  2  \Pr  \{ \ \hat \sigma^{-1}_A (\theta_A - \hat \theta_A) ] < -z \} \\
& \leq & 2 \Phi(-z) + 2\gamma'_N = 2\alpha + 2\gamma'_N,
\end{eqnarray*}
where the first equality reused the previous calculation, the first inequality holds by definition of the numerical median, the second inequality holds by Markov inequality, and the equality that follows  holds by 
\begin{eqnarray*}
 2 \Ep \left[ \frac{1}{\mathcal{A}} \sum_{a \in \mathcal{A}} 
    1 \{ \ \hat \sigma^{-1}_a (\theta_a - \hat \theta_a) ] < -z \} \right]
 & = &  2 \Ep \Pr \left (  \ \hat \sigma^{-1}_A (\theta_A - \hat \theta_A)  < -z\mid \mathrm{Data} \right ) \\
&=&   2  \Pr \{ \ \hat \sigma^{-1}_A (\theta_A - \hat \theta_A) ] < -z\}, 
\end{eqnarray*}
and the last inequality follows from the approximate normality assumption (\ref{reg:basic}).


\subsection*{Proof of Theorem \ref{theorem:CI2}}  Define $\mathcal{D} = \{ \theta_a, [L_a, U_a]: a \in \mathcal{A}\}$, and let $A \sim U(\mathcal{A})$ given $\mathcal{D}$. Then, 
\begin{eqnarray*}
\Pr ( \theta_A < L)    &  = & \Ep \Pr ( \theta_A < L \mid \mathcal{D})   =  \Ep \left[ \frac{1}{|\mathcal{A}|}\sum_{a \in \mathcal{A} } 
1\{ \theta_a < L \} \right]\\
 & = & \Ep \left[\frac{1} {|\mathcal{A}|}\sum_{a \in \mathcal{A} }
1\{ \theta_a < L , \ L_a < L \} + 
 \frac{1}{|\mathcal{A}|}\sum_{a \in \mathcal{A} }
1\{ \theta_a < L , \ L_a \geq L \} \right] \\
& \leq & \Ep \left[\frac{1}{|\mathcal{A}|}\sum_{a \in \mathcal{A} }
1\{ L_a < L \} \right] + 
\Ep  \left[ \frac{1}{|\mathcal{A}|}\sum_{a \in \mathcal{A} }
1\{ \theta_a < L_a \} \right] \\
& \leq & \Ep (\beta) + 
\Ep  \left[\frac{1}{|\mathcal{A}|}\sum_{a \in \mathcal{A} }
1\{ \theta_a < L_a \}\right]  \\
& \leq & \beta + 
\Ep  \Pr( \theta_A < L_A \mid \mathcal{D}) \\
& \leq &  \beta + \Pr\{\theta_A < L_A\} \leq  \beta + \alpha/2 + o(1),
\end{eqnarray*}
where the first equality holds by the law of iterated expectations; the second by the fact that, given $\mathcal{D}$, $L$ is fixed but $A \sim U(\mathcal{A})$; the second inequality holds by definition of $L$:
$$
\frac{1}{|\mathcal{A}|}\sum_{a \in \mathcal{A} }
1\{ L_a < L \} 
\leq \frac{1}{|\mathcal{A}|}\sum_{a \in \mathcal{A} }
1\{ L_a < \overline{Q}_{\beta}[L_A \mid \mathrm{Data}] \}
\leq \beta,
$$
by the definition of the upper quantile; and the third by the same argument as the second equality, the penultimate inequality holds by the law of iterated expectations, and the last inequality holds by assumption  (\ref{eq:brackets}).  We conclude similarly
\begin{eqnarray*}
\Pr ( \theta_A > U)  \leq  \beta + \Pr\{\theta_A > U_A\}  \leq  \beta + \alpha/2 + o(1). 
\end{eqnarray*}
The asserted result follows. \qed

\subsection*{Proof of Theorem \ref{theorem:CI}} In the proof let $z = \Phi^{-1}(1-\alpha/2)$, and use $M_{\mathcal{A}}[\cdot]$ and $Q_{\mathcal{A}}[\cdot]$ as short hand for $\mathrm{M[\cdot | Data]}$ and $\mathrm{Q[\cdot | Data]}$, respectively,  with overlined and underlined versions defined similarly. 

We first note
\begin{eqnarray*}
&& \hspace{-.25in }\Pr (  \theta^* \not \in [L,U])
 =  \Pr( \theta^* > {M}_{\mathcal{A}} ( \hat \theta_A + z \hat \sigma_A)) + \Pr( \theta^* < {M}_{\mathcal{A}}  ( \hat \theta_A - z \hat \sigma_A)) \\
& = & \Pr( 0 > {M}_{\mathcal{A}}  ( \hat \theta_A - \theta^* + z \hat \sigma_A)) + \Pr( 0 < {M}_{\mathcal{A}} ( \hat \theta - \theta^* - z \hat \sigma_A)). 
\end{eqnarray*}

To show part (i) with $\beta=1/2$,  
\begin{eqnarray*}
\Pr( 0 < {M}_{\mathcal{A}} ( \hat \theta - \theta^* - z \hat \sigma_A))
& \leq & \Pr \left ( 
\frac{1}{|\mathcal{A}|} \sum_{a \in \mathcal{A}}
1 \left ( \hat \sigma_a^{-1}(\hat \theta_a - \theta^*) > z   \right) \geq 1/2 
\right)  \\
& \leq  &  2 \Ep \left[
\frac{1}{|\mathcal{A}|} \sum_{a \in \mathcal{A}}
  1 \left ( \hat \sigma_a^{-1}(\hat \theta_a - \theta^*)>  z \right)  \right]  \\
& = & 2 \Ep \Pr \left ( \hat \sigma_A^{-1}( \hat \theta_A - \theta^*) > z \mid \mathrm{Data} \right ) \\
& = & 2 \Pr \left ( \hat \sigma_A^{-1}( \hat \theta_A - \theta^*) > z\right ) \\
& \leq & 2 \Pr \left ( \hat \sigma_A^{-1}( \hat \theta_A - \theta_A) > z -r_N\right) + 2 \gamma'''_N \\
 & \leq &  2(1-\Phi(z-r_N)) + 2\gamma'_N+ 2 \gamma'''_N  \\ 
 &\leq & 2\alpha/2 + 2r_N/\sqrt{2 \pi} + 2\gamma'_N + 2 \gamma'''_N,
 \end{eqnarray*}
where the first inequality follows from the definition of the numerical median, the second from the Markov
inequality; the first equality holds
by $A \sim U(\mathcal{A})$ given $\mathrm{Data}$, the second by the law of iterated expectations; the third inequality holds by the concentration condition (R3) and the union bound, the penultimate inequality holds by the approximation normality conditions (R1), and the last from the properties of $\Phi$.  

We derive similarly that
\begin{eqnarray*}
 \Pr( 0 > {M}_{\mathcal{A}}  ( \hat \theta_A - \theta^* + z \hat \sigma_A)  )  
 \leq  2\alpha/2 + 2r_N/\sqrt{2 \pi} + 2\gamma'_N + 2 \gamma'''_N.
\end{eqnarray*}

Therefore the part (i)  holds for the term $$o(1):=  4r_N/\sqrt{2 \pi} +  4(\gamma'_N+\gamma'''_N).$$

To show part (ii),  from the analysis of part (i),
\begin{eqnarray*}
&& \Pr( 0 < {M}_{\mathcal{A}} ( \hat \theta - \theta^* - z \hat \sigma_A))
\leq  \Pr \left ( 
\frac{1}{|\mathcal{A}|} \sum_{a \in \mathcal{A}}
1 \left ( \hat \sigma_a^{-1}(\hat \theta_a - \theta^*) > z   \right) \geq 1/2
\right).  
 \end{eqnarray*}
Then we bound
$$
T=\frac{1}{|\mathcal{A}|} \sum_{a \in \mathcal{A}}
1 \left ( \hat \sigma_a^{-1}(\hat \theta_a - \theta^*) > z   \right)
\leq  T_1 + T_2 $$
$$
T_1 := \frac{1}{|\mathcal{A}|} \sum_{a \in \mathcal{A}}
1 \left ( \hat \sigma_a^{-1}(\hat \theta_a - \theta_a) > z - r_N  \right), \quad T_2 := 
\frac{1}{|\mathcal{A}|} \sum_{a \in \mathcal{A}}
1 \left ( \hat \sigma_a^{-1} ( \theta_a - \theta^*)> r_N \right).
$$

By the union bound
$$\Pr(T \geq 1/2) \leq
\Pr \left (T_1 > 1/2- 2\sqrt{\gamma_N'''} \right) + 
\Pr \left ( T_2 \geq \sqrt{\gamma_N'''} \right ).$$

Then for $\beta_N =1/2-2\sqrt{\gamma_N'''}$, 
\begin{eqnarray*}
 \Pr (T_1 > \beta_N)
& \leq & \Pr\left ( \overline{\mathrm{Q}}_{\beta_N,\mathcal{A}}[
\hat \sigma_A^{-1}( \theta_A- \hat \theta_A)]< -z +r_N \right) \\
& \leq & \Pr\left ( \mathrm{Q}_{\beta_N,\mathcal{A}}[
\hat \sigma_A^{-1}( \theta_A- \hat \theta_A)]< -z +r_N \right) \\
&\leq&  \Pr\left (
\hat \sigma_A^{-1}( \theta_A- \hat \theta_A)< -z +r_N \right) + \gamma''_N \\
& \leq &  \Phi(-z+r_N) + \gamma'_N+  \gamma''_N  \\ 
 &\leq & \alpha/2 + r_N/\sqrt{2 \pi} + \gamma'_N +  \gamma''_N,
\end{eqnarray*}
where first inequality holds by the definition of the numerical quantile, the third by the concentration of medians assumption (R2), and  the fourth by the approximate normality (R1). 

Also, by Markov inequality
\begin{eqnarray*}
\Pr\left ( T_2 \geq \sqrt{\gamma'''_N} \right)
\leq \Ep T_2/\sqrt{\gamma'''_N} 
= \Pr \left ( \sigma_A^{-1} | \theta_A - \theta^*|> r_N \right)/\sqrt{\gamma'''_N} 
\leq \gamma'''_N/\sqrt{\gamma'''_N},
\end{eqnarray*}
where we are using (R3) and the relation
\begin{eqnarray*}
\Ep T_2 &=&  \Ep \left[ \frac{1}{\mathcal{A}}  \sum_{a \in \mathcal{A}} 1 \left ( \sigma_a^{-1}| \theta_a - \theta^*|> r_N \right) \right] \\
&= & \Ep \Pr ( \sigma_A^{-1} | \theta_A - \theta^*|> r_N  \mid \mathrm{Data})=\Pr \left ( \sigma_A^{-1} | \theta_A - \theta^*|> r_N \right),
\end{eqnarray*}
using our formalism that $A \sim U(\mathcal{A})$ independently of $\mathrm{Data}.$

Collecting terms conclude 
$$
\Pr( 0 < {M}_{\mathcal{A}} ( \hat \theta - \theta^* - z \hat \sigma_A)) \leq 
\alpha/2 + r_N/\sqrt{2 \pi} + \gamma'_N +  \gamma''_N+ \sqrt{\gamma'''_N}.
$$

We derive similarly that
\begin{eqnarray*}
 \Pr( 0> {M}_{\mathcal{A}}  ( \hat \theta_A - \theta^* + z \hat \sigma_A ) )  
 \leq  \alpha/2 + r_N/\sqrt{2 \pi} + \gamma'_N +  \gamma''_N+ \sqrt{\gamma'''_N}.
\end{eqnarray*}

Therefore the part (ii)  holds for the term 
$$
o(1) =  2\left (r_N/\sqrt{2 \pi} + \gamma'_N +  \gamma''_N+ \sqrt{\gamma'''_N} \right). 
$$

To show claim (iii) note that by construction $L \leq \hat \theta \leq U$ and the coverage event 
$L \leq \theta^* \leq U$  implies  that $|\hat \theta - \theta^*|\leq U-L.$ \qed

\subsection*{Concentration of Estimands Around Their Median}\label{sec:stability}

The purpose of this section is to demonstrate that the concentration assumptions made in the inference section are plausible. To show this
we focus on the BLP parameter
$$
\theta_A =\frac{\mathrm{Cov}_Z (s_0(Z), S_A(Z)}{ \Var_Z S_A(Z)) },
$$
where $A$ is a uniform variable on $\mathcal{A}$, and the variance and covariance are taken with respect to the marginal distribution of $Z$. We want to show the concentration of this parameter around
$$
\theta^* = \mathrm{Med}[\theta_A \mid \mathrm{Data}].
$$
We show the difference can be bounded using measures of estimation and algorithmic stabilities; we derive inspiration from \cite{CWZ:JASA} and \cite{vasilis:stable}).  

In what follows, we assume the same set-up as in the main text, in particular the exchangeability.

\subsection*{Estimation Stability or Pseudo-Consistency}
Statistical learning theory, for example, results in Section 5, provides bounds on estimation errors of the form:
$$
\Ep \Ep_Z (S_A(Z) - s_\bullet(Z))^2 = \Ep  (S_A(Z) - s_\bullet(Z))^2 \leq R^2_{|A|},
$$
where $s_\bullet$ is a fixed ``pseudo-true" function that does not depend on $A$, and this function does not have to be the CATE $s_0$ in the misspecified case. Here $\Ep_Z$ denotes the expectation taken with respect to the marginal distribution of $Z$.  For example, in Section 5, $s_\bullet$ minimizes the mean square approximation error $$
\min_{s \in \mathcal{S}} \Ep [s_0(Z) - s(Z)]^2
= \Ep [s_0(Z) - s_\bullet(Z)]^2,
$$
but $s_\bullet$ above does not to be defined in this way.

Define the BLP parameter corresponding to $s_\bullet$ as:
$$
\theta_\bullet =\frac{\mathrm{Cov}_Z (s_0(Z), s_\bullet(Z))}{ \Var_Z s_\bullet(Z)) }.
$$
This is a fixed estimand.

If $R_{|A|} \to 0$ as $|A| \to \infty$, then $S_A$ converges to the pseudo-true value $s_\bullet$. We call this property ``pseudo"-consistency. The lemma shows that in this case, the random estimand $\theta_A$ approaches $\theta_\bullet$ at the rate $R_{|A|}$.

\begin{lemma}[Concentration from ``Pseudo''-Consistency]\label{lemma:e-stability} Assume that $S_a \in \mathcal{S}$ for all $a \in \mathcal{A}$ and $s_\bullet \in \mathcal{S}$, that the elements of $\mathcal{S}$ and $s_0$ are all bounded above by a finite constant $K$, and that $\Var_Z S(Z)$ is bounded below by a positive constant $k>0$ for all $S \in \mathcal{S}$. Then 
$$
\Ep |\theta_A - \theta_\bullet| \leq C_{K,k}  [R_{|A|} \wedge  1]
$$
where $C_{K,k}$ is a numeric constant that only depends on $K$ and $k$.
\end{lemma}

\subsection*{Concentration under Algorithmic Stability}
On the other hand, algorithmic or statistical influence analysis often implies that
$$
\Ep \Ep_Z (S_A(Z) - S_{A'}(Z))^2 \leq {R'}^2_{|A|},
$$
where $A$ and $A'$ are independent uniform variables on  $\mathcal{A}$.  To explain the notion, let $M$ and $M'$ be the complements of $A$ and $A'$ relative to $\{1,...,N\}$.  The symmetric difference between $A$ and $A'$ is  $M \cap M'$. If the latter set is small in cardinality relative to the cardinality of $A$, then we would expect $S_A$ and $S_{A'}$ not to differ if the machine producing $S$'s is a smooth function of data. The definition above provides one way to measure this stability. We provide further discussion below.

By triangle inequality, the algorithmic stability can be bounded by estimation stability:
$$
\sqrt{\Ep \Ep_Z (S_A(Z) - S_{A'}(Z))^2} \leq 2 \sqrt{\Ep \Ep_Z (S_A(Z) - S_{\bullet}(Z))^2}
$$
Therefore algorithmic stability is more general. 

\begin{lemma}[Concentration from Algorithmic Stability]\label{lemma:a-stability} Suppose the assumptions of the previous lemma hold. Then if $R'_{|A|} \to 0$ as $|A| \to \infty$, then 
$$
\Ep |\theta_A - \theta_{A'}|  \leq C_{K,k}  [R'_{|A|} \wedge  1],
$$
where $C_{K,k}$ is a numeric constant that only depends on $K$ and $k$.
\end{lemma}

\subsection*{Putting it Together: Concentration Around Median}  

The following result shows that the desired concentration condition holds if either estimation stability or algorithmic stability is strong enough.

\begin{lemma}[Stability of Median Target from Estimation or Algorithmic Stability]\label{lemma:m-stability} Suppose the assumptions of the previous lemma hold. Then
$$
\Ep | \theta_A - \theta^*| \leq \Ep |\theta_A - \theta_{A'}| \wedge \Ep |\theta_A - \theta_\bullet|.
$$
Therefore, if 
\begin{equation}\label{eq:side}
\sqrt{n} C_{K,k} [R'_{|A|} \wedge R_{|A|}] \leq \delta_N
\end{equation}
for $\delta_N \to 0$ as $N \to \infty$, then 
$$
\Pr ( \sqrt{n}| \hat \theta_A - \theta^* | > \sqrt{\delta_N}) \leq \sqrt{\delta_N}.
$$
\end{lemma}
The latter implies the condition we want provided $\hat \sigma_A/\sqrt{n} + 
\sqrt{n}/\hat \sigma_A = O_P(1)$.

We conclude here with some comparisons of the two notions of stability.
Estimation stability readily follows from the available statistical learning theory. In particular $R^2_{|A|}$ scales like $d/|A|$ where $d$ is the intrinsic dimension of the function class $\mathcal{S}$, as we discussed in Section 5. Therefore, $n$ needs to be much smaller than $d (N-n)$ to satisfy the last condition of the last lemma.

Algorithmic stability does not require estimation stability, even though the latter property seems quite mild. On the other hand, its characterizations are not well-studied and are much less available. See \cite{CWZ:JASA} for analysis of constrained Lasso and Ridge that is applicable here; see also \cite{vasilis:stable}  for leave-one-out stability analysis for bagged estimators over the subsamples (this analysis requires extension to the present framework).

It is useful to give a simple example to compare the two measures of stability. If $S_A(Z)$'s are generated by linear least squares $Z'\hat \beta_A$ with $d= \dim(Z)$, then we have a crude upper bound on the algorithmic stability bound ${R'}^2_{A}$ scaling like $n d/(N-n)^2$.  This is generally smaller than $R_A$ scaling like $d/(N-n)$. It implies a weaker same qualitative requirement on $n$: $n$ needs to be smaller than $\sqrt{d} (N-n)$ to satisfy the condition (\ref{eq:side}) of Lemma \ref{lemma:m-stability}.

\subsection*{Proof of Lemmas \ref{lemma:e-stability}- \ref{lemma:m-stability}}

To show Lemma \ref{lemma:e-stability}, it is convenient to define $f^o(Z) = f(Z) - \Ep_Z f(Z)$.  Then, using  the boundedness assumption we have
\begin{eqnarray*}
|\mathrm{Cov}_Z(s_0(Z), S_A(Z)) -\mathrm{Cov}_Z(s_0(Z), s_\bullet(Z))|  &= & 
|\Ep_Z [s_0^o(Z) S_A(Z)] - \Ep  [s_0^o(Z) s_\bullet(Z)]|\\
&\leq &  K \Ep_Z |S_A(Z) -  s_\bullet(Z))|,
\end{eqnarray*}\begin{eqnarray*}
|\mathrm{Var}_Z(S_A(Z)) -\mathrm{Var}_Z(s_\bullet(Z))|  &= & 
|\Ep_Z (S^o_A(Z))^2 - \Ep_Z (s^o_\bullet (Z))^2| \\
&\leq &  \Ep_Z |S^o_A(Z) +  s^o_\bullet(Z))| |S^o_A(Z) -  s^o_\bullet(Z))| \\
&\leq & 2 K \Ep_Z |S^o_A(Z) -  s^o_\bullet(Z))|.
\end{eqnarray*}
Then using elementary inequalities
and boundedness assumptions conclude 
$$
|\theta_A - \theta_{\bullet}| 
\leq (k^{-1} K + 2 k^{-2} K^2) \Ep_Z |S_A(Z) -  s_\bullet(Z)|$$
Taking expectation over $A$, 
$$
\Ep |\theta_A - \theta_{\bullet}| \leq (k^{-1} K + 2 k^{-2} K^2) \Ep \Ep_Z |S_A(Z) -  s_\bullet(Z))|
\leq C_{K,k} R_{|A|},$$
where the last inequality follows from the  norm inequality.

Lemma \ref{lemma:a-stability} follows analogously, replacing $s_\bullet$ with $S_{A'}$ to obtain
$$
|\theta_A - \theta_{A'}| 
\leq (k^{-1} K + 2 k^{-2} K^2) \Ep_Z |S_A(Z) -  S_{A'}(Z)|$$
Taking expectation over $(A, A')$, we obtain:
$$
\Ep |\theta_A - \theta_{A'}| \leq (k^{-1} K + 2 k^{-2} K^2) \Ep \Ep_Z |S_A(Z) -  S_{A'}(Z))|
\leq C_{K,k} R'_{|A|},$$
where the last inequality follows from the  norm inequality.

To show Lemma \ref{lemma:m-stability}, we note that
\begin{eqnarray*}
 \Ep | \theta_A - \theta^*|
 & = &   \Ep \Ep \left[ \frac{1}{\mathcal{A}} \sum_{a \in \mathcal{A}}| \theta_a - \theta^*|   \mid \mathrm{Data} \right] \\
& \leq &   \Ep \Ep \left[ \frac{1}{\mathcal{A}} \sum_{a \in \mathcal{A}}| \theta_a - \theta_\bullet|   \mid \mathrm{Data} \right]  =    \Ep | \theta_A - \theta_\bullet|,
\end{eqnarray*}
where the first property holds by the law of iterated expectation and by $A \sim U( \mathcal{A})$ independently of the $\mathrm{Data}$, the inequality holds by definition of $\theta^*$ as the median of the sample $\{\theta_a: a \in \mathcal{A}\}$,
and the last equality holds by iterating expectations again.

Similarly, we note 
\begin{eqnarray*}
\Ep | \theta_A - \theta^*|
&=&  \Ep \Ep \left[ \frac{1}{\mathcal{A}} \sum_{a \in \mathcal{A}}| \theta_a - \theta^*|   \mid \mathrm{Data} \right]\\
& = &
  \Ep \frac{1}{|\mathcal{A}|} \sum_{a' \in \mathcal{A}} \Ep \left[  \frac{1}{|\mathcal{A}|} \sum_{a \in \mathcal{A}}| \theta_a - \theta^*|   \mid \mathrm{Data} \right] \\
 & \leq &
  \Ep \frac{1}{|\mathcal{A}|} \sum_{a' \in \mathcal{A}} \Ep \left[  \frac{1}{|\mathcal{A}|} \sum_{a \in \mathcal{A}}| \theta_a - \theta_{a'}|   \mid \mathrm{Data} \right]  \\
  & =&  
 \Ep \left[ \frac{1}{|\mathcal{A}|} \sum_{a' \in \mathcal{A}} \frac{1}{|\mathcal{A}|} \sum_{a \in \mathcal{A}}| \theta_a - \theta_{a'}|    \mid \mathrm{Data} \right]  =  \Ep | \theta_A - \theta_{A'}|,
\end{eqnarray*}
where the first property holds by the law of iterated expectation and by $A \sim U( \mathcal{A})$ independently of the $\mathrm{Data}$, the inequality holds by definition of $\theta^*$ as the median of $\{\theta_A: a \in \mathcal{A}\}$, the last equality holds by iterating expectations and independence of $A$ and $A'$.

Finally, the second claim of the Lemma follows by the Markov inequality. \qed

\section{Deferred Discussion and Proofs for Section \ref{sec:further}}\label{app:sec5}

\begin{remark}[Extensions of Theorem \ref{lemma:rate}]
The result of Theorem \ref{lemma:rate} follows from combining Theorem 3 of \cite{rakhlin:offset} with Theorem \ref{lemma:oracle}. We assumed boundedness conditions to make the statement as simple as possible.  Bounds on excess risk without the boundedness conditions follow from Theorem 4 in \cite*{rakhlin:offset}.  If the class $\mathcal{S}$ is not convex, similar performance bound is attained by \cite{audibert2007progressive}'s "star" modification of the optimizer $S$, by Theorems 3 and 4 by \cite{rakhlin:offset}.  We refer to \cite{rakhlin:offset} and \cite{suhas:offset} for detailed general discussion.  
\end{remark}

\subsection*{Using Losses \eqref{eq:A} and \eqref{eq:B} for Choosing the Best ML Method}

The loss functions \eqref{eq:A} and \eqref{eq:B}  can also be used to aggregate several learning methods using separate auxiliary subsets.\footnote{This is in contrast to our main proposal, where we choose the best ML method based on goodness-of-fit measures in the second stage.} To fix ideas, suppose we have a  set of methods giving scores $S_k$ and $B_k$, $k=1,...,K$, where $K$ is small, obtained using a subset $A_1 \subset A$.  Then,
we can combine these scores into  
$$
S(Z) = \sum_{k=1}^K \lambda^S_k S_k(Z); \ \ B(Z) =
\sum_{k=1}^K \lambda^B_k B_k(Z),
$$
and then we can learn the weights $\lambda^S$ and $\lambda^B$  by optimizing the loss functions (A) or (B) evaluated on subset $A_2$, such that $A_2$ does not overlap with $A_1$, e.g. $A_2 = A \setminus A_1$. 

\begin{remark}[Learning Guarantee for Aggregation] Let $\hat \lambda^S$ and $\hat \lambda^B$ denote the weights learned in this way.  Then, another application of the results of \cite{rakhlin:offset} for linear regression, under the assumption that $|Y|$, $|B|$, $|S|$, $|w(Z)|$ are all bounded by $R$, gives the excess risk bound:
$$
\Ep\Big [ s_0(Z) - \sum_{k=1}^K \hat \lambda^S_k S_k(Z) \Big]^2
- \overbracket{\min_{\{\lambda^S_k\}_{k=1}^K}\Ep\Big[ s_0(Z) - \sum_{k=1}^K \lambda^S_k S_k(Z)\Big]^2}^{\mathrm{ oracle \ \ risk}}\leq  C_R K/|A_2|,
$$
where $C_R$ is some constant that depends on $R$ and $|A_2|$ is the sample size used to perform the aggregation.  Thus, if the right-hand side is small, the excess risk of this estimator relative to the oracle aggregation method is negligible.  Since the oracle aggregation risk here is weakly smaller than the oracle risk of choosing the best prediction rule $\min_k \Ep[ s_0(Z) -  S_k(Z)]^2$, convex aggregation here is approximately better than choosing the best ML method.

\end{remark}

\begin{remark}[Large $K$] The method above gives a small excess risk when $K/|A_2|$ is small; otherwise, the excess risk can be large.  In the latter case, we can apply Lasso to select a sparse linear combination of rules, and the sharp bounds on excess risk follow from Example 4 in \cite{KLT}.  Finally, 
we may choose the ``best" machine learning algorithm using objective functions \eqref{eq:A} and \eqref{eq:B}  evaluated on the data subset $A_2$. 
Results of \cite{wegkamp2003}  imply certain good guarantees for the "best" approach, but sharp bounds on the excess risk that scale like $\log K/|A_2|$ only hold for the "star" modification of the "best" method \citep{audibert2007progressive}. 
\end{remark}

\subsection*{Proof of Theorem \ref{lemma:oracle}}  For the  objective \eqref{eq:B}, write 
 $Y = b_0(Z) + D s_0(Z) + \epsilon$, where $\Ep [\epsilon \mid D, Z]=0$. Then
$$
YH = \{H b_0(Z) + (HD-1) s_0(Z)\} + s_0(Z) + \epsilon H,
$$
where the first term can be expressed as: $$
\{H b_0(Z) + (HD-1) s_0(Z)\}
= H (b_0(Z) + (1-p(Z)) s_0(Z)) = H \bar b_0(Z).
$$
So that we can decompose:
$$
YH - b(Z) H - s(Z) =  \{ H (\bar b_0(Z) - b(Z))\} + \{s_0(Z) - s(Z)\} + \epsilon H.
$$
Then  the result follows taking the square and expectation, by using  (i) orthogonality of the three terms in the decomposition above:
$$
\Ep [ \epsilon H^2 (\bar b_0(Z)- b(Z))] = 0, \ \ \Ep[ \epsilon H (s_0(Z)- s(Z))] = 0, \ \ \Ep[ H (\bar b_0(Z)- b(Z)) ( s_0(Z) - s(Z)] =0,
$$
where the last relation follows from $\Ep [H \mid Z] =0$, and (ii) also noting that $\Ep [H^2|Z] = w(Z)$.

For the objective \eqref{eq:A}, write similarly,
$$
Y - b(Z) - (D-p(Z)) s(Z) 
=  [\tilde b_0(Z) - b(Z)] 
+  [D-p(z)] (s_0(Z) - s(Z))  + \epsilon,
$$
and then conclude that the three terms are orthogonal to each other.  The result follows by completing the square and taking expectation, where we also observe that
$$
\Ep w(Z) (D-p(Z))^2(s_0(Z)-s(Z))^2 = \Ep (s_0(Z)-s(Z))^2 ,
$$
since $\Ep [w(Z) (D-p(Z))^2\mid Z]=1$. \qed

\subsection*{Proof of Theorem \ref{lemma:rate}} We demonstrate the result for type B loss; the demonstration for type A follows similarly. Application of Theorem 3 of \cite{rakhlin:offset} gives the following
bound on the excess risk $\mathcal{R}$ of the estimator $(B, S)$:
$$
0 \leq \mathcal{R} := \Ep [YH - B(Z) H - S(Z) ]^2 
- \Ep [YH - b_\bullet(Z) H - s_\bullet(Z) ]^2  \leq C_K \mathcal{R}^o(A, \mathcal{H}, c_K),
$$
where $(b_\bullet, s_\bullet)$ minimize $\Ep [YH - b_\bullet(Z) H - s_\bullet(Z) ]^2$ over $b \in \mathcal{B}$ and $s \in \mathcal{S}$, and $C_K$ and $c_K$ are positive constants that only depend on $K$, and $\mathcal{H}:= 4(H\mathcal{B}+ \mathcal{S})$. Theorem \ref{lemma:oracle} then implies that
\begin{eqnarray*}
\mathcal{R}  &=&   \Ep [s_0(Z)- S(Z) ]^2  - \Ep [s_0(Z)- s_\bullet(Z) ]^2 \\
& + & \Ep [w(Z) (\bar b_0(Z) - B(Z)) ]^2  - \Ep [w(Z) (\bar b_0(Z) -  b_\bullet(Z)) ]^2,
\end{eqnarray*}
where the second term is non-negative. Therefore,
$$
 \Ep [s_0(Z)- S(Z) ]^2  - \Ep [s_0(Z)- s_\bullet(Z) ]^2 \leq C_K \mathcal{R}^o(A, \mathcal{H}, c_K).
$$
The lower bound 
$$
 \Ep [s_0(Z)- S(Z) ]^2  - \Ep [s_0(Z)- s_\bullet(Z) ]^2 \geq \Ep[S(Z) - s_\bullet(Z)]^2
$$
follows from Pythagorian inequality for obtuse triangles and the fact that $s_\bullet$ minimizes $\Ep[s_0(Z) - S(Z)]^2$ over the convex set $\mathcal{S}$. \qed

\section{Gaussian Approximation for Split-Sample Least Squares Uniformly Over Convex Sets and in $P$.}

We present a set-up that covers not only the split-sample least square estimators of the main text, but also other potential cases of interest. Let $W$ denote a generic data vector.  All the linear regressions or mean estimators used on the main sample $M$ could be viewed as ordinary least squares with a suitable definition of $W$. 

Throughout we assume that  $\{(W_i)\}_{i=1}^N$ are i.i.d. copies of vector $W$ that has law $P$.  We abbreviate $
(\mathcal{D}_A, \mathcal{D}_M) :=(\mathrm{Data}_A, \mathrm{Data}_M).$ There is a learning algorithm that inputs $\mathcal{D}_A$ and outputs a map $f(\cdot; \mathcal{D}_A)$, which maps the support of $W$ to $\Bbb{R}^{d+1}$ for a fixed $d$. This map  defines the split-specific outcome and regressors:
$$(Y_{A,i}, X_{A,i})= f(W_i; \mathcal{D}_A), \quad i \in M.$$ Let $\hat \beta_A$ be a solution to 
 $
 \mathbb{E}_{N,M} [X_{A,i} \hat \epsilon_{A,i}  ] = 0$ for $\hat \epsilon_{A,i} = Y_{A,i} - X_{A,i}'\hat \beta_A.
 $
Let $\hat V_{A}$  denote the Eicker-Huber-White sandwich $$\hat V_{A} :=    (\mathbb{E}_{N,M}  X_{A,i} X_{A,i}')^{-1}   \mathbb{E}_{N,M} \hat \epsilon^2_{A,i}  X_{A,i} X_{A,i}'    (\mathbb{E}_{N,M}  X_{A,i} X_{A,i}')^{-1},$$
whenever it exists.

Fix some positive finite constants $c$ and $C$. Let $\beta_A$ denote a solution to
$\Ep[ X_A \epsilon_A] = 0$, for $\epsilon_A = Y_A - X_A'\beta_A,$ if it exists. And let  $$V_{A} :=    (\Ep_P [X_A X_A' \mid \mathcal{D}_A])^{-1}   \Ep_P [\epsilon^2_A  X_A X_A'  \mid \mathcal{D}_A] (\Ep_P [X_A X_A' \mid \mathcal{D}_A])^{-1},$$
if it exists. Let $\mathcal{E}_{A,N}$ be the event that 
\begin{eqnarray*}
&& \Ep_P |Y_A|^{4+\delta} + \Ep_P[\|X_A \|^{4+\delta} \mid  \mathcal{D}_A] \leq C, \quad \min_{\|a\|=1} \Ep_P [(a'X_A)^2 \mid  \mathcal{D}_A] > c.
\end{eqnarray*}
On this event $\beta_A$ and $\epsilon_A$ are well defined. Let $\mathcal{E}'_{A,N} \subset \mathcal{E}_{A,N}$ be the event such that 
$$
\min_{\|a\|=1} \Ep_P[ (\epsilon_A a'X_A)^2  \mid \mathcal{D}_A] >c.
$$
On this event $V_{N}$ is well-defined.  Let $CS(\Bbb{R}^d)$ denote the collection of the convex sets in $\Bbb{R}^d$.

We observe that, by the i.i.d. sampling and $A \sim U(\mathcal{A})$ independently of $\mathrm{Data}$, 
$(\mathcal{D}_A, \mathcal{D}_M)$ has the same distribution as $(\mathcal{D}_a, \mathcal{D}_m)$,
for a fixed partition $(a, m)$. 
This is an exchangeability property. Therefore, we can fix $(A,M)$ to be a fixed partition $\{a,m\}$ in what follows. Moreover $(X_{a,i}, Y_{a,i})_{i=1}^{N-m}$ are i.i.d. conditional on $\mathcal{D}_a$. These observations simplify the verification of the following result.

\begin{lemma}[Gaussian Approximation]\label{lemma: estimation} Using the setup above, let $\gamma_N$ be a sequence of positive constants tending to zero. Suppose that for all $P \in \mathcal{P}$, we have $\Pr_P(\mathcal{E}'_{N,A}) \geq 1- \gamma_N$.  Then, uniformly in $P \in \mathcal{P}$, as $(n,N) \to \infty$:
$$
\sup_{R \in CS(\Bbb{R}^d) } \left |\Pr_P[ \hat V_{A}^{-1/2}  (\hat \beta_A - \beta_A)  \in  R  \mid \mathcal{D}_A] - \Pr ( N(0,I_d) \in R) \right | \overset{\Pr_P}{\longrightarrow} 0,
$$
$$
\sup_{R \in CS(\Bbb{R}^d)} \left |\Pr_P [ \hat V_{A}^{-1/2} (\hat \beta_A - \beta_A) \in R] - \Pr ( N(0,I_d) \in R) \right | {\longrightarrow} \ 0,
$$
and the same results hold with $\hat V_{A}$ replaced by $V_{A}$; moreover, $\hat V_N V^{-1}_N \to_{\Pr_P} I$ both conditional on $\mathcal{D}_N$ and unconditionally.

\end{lemma}

\subsection*{Proof of Lemma \ref{lemma: estimation}} 
It suffices to demonstrate the argument for an arbitrary sequence $\{P_N\}$ in $\mathcal{P}$. 
Let
\begin{eqnarray*}
    \hat t_{A} := \hat V_{A}^{-1/2} (\hat \beta_A - \beta_A), \ 
     t_{A} :=  V_{A}^{-1/2} (\hat \beta^o_A - \beta_A),  \ \hat \beta^o_A
    := [\Ep X_{A} X_{A}']^{-1} \mathbb{E}_{N,M} X_A Y_A.
\end{eqnarray*}

Consider the event $\mathcal{E}_{N,A}'' \subseteq \mathcal{E}'_{N,A}$ such that:
$$
\mathcal{E}_{N,A}'' = \left\{ (\hat t_{A}, t_{A}, \hat V_N) \text{ exist and }  \| \hat t_{A} - t_{A}\| + \| \hat V_N- V_N\| \leq r_N \right\}.
$$
It follows from the standard arguments for asymptotic theory for least squares under i.i.d. sampling of data arrays $(Y_{A,i}, X_{A,i})_{i=1}^N$, e.g. \cite{gallant:white}, that there exists a sequence of positive constants $\{r_N, \delta_N\} \searrow 0$ such that  $
\Pr \left( \mathcal{E}_{N,A}'' \mid \mathcal{D}_{A}\right) \geq 1- \delta_N$ on the event  $\mathcal{E}'_{N,A}$. Therefore by the union bound
\begin{equation}\label{eq:event}
\Pr \left( \mathcal{E}_{N,A}''\right) \geq 1- \delta_N - \gamma_N,
    \end{equation}
for $\gamma_N$ defined in the statement of the lemma.    
For $r>0$ let $R^{r}=\{x \in \Bbb{R}^d: d(x, R) \geq r\}$ and $R^{-r}=\{x \in R: d(x, \Bbb{R}^d\setminus R) \geq r\}$, where $d(x, R) := \min_{x' \in R} \| x'- x\|$. Note that $R^{-r}$ can be an empty set. Then, on the event $\mathcal{E}''_{N,A}$,
\begin{eqnarray*}
\Pr \left ( \hat t_A \in R \mid  \mathcal{D}_{A}\right) & \geq & 
\Pr \left ( t_A \in R^{-r_N} \mid  \mathcal{D}_{A} \right)\\
& \geq & \Pr (N(0,I_d) \in R^{-r_N}) - B_N d^{1/4}/\sqrt{n}, \\
& \geq & \Pr (N(0,I_d) \in R) - 4 d^{1/4} r_N 
 - B_N d^{1/4} /\sqrt{n},
\end{eqnarray*}
where $B_N  = C'\Ep [  \| V_N^{-1/2} X_A \epsilon_A\|^3 \mid \mathcal{D}_{A}],$ where $C'$ is a numerical constant.  The second inequality follows by the Bentkus bounds \citep{bentkus,raic}, which extend the Berry-Essen bounds to the multidimensional case, and the last inequality follows from the Ball's reverse isoperimetric inequality of the standard Gaussian vector \citep{ball:isoperimetry}.  It follows similarly that \begin{eqnarray*}
\Pr \left ( \hat t_A \in R \mid  \mathcal{D}_{A}\right) & \leq & 
\Pr \left ( t_A \in R^{r_N} \mid  \mathcal{D}_{A} \right)\\
& \leq & \Pr (N(0,I_d) \in R^{r_N}) + B_N/\sqrt{n}, \\
& \leq & \Pr (N(0,I_d) \in R) + 4 d^{1/4} r_N 
+ B_N d^{1/4}/\sqrt{n}.
\end{eqnarray*}
 Since  $R$ above is arbitrary convex subset of $ \Bbb{R}^d$, we have that on the event $\mathcal{E}''_{N,A}$:
 \begin{eqnarray*}
&& \sup_{R \in CS(\Bbb{R}^d)} |\Pr \left ( \hat t_A \in R \mid  \mathcal{D}_{A}\right) -\Pr (N(0,I_d) \in R) | \\
&& \leq \sup_{R \in CS(\Bbb{R}^d)}|\Pr \left ( \hat t_A \in R \mid  \mathcal{D}_{A}\right) -\Pr (N(0,I_d) \in R) |  \\
& & \leq
4 d^{1/4} r_N 
 + d^{1/4} B_N/\sqrt{n}.
 \end{eqnarray*}
 Using Holder inequalities, we can check  that $B_N \leq B$ on the event $\mathcal{E}_N''$, for some constant $B$ that depends only on $(c,C, d, \delta)$. 
The first claim follows combining this inequality with (\ref{eq:event}). 

 To show that second claim note that
 \begin{eqnarray*}
&&\sup_{R \in CS(\Bbb{R}^d)} |\Ep \Pr \left ( \hat t_A \in R \mid  \mathcal{D}_{A}\right) -\Pr (N(0,I_d) \in R) | \\
&& \quad  \leq 4 d^{1/4} r_N 
 + d^{1/4} B_N/\sqrt{n}
 +  (1- \Pr (\mathcal{E}''_{N,A}))
 \leq 4  d^{1/4} B_N/\sqrt{n}
 +  \gamma_N + \delta_N.
 \end{eqnarray*}

Finally,  $\|\hat V_N V^{-1}_N - I\| \leq \| V^{-1}_N\| r_N \leq c r_N$ conditional on $\mathcal{D}_{A}$ and on the event $\mathcal{E}''_N$. The conditional convergence claim follows from (\ref{eq:event}).  It then follows that $\Ep \Pr(\|\hat V_N V^{-1}_N - I\| \leq c r_N \mid \mathcal{D}_A) \geq \Pr( \mathcal{E}_{N,A}) \geq 1-\delta_N - \gamma_N$. \qed

%% file: Appendix-AdditionalTheory.tex
\section{Extension to Unbiased Signal Framework} \label{app:sec7}

Our inference approach generalizes to any problem of the following sort, studied in \cite{semenova2020} using more conventional inference approaches. Suppose we can construct an \textit{unbiased signal} $\tilde Y$ such that $$\Ep [\tilde Y \mid Z]  = s_0(Z),$$ where $s_0(Z)$ is now a generic target function. Let $S(Z)$ denote an ML proxy for $s_0(Z)$.  In experimental settings the unbiased signals arise from multiplying an outcome with a Riesz representer for the effect of interest, as we explain below.

Then, using previous arguments, we immediately can generate the following conclusions:
\begin{enumerate}
\item The projection of $\tilde Y$ on the ML proxy $S(Z)$ identifies the BLP of $s_0(Z)$ on $S(Z)$.
\item The grouped average of the target (GAT)  $\Ep[ s_0(Z) \mid G_k] $ is identified by  $\Ep[ \tilde Y \mid G_k] $.
\item Using ML tools we can train proxy predictors $S(Z)$ to predict $\tilde Y$  in auxiliary samples.
\item We  can post-process $S(Z)$ in the main sample, by estimating the  BLP and GATs.
\item We can perform split-sample robust inference on functionals of the BLP and GATs.
\end{enumerate}

\begin{example}[ Forecasting or Predicting   Regression Functions using ML proxies]   This is the most common type of the problem arising in forecasting. Here the target is the best predictor of $Y$ using $Z$, namely $s_0(Z) = \Ep[Y \mid Z]$, and $\tilde Y = Y$ trivially serves as the unbiased signal. The interesting part here is the use of the inference tools developed in this paper for constructing confidence intervals for the predicted values produced by the estimated BLP of $s_0(Z)$ using $S(Z)$.
\end{example}


\begin{example}[Predicting  Structural Derivatives using ML proxies]  Suppose we are interested in 
 predicting the conditional average partial derivative $s_0(z) = \Ep[g'(D,Z) \mid Z=z]$, where $g'(d,z) = \partial g(d,z)/ \partial x$ and  $g(d,z)= \Ep [Y \mid D= d, Z=z]$.
In the context of demand analysis, $Y$ is the log of individual demand, $D$ is the log-price of a product, and $Z$ includes prices of other products and individual characteristics. Then, the unbiased signal  is given by $ \tilde Y = -  Y [\partial  \log p(D \mid Z) /\partial d],$ where  $p(\cdot \mid \cdot)$ is the conditional density function of $D$ given $Z$, which is known if $D$ is generated experimentally conditional on $Z$. That is, using the integration by parts formula, $\Ep [ \tilde Y  \mid Z] = s_0(Z)$ under mild conditions on the density.  

\end{example}

\begin{example}[Other Causal Objects] \cite{CNS:AutoLocal} presented a number of other examples where a causal parameter of interest $s_0(Z)$ is expressed as a linear functional of the regression function $g(D,Z)= \Ep[Y \mid D, Z]$, that is, $s_0(Z) = \Ep[ m(Y,D,Z, g)\mid Z]$, for some moment function $m$ that is linear in $g$; this includes the examples above for instance. Then, under mild conditions, we can construct an unbiased signal 
\begin{equation}
\tilde Y = Y \alpha(D,Z),
\end{equation}
where $\alpha(D,Z)$ is the Riesz Representer, such that $\Ep[Y \alpha(D,Z)\mid Z] = s_0(Z)$. For instance, in CATE, the representer $\alpha(D,Z)$ is the HT transform $H$; in Example 2, $\alpha(D,Z)=[\partial  \log p(X \mid Z) /\partial x]$; and in Example 1, the representer $\alpha(D,Z)$ is just $1$. In addition to these examples, other examples that fall in this framework include causal effects from transporting covariates and from distributional shift in covariates induced by policies; see
\cite{CNS:AutoLocal} for more details.  In experimental settings,  $\alpha(D,Z)$ will typically be known. 

The noise reduction strategies, like the ones we used in the context of H-transformed outcomes, can be useful in these cases as well. For this purpose we can use terms of the form $ \{\alpha(D,Z) - \Ep [\alpha(D,Z) \mid Z]\} B(Z)$ for denoising where $\alpha(D,Z)$ now plays the same role as $H$ before. 
\end{example}

\clearpage

%% file: AdditionalEmpirical.tex
\section{Additional Empirical Results}  \label{app:empirical}

\begin{table}[H]
\centering
\caption{CLAN of Immunization Incentives: Other Covariates-1}\label{table:Haryana_CLAN1}
\fontsize{7}{7}\selectfont
\begin{tabular}{lcccccc}
  \hline
  \hline
  \\ [-1mm]
       & \multicolumn{3}{c}{Elastic Net}  & \multicolumn{3}{c}{Nnet} \\   [1mm]
& 20\% Most  & 20\% Least  & Difference & 20\% Most  & 20\% Least  & Difference\\ [0.5mm] 
 & ($\delta_{5}$) &  ($\delta_1$)  & ($\delta_{5}-\delta_{1}$) &   ($\delta_{5}$) &  ($\delta_1$)  & ($\delta_{5}-\delta_{1}$) \\ [1mm] 
 \cline{2-7}  \\ [-1mm]
  Fraction Participating in Employment Generating Schemes  & 0.122 & 0.020 & 0.097 & 0.070 & 0.030 & 0.037 \\ 
   & (0.095,0.146) & (-0.002,0.042) & (0.064,0.130) & (0.046,0.094) & (0.008,0.052) & (0.005,0.068) \\ 
   & - & - & [0.000] & - & - & [0.051] \\ 
  Fraction Below Poverty Line (BPL) & 0.181 & 0.194 & -0.016 & 0.183 & 0.177 &  0.007 \\ 
   & (0.126,0.233) & (0.143,0.247) & (-0.089,0.057) & (0.132,0.234) & (0.133,0.223) & (-0.061,0.072) \\ 
   & - & - & [1.000] & - & - & [1.000] \\ 
  Average Financial Status (1-10 scale) & 3.271 & 3.531 & -0.267 & 3.337 & 3.741 & -0.376 \\ 
   & (3.016,3.534) & (3.284,3.762) & (-0.611,0.072) & (3.095,3.587) & (3.499,3.975) & (-0.708,-0.048) \\ 
   & - & - & [0.227] & - & - & [0.047] \\ 
  Fraction Scheduled Caste-Scheduled Tribes (SC/ST) & 0.169 & 0.148 & 0.022 & 0.190 & 0.145 & 0.046 \\ 
   & (0.125,0.213) & (0.106,0.191) & (-0.039,0.084) & (0.146,0.233) & (0.105,0.186) & (-0.014,0.104) \\ 
   & - & - & [0.917] & - & - & [0.261] \\ 
  Fraction Other Backward Caste (OBC) & 0.268 & 0.205 & 0.060 & 0.331 & 0.179 & 0.154 \\ 
   & (0.215,0.319) & (0.155,0.253) & (-0.012,0.133) & (0.284,0.378) & (0.133,0.224) & (0.089,0.220) \\ 
   & - & - & [0.204] & - & - & [0.000] \\ 
  Fraction Minority caste  & 0.005 & 0.005 &  0.000 & 0.004 & 0.006 & -0.002 \\ 
   & (-0.002,0.013) & (-0.002,0.014) & (-0.011,0.010) & (-0.002,0.010) & (0.000,0.011) & (-0.009,0.006) \\ 
   & - & - & [1.000] & - & - & [1.000] \\ 
  Fraction General Caste & 0.217 & 0.464 & -0.239 & 0.230 & 0.500 & -0.273 \\ 
   & (0.154,0.280) & (0.403,0.525) & (-0.331,-0.142) & (0.168,0.293) & (0.439,0.564) & (-0.362,-0.177) \\ 
   & - & - & [0.000] & - & - & [0.000] \\ 
  Fraction No Caste & 0.000 & 0.000 & 0.000 & 0.000 & 0.000 & 0.000 \\ 
   & (0.000,0.000) & (0.000,0.000) & (0.000,0.000) & (0.000,0.000) & (0.000,0.000) & (0.000,0.000) \\ 
   & - & - & [1.000] & - & - & [1.000] \\ 
  Fraction Other Caste & 0.000 & 0.000 & 0.000 & 0.001 & 0.000 & 0.001 \\ 
   & (0.000,0.000) & (0.000,0.000) & (0.000,0.000) & (0.000,0.002) & (-0.001,0.001) & (-0.001,0.002) \\ 
   & - & - & [1.000] & - & - & [0.912] \\ 
  Fraction Dont Know Caste & 0.335 & 0.173 & 0.162 & 0.241 & 0.162 & 0.076 \\ 
   & (0.274,0.399) & (0.115,0.228) & (0.077,0.244) & (0.186,0.294) & (0.112,0.214) & (0.004,0.151) \\ 
   & - & - & [0.000] & - & - & [0.080] \\ 
  Fraction Hindu & 0.806 & 0.956 & -0.142 & 0.955 & 0.955 & -0.002 \\ 
   & (0.725,0.887) & (0.884,1.026) & (-0.253,-0.035) & (0.921,0.987) & (0.907,0.991) & (-0.040,0.037) \\ 
   & - & - & [0.021] & - & - & [1.000] \\ 
  Fraction Muslim& 0.169 & 0.017 & 0.143 & 0.026 & 0.017 & 0.006 \\ 
   & (0.091,0.247) & (-0.047,0.083) & (0.040,0.246) & (0.006,0.048) & (-0.005,0.052) & (-0.013,0.030) \\ 
   & - & - & [0.013] & - & - & [1.000] \\ 
  Fraction Christian & 0.000 & 0.004 & -0.004 & 0.000 & 0.004 & -0.004 \\ 
   & (-0.008,0.008) & (-0.004,0.011) & (-0.015,0.007) & (-0.008,0.008) & (-0.004,0.011) & (-0.014,0.007) \\ 
   & - & - & [1.000] & - & - & [1.000] \\ 
  Fraction Buddhist & 0.000 & 0.000 & 0.000 & 0.000 & 0.000 & 0.000 \\ 
   & (0.000,0.000) & (0.000,0.000) & (0.000,0.000) & (0.000,0.000) & (0.000,0.000) & (0.000,0.000) \\ 
   & - & - & [1.000] & - & - & [1.000] \\ 
  Fraction Sikh & 0.000 & 0.000 & 0.000 & 0.000 & 0.000 & 0.000 \\ 
   & (0.000,0.000) & (0.000,0.000) & (0.000,0.000) & (0.000,0.000) & (0.000,0.000) & (0.000,0.000) \\ 
   & - & - & [1.000] & - & - & [1.000] \\ 
  Fraction Jain & 0.000 & 0.000 & 0.000 & 0.000 & 0.000 & 0.000 \\ 
   & (0.000,0.000) & (0.000,0.000) & (0.000,0.000) & (0.000,0.000) & (0.000,0.000) & (0.000,0.000) \\ 
   & - & - & [1.000] & - & - & [1.000] \\ 
  Fraction Other Religion & 0.000 & 0.000 & 0.000 & 0.000 & 0.000 & 0.000 \\ 
   & (0.000,0.000) & (0.000,0.000) & (0.000,0.000) & (0.000,0.000) & (0.000,0.000) & (0.000,0.000) \\ 
   & - & - & [1.000] & - & - & [1.000] \\ 
   \hline \hline
\end{tabular}
\caption*{\footnotesize Notes: Medians over 250 splits.   Median confidence interval $(\alpha =.05)$ in parenthesis.  P-values for the hypothesis that the parameter is equal to zero against the two-sided alternatives in brackets.}
\end{table}

\newpage

\begin{table}[H]
\centering
\caption{CLAN of Immunization Incentives: Other Covariates-2}\label{table:Haryana_CLAN2}
\fontsize{7}{7}\selectfont
\begin{tabular}{lcccccc}
  \hline
  \hline
  \\ [-1mm]
       & \multicolumn{3}{c}{Elastic Net}  & \multicolumn{3}{c}{Nnet} \\   [1mm]
& 20\% Most  & 20\% Least  & Difference & 20\% Most  & 20\% Least  & Difference\\ [0.5mm] 
 & ($\delta_{5}$) &  ($\delta_1$)  & ($\delta_{5}-\delta_{1}$) &   ($\delta_{5}$) &  ($\delta_1$)  & ($\delta_{5}-\delta_{1}$) \\ [1mm] 
 \cline{2-7}  \\ [-1mm]
Fraction Dont Know religion & 0.031 & 0.023 &  0.007 & 0.015 & 0.023 & -0.008 \\ 
   & (0.013,0.047) & (0.006,0.040) & (-0.017,0.031) & (0.000,0.031) & (0.008,0.039) & (-0.030,0.013) \\ 
   & - & - & [1.000] & - & - & [0.818] \\ 
  Fraction Literate & 0.779 & 0.797 & -0.017 & 0.820 & 0.786 &  0.032 \\ 
   & (0.756,0.801) & (0.775,0.817) & (-0.048,0.015) & (0.804,0.836) & (0.769,0.803) & (0.010,0.052) \\ 
   & - & - & [0.626] & - & - & [0.008] \\ 
  Fraction Single  & 0.053 & 0.046 & 0.006 & 0.051 & 0.043 & 0.007 \\ 
   & (0.046,0.060) & (0.040,0.053) & (-0.004,0.016) & (0.044,0.058) & (0.037,0.049) & (-0.001,0.016) \\ 
   & - & - & [0.465] & - & - & [0.193] \\ 
  Fraction of adults Married (living with spouse) & 0.490 & 0.521 & -0.032 & 0.516 & 0.521 & -0.007 \\ 
   & (0.475,0.506) & (0.508,0.536) & (-0.053,-0.011) & (0.504,0.527) & (0.507,0.534) & (-0.024,0.011) \\ 
   & - & - & [0.006] & - & - & [0.894] \\ 
  Fraction of adults Married (not living with spouse)  & 0.002 & 0.004 & -0.001 & 0.003 & 0.003 &  0.000 \\ 
   & (0.000,0.005) & (0.001,0.006) & (-0.005,0.003) & (0.001,0.005) & (0.001,0.005) & (-0.002,0.003) \\ 
   & - & - & [1.000] & - & - & [1.000] \\ 
  Fraction of adults Divorced or Separated  & 0.006 & 0.002 & 0.005 & 0.005 & 0.001 & 0.004 \\ 
   & (0.004,0.009) & (-0.001,0.004) & (0.001,0.008) & (0.003,0.007) & (-0.001,0.002) & (0.001,0.007) \\ 
   & - & - & [0.010] & - & - & [0.008] \\ 
  Fraction of adults Widow or Widower  & 0.034 & 0.036 & -0.001 & 0.036 & 0.040 & -0.004 \\ 
   & (0.029,0.040) & (0.031,0.041) & (-0.009,0.006) & (0.030,0.041) & (0.034,0.045) & (-0.012,0.003) \\ 
   & - & - & [1.000] & - & - & [0.579] \\ 
  Fraction Marriage Status Unknown & 0.000 & 0.000 & 0.000 & 0.000 & 0.000 & 0.000 \\ 
   & (0.000,0.000) & (0.000,0.000) & (0.000,0.000) & (0.000,0.000) & (0.000,0.000) & (0.000,0.000) \\ 
   & - & - & [1.000] & - & - & [1.000] \\ 
  Fraction Marriage status "NA" & 0.413 & 0.390 &  0.025 & 0.389 & 0.392 & -0.001 \\ 
   & (0.394,0.433) & (0.373,0.406) & (0.000,0.050) & (0.375,0.401) & (0.377,0.406) & (-0.020,0.019) \\ 
   & - & - & [0.103] & - & - & [1.000] \\ 
  Fraction who received Nursery level educ. or less & 0.156 & 0.158 & -0.003 & 0.133 & 0.166 & -0.032 \\ 
   & (0.140,0.171) & (0.144,0.172) & (-0.024,0.018) & (0.121,0.144) & (0.155,0.177) & (-0.047,-0.017) \\ 
   & - & - & [1.000] & - & - & [0.000] \\ 
  Fraction who received Class 4 level educ. & 0.077 & 0.087 & -0.009 & 0.081 & 0.090 & -0.009 \\ 
   & (0.069,0.086) & (0.079,0.095) & (-0.021,0.002) & (0.073,0.090) & (0.082,0.098) & (-0.021,0.002) \\ 
   & - & - & [0.218] & - & - & [0.222] \\ 
  Fraction who received Class 8 level educ. & 0.171 & 0.159 & 0.013 & 0.160 & 0.154 & 0.008 \\ 
   & (0.159,0.183) & (0.148,0.170) & (-0.003,0.030) & (0.148,0.171) & (0.144,0.165) & (-0.009,0.023) \\ 
   & - & - & [0.220] & - & - & [0.736] \\ 
  Fraction who received Class 12 level educ. & 0.204 & 0.232 & -0.028 & 0.246 & 0.226 &  0.018 \\ 
   & (0.185,0.224) & (0.213,0.250) & (-0.056,0.001) & (0.230,0.263) & (0.210,0.241) & (-0.005,0.041) \\ 
   & - & - & [0.119] & - & - & [0.251] \\ 
  Fraction who received Graduate or Other Diploma  & 0.076 & 0.093 & -0.017 & 0.085 & 0.095 & -0.011 \\ 
   & (0.062,0.090) & (0.080,0.106) & (-0.036,0.003) & (0.072,0.098) & (0.082,0.108) & (-0.028,0.007) \\ 
   & - & - & [0.185] & - & - & [0.492] \\ 
  Fraction with education level Other or Dont know  & 0.310 & 0.264 & 0.046 & 0.293 & 0.262 & 0.031 \\ 
   & (0.298,0.323) & (0.252,0.276) & (0.029,0.063) & (0.283,0.305) & (0.252,0.272) & (0.016,0.046) \\ 
   & - & - & [0.000] & - & - & [0.000] \\ 
   \hline \hline
\end{tabular}
\caption*{\footnotesize Notes: Medians over 250 splits.   Median confidence interval $(\alpha =.05)$ in parenthesis.  P-values for the hypothesis that the parameter is equal to zero against the two-sided alternatives in brackets.}
\end{table}

\newpage

\begin{figure}[H] 
\includegraphics[width=0.9\textwidth, height=.58\textwidth]{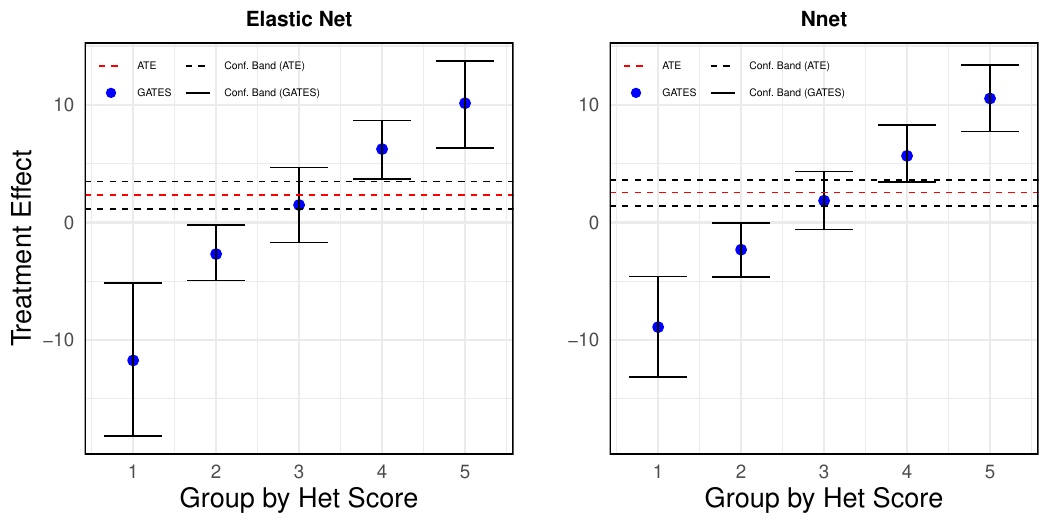}
\caption{\footnotesize GATES of Immunization Full package compared to Ambassadors and SMS only. Point estimates and 90\% adjusted confidence intervals uniform across groups based on 250 random splits in half.}
\label{figure:Haryana_GATESalt}
\end{figure}

\newpage

\begin{table}[H]
\caption{Cost effectiveness for GATE Quintiles, Comparing Full treatment to most cost-effective treatment}\label{table:Haryana_CEalt}

\centering
\fontsize{7}{7}\selectfont
\begin{tabular}{lllllll}
  \hline
& \multicolumn{3}{c}{Elastic Net}   & \multicolumn{3}{c}{Nnet} \\
 & Mean in Treatment  & Mean in Control  & Difference & Mean in Treatment  & Mean in Control & Difference\\ [0.5mm] 
 & ($\hat{\mathrm{E}} [X \mid D =1, G_k  ]) $ & ($\hat{\mathrm{E}} [X \mid D = 0, G_k  ]$) &  & ($\hat{\mathrm{E}} [X \mid D =1, G_k  ]) $ & ($\hat{\mathrm{E}} [X \mid D = 0, G_k  ]$) &\\    \hline
  \hline
Imm. per dollar (All) & 0.036 & 0.043 & -0.006 & 0.036 & 0.042 & -0.006 \\ 
& (0.034,0.038) & (0.041,0.044) & (-0.008,-0.004) & (0.034,0.038) & (0.041,0.044) & (-0.009,-0.004) \\ 
& - & - & [0.000] & - & - & [0.000] \\
Imm. per dollar ($G_1$) & 0.027 & 0.049 & -0.021 & 0.029 & 0.049 & -0.019 \\ 
   & (0.023,0.032) & (0.048,0.050) & (-0.026,-0.016) & (0.026,0.033) & (0.048,0.050) & (-0.023,-0.016) \\ 
   & - & - & [0.000] & - & - & [0.000] \\ 
Imm.per dollar ($G_2$)   & 0.032 & 0.046 & -0.014 & 0.034 & 0.046 & -0.012 \\ 
   & (0.029,0.036) & (0.045,0.048) & (-0.018,-0.010) & (0.031,0.037) & (0.045,0.048) & (-0.016,-0.009) \\ 
   & - & - & [0.000] & - & - & [0.000] \\ 
Imm.per dollar ($G_3$)  & 0.034 & 0.044 & -0.010 & 0.037 & 0.044 & -0.007 \\ 
   & (0.030,0.037) & (0.042,0.046) & (-0.014,-0.006) & (0.034,0.040) & (0.042,0.046) & (-0.011,-0.003) \\ 
   & - & - & [0.000] & - & - & [0.001] \\ 
Imm. per dollar ($G_4$)  & 0.039 & 0.044 & -0.004 & 0.038 & 0.044 & -0.005 \\ 
   & (0.036,0.042) & (0.042,0.046) & (-0.008,-0.001) & (0.035,0.041) & (0.042,0.046) & (-0.009,-0.002) \\ 
   & - & - & [0.015] & - & - & [0.008] \\ 
Imm. per dollar ($G_5$)  & 0.038 & 0.039 & -0.001 & 0.037 & 0.040 & -0.003 \\ 
   & (0.035,0.041) & (0.036,0.042) & (-0.005,0.003) & (0.033,0.040) & (0.037,0.042) & (-0.007,0.001) \\ 
   & - & - & [1.000] & - & - & [0.310] \\ 
\hline
\end{tabular}
\caption*{\footnotesize Notes: Medians over 250 splits.   Median confidence interval $(\alpha =.05)$ in parenthesis.  P-values for the hypothesis that the parameter is equal to zero against the two-sided alternatives in brackets.}
\end{table}

\newpage

\section{Figures and tables - Predictive Learners} \label{sec:predictive2}

\begin{figure}[H] 
\includegraphics[scale=0.9]{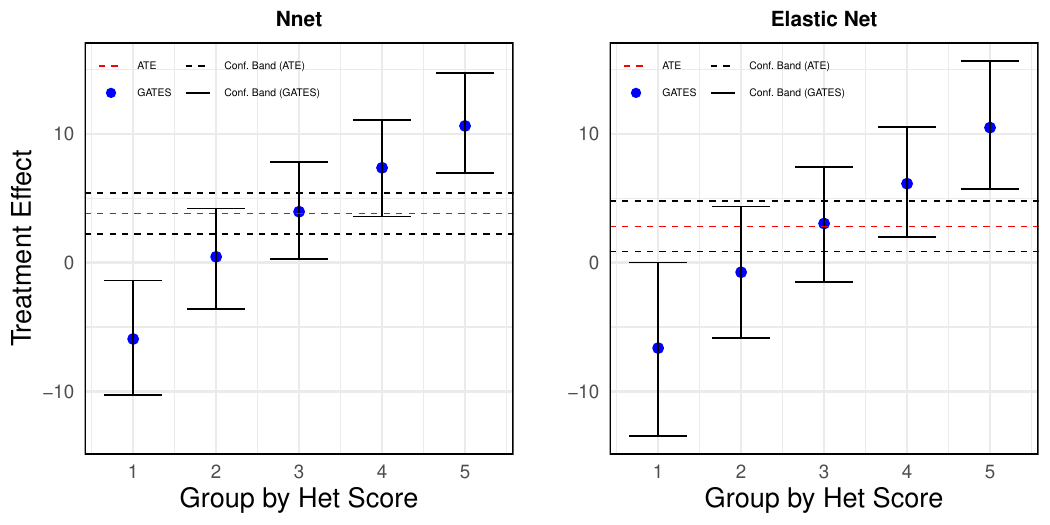}
\caption{\footnotesize GATES of Immunization Incentives, based upon Predictive Learners. Median point estimates and  Median confidence interval $(\alpha =.05)$ in parenthesis, over 250 splits.}
\label{figure:Haryana_GATES_pred2}
\end{figure}

\newpage


\begin{table}[H] 
\centering
\small
\caption{Comparison of ML Methods: Immunization Incentives}
\label{table:Haryana_Best_pred2}
\begin{tabular}{lcccc}
  \hline
    \hline
  \\ [-3mm]
 & Elastic Net & Boosting & Neural Network & Random Forest \\  [1mm]
  \hline \\ [-3mm]
Best BLP ($\Lambda$) & 63.670 & 31.480 & 51.680 & 23.400 \\ 
& [48.879, 77.659] & [23.983, 45.833] & [41.557, 65.97] & [17.538, 31.659] \\ 
Best GATES ($\bar \Lambda$)  & 7.950 &  5.019 &  5.857 &  4.185 \\ 
 & [6.803, 8.938] & [4.194, 6.379] & [5.097, 6.459] & [3.142, 5.158] \\ 
   \hline\hline
\end{tabular}
\caption*{\footnotesize Notes: Medians over 250 splits in half, based upon Predictive Learners. The brackets report interquartile ranges for goodness-of-fit statistics.}
\end{table}


\begin{table}[H]
\caption{BLP of Immunization Incentives}\label{table:Haryana_BLP_pred2}
\small
\begin{tabular}{lcccc}
  \hline
    \hline
  \\ [-3mm]
   & \multicolumn{2}{c}{Elastic Net}  & \multicolumn{2}{c}{Neural Network} \\  [1mm]
 & ATE ($\beta_{1}$) & HET ($\beta_{2}$) & ATE ($\beta_{1}$) & HET ($\beta_{2}$) \\ 
 \cline{2-5}  \\ [-3mm]
&2.806 & 1.070 & 2.462 & 0.902 \\ 
& (1.123,4.681) & (0.840,1.312) & (0.879,3.937) & (0.684,1.116) \\  \relax 
& [0.003] & [0.000] & [0.005] & [0.000] \\    [1mm]
   \hline\hline
\end{tabular}
\caption*{\footnotesize Notes: Medians over 250 splits, based upon Predictive Learners.  Median confidence interval $(\alpha =.05)$ in parenthesis.  P-values for the hypothesis that the parameter is equal to zero against the two-sided alternatives in brackets.}
\end{table}


\begin{table}[H]
\centering
\caption{GATES of  20\% Most and Least Affected Groups}\label{table:Haryana_GATES_pred2}
\footnotesize
\begin{tabular}{lcccccc}
\hline
\hline
\\ [-3mm]
   & \multicolumn{3}{c}{Elastic Net}  & \multicolumn{3}{c}{Nnet} \\   [1mm]
& 20\% Most  &  20\% Least  & Difference & 20\% Most  & 20\% Least  & Difference  \\  [-1mm]
 & ($G_5$) & ($G_1$)  &  & ($G_5$) & ($G_1$)  &    \\
 \cline{2-7}  \\ [-2mm]
GATE $\gamma_k := $  $ \hat{\Ep} [ s_0(Z) \mid G_k  ]$ &  13.300 & -7.362 & 20.88 & 11.260 & -6.063 & 17.33 \\ 
& (8.016,18.89) & (-12.63,-2.005) & (12.94,28.43) & (7.866,14.80) & (-9.792,-2.213) & (12.15,22.75) \\ 
& [0.000] & [0.016] & [0.000] & [0.000] & [0.006] & [0.000] \\ 
\hline\hline
\end{tabular}
\caption*{\footnotesize Notes: Medians over 250 splits, based upon Predictive Learners..   Median confidence interval $(\alpha =.05)$ in parenthesis.  P-values for the hypothesis that the parameter is equal to zero against the two-sided alternatives in brackets.}
\end{table}

\newpage

\begin{table}[H]
\centering
\caption{CLAN of Immunization Incentives }\label{table:Haryana_CLAN_C1_pred2}
\fontsize{8}{8}\selectfont
\begin{tabular}{lcccccc}
  \hline
  \hline
  \\ [-1mm]
       & \multicolumn{3}{c}{Elastic Net}  & \multicolumn{3}{c}{Nnet} \\   [1mm]
& 20\% Most  & 20\% Least  & Difference & 20\% Most  & 20\% Least  & Difference\\ [0.5mm] 
 & ($\delta_{5}$) &  ($\delta_1$)  & ($\delta_{5}-\delta_{1}$) &   ($\delta_{5}$) &  ($\delta_1$)  & ($\delta_{5}-\delta_{1}$) \\ [1mm] 
 \cline{2-7}  \\ [-1mm]
  Number of vaccines  & 2.182 & 2.284 & -0.093 & 2.186 & 2.282 & -0.107 \\ 
  to pregnant mother   & (2.110,2.252) & (2.213,2.355) & (-0.193,0.005) & (2.126,2.246) & (2.216,2.343) & (-0.193,-0.019) \\ 
   & - & - & [0.126] & - & - & [0.033] \\ 
  Number of vaccines  & 4.034 & 4.678 & -0.617 & 4.275 & 4.722 & -0.454 \\ 
  to child since birth     & (3.809,4.260) & (4.474,4.891) & (-0.939,-0.327) & (4.103,4.439) & (4.547,4.895) & (-0.690,-0.213) \\ 
   & - & - & [0.000] & - & - & [0.000] \\ 
  Fraction of children & 0.998 & 1.000 & -0.002 & 1.000 & 1.000 &  0.000 \\ 
  received polio drops    & (0.995,1.001) & (0.997,1.003) & (-0.006,0.002) & (1.000,1.000) & (1.000,1.000) & (0.000,0.000) \\ 
   & - & - & [0.686] & - & - & [0.879] \\ 
  Number of polio & 2.954 & 2.993 & -0.039 & 2.966 & 2.998 & -0.031 \\ 
    drops to child  & (2.933,2.974) & (2.975,3.012) & (-0.066,-0.012) & (2.954,2.978) & (2.985,3.009) & (-0.048,-0.014) \\ 
   & - & - & [0.009] & - & - & [0.001] \\ 
Fraction of children & 0.798 & 0.930 & -0.133 & 0.910 & 0.929 & -0.018 \\ 
    received immunization card  & (0.749,0.848) & (0.885,0.975) & (-0.198,-0.062) & (0.888,0.935) & (0.902,0.956) & (-0.052,0.011) \\ 
   & - & - & [0.001] & - & - & [0.443] \\ 
  Fraction of children received & 0.127 & 0.250 & -0.122 & 0.125 & 0.258 & -0.130 \\ 
  Measles vaccine by 15 months of age   & (0.092,0.163) & (0.216,0.283) & (-0.172,-0.074) & (0.094,0.160) & (0.226,0.289) & (-0.178,-0.085) \\ 
   & - & - & [0.000] & - & - & [0.000] \\ 
 Measles at credible locations  & 0.286 & 0.406 & -0.121 & 0.288 & 0.427 & -0.142 \\ 
   & (0.237,0.333) & (0.362,0.448) & (-0.187,-0.056) & (0.245,0.331) & (0.388,0.470) & (-0.200,-0.083) \\ 
   & - & - & [0.001] & - & - & [0.000] \\ 
   \hline \hline
\end{tabular}
\caption*{\footnotesize Notes: Medians over 250 splits, based upon Predictive Learners.   Median confidence interval $(\alpha =.05)$ in parenthesis.  P-values for the hypothesis that the parameter is equal to zero against the two-sided alternatives in brackets.}
\end{table}

\newpage

\newpage

\section{Figures and Tables - Prediction Intervals} \label{sec:mp_causal2}

\begin{figure}[H] 
\includegraphics[scale=0.9]{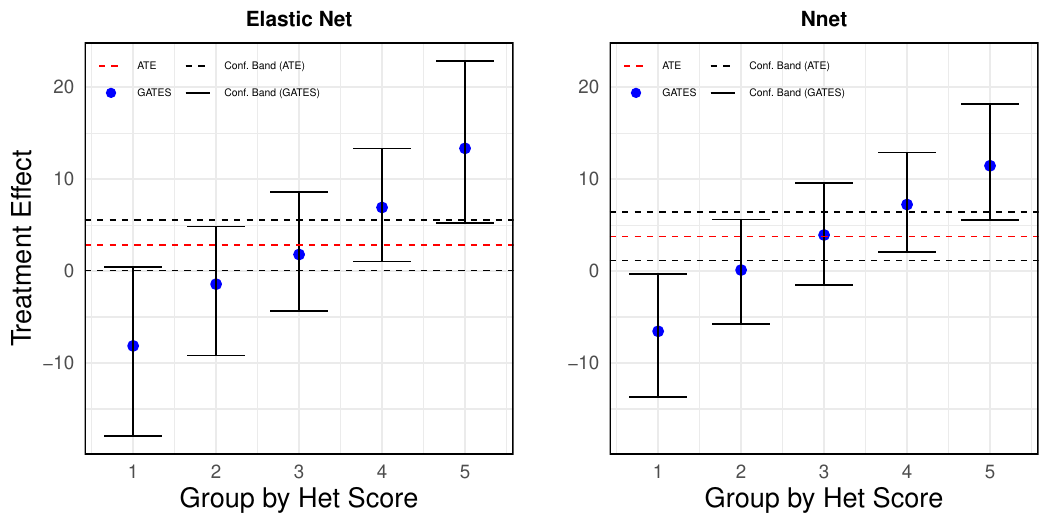}
\caption{\footnotesize GATES of Immunization Incentives, based upon prediction intervals presented in Section \ref{sec:MP}. Median point estimates and  Median confidence interval  $(\alpha =.05, \beta =.225)$, over 250 splits.}
\label{figure:Haryana_GATES_mpcausal2}
\end{figure}

\newpage


\begin{table}[H] 
\centering
\small
\caption{Comparison of ML Methods: Immunization Incentives}
\label{table:Haryana_Best_mpcausal2}
\begin{tabular}{lcccc}
  \hline
    \hline
  \\ [-3mm]
 & Elastic Net & Boosting & Neural Network & Random Forest \\  [1mm]
  \hline \\ [-3mm]
Best BLP ($\Lambda$) & 67.750 & 32.900 & 53.420 & 25.200 \\ 
& [51.491, 82.368] & [23.246, 44.665] & [42.516, 67.647] & [18.328, 34.705] \\ 
Best GATES ($\bar \Lambda$)  & 8.254 &  5.104 &  6.001 &  4.492 \\ 
& [7.329, 9.314] & [4.27, 6.079] & [5.087, 6.888] & [3.339, 5.507] \\  
   \hline\hline
\end{tabular}
\caption*{\footnotesize Notes: Medians over 250 splits in half. Note that we used Neural Network Causal Boosting for all methods, using Algorithm \ref{alg2:implementation}. The brackets report interquartile ranges for goodness-of-fit statistics.}
\end{table}


\begin{table}[H]
\caption{BLP of Immunization Incentives}\label{table:Haryana_BLP_mpcausal2}
\small
\begin{tabular}{lcccc}
  \hline
    \hline
  \\ [-3mm]
   & \multicolumn{2}{c}{Elastic Net}  & \multicolumn{2}{c}{Neural Network} \\  [1mm]
 & ATE ($\beta_{1}$) & HET ($\beta_{2}$) & ATE ($\beta_{1}$) & HET ($\beta_{2}$) \\ 
 \cline{2-5}  \\ [-3mm]
& 2.814 & 1.047 & 2.441 & 0.899 \\ 
& (1.087,4.506) & (0.826,1.262) & (0.846,3.979) & (0.685,1.107) \\  \relax 
& [0.004] & [0.000] & [0.004] & [0.000] \\    [1mm]
   \hline\hline
\end{tabular}
\caption*{\footnotesize Notes: Medians over 250 splits, based upon prediction intervals presented in Section \ref{sec:MP}. Confidence Intervals for Median Parameter $(\alpha =.05, \beta =.225)$ in parenthesis.}
\end{table}


\begin{table}[H]
\centering
\caption{GATES of  20\% Most and Least Affected Groups}\label{table:Haryana_GATES_mpcausal2}
\footnotesize
\begin{tabular}{lcccccc}
\hline
\hline
\\ [-3mm]
   & \multicolumn{3}{c}{Elastic Net}  & \multicolumn{3}{c}{Nnet} \\   [1mm]
& 20\% Most  &  20\% Least  & Difference & 20\% Most  & 20\% Least  & Difference  \\  [-1mm]
 & ($G_5$) & ($G_1$)  &  & ($G_5$) & ($G_1$)  &    \\
 \cline{2-7}  \\ [-2mm]
GATE $\gamma_k := $  $ \hat{\Ep} [ s_0(Z) \mid G_k  ]$ & 13.230 & -8.000 & 21.60 & 11.210 & -6.551 & 18.13 \\ 
& (6.001,21.43) & (-16.38,-0.192) & (9.310,33.85) & (5.432,17.07) & (-12.85,-0.971) & (9.292,26.56) \\  
& [0.000] & [0.009] & [0.000] & [0.000] & [0.002] & [0.000] \\   [1mm]
\hline\hline
\end{tabular}
\caption*{\footnotesize Notes: Medians over 250 splits, based upon prediction intervals presented in Section \ref{sec:MP}. Confidence Intervals for Median Parameter $(\alpha =.05, \beta =.225)$ in parenthesis.}
\end{table}

\newpage

\begin{table}[H]
\centering
\caption{CLAN of Immunization Incentives }\label{table:Haryana_CLAN_C1_mpcausal2}
\fontsize{8}{8}\selectfont
\begin{tabular}{lcccccc}
  \hline
  \hline
  \\ [-1mm]
       & \multicolumn{3}{c}{Elastic Net}  & \multicolumn{3}{c}{Nnet} \\   [1mm]
& 20\% Most  & 20\% Least  & Difference & 20\% Most  & 20\% Least  & Difference\\ [0.5mm] 
 & ($\delta_{5}$) &  ($\delta_1$)  & ($\delta_{5}-\delta_{1}$) &   ($\delta_{5}$) &  ($\delta_1$)  & ($\delta_{5}-\delta_{1}$) \\ [1mm] 
 \cline{2-7}  \\ [-1mm]
  Number of vaccines  & 2.187 & 2.277 & -0.081 & 2.174 & 2.285 & -0.112 \\ 
     to pregnant mother & (2.115,2.259) & (2.212,2.342) & (-0.180,0.015) & (2.111,2.234) & (2.224,2.345) & (-0.202,-0.028) \\ 
   & - & - & [0.190] & - & - & [0.019] \\ 
  Number of vaccines  & 4.077 & 4.639 & -0.562 & 4.264 & 4.734 & -0.490 \\ 
   to child since birth   & (3.858,4.304) & (4.444,4.859) & (-0.863,-0.260) & (4.091,4.434) & (4.549,4.900) & (-0.739,-0.250) \\ 
   & - & - & [0.001] & - & - & [0.000] \\ 
  Fraction of children & 0.998 & 1.000 & -0.002 & 1.000 & 1.000 &  0.000 \\ 
    received polio drops  & (0.995,1.001) & (0.997,1.003) & (-0.006,0.002) & (1.000,1.000) & (1.000,1.000) & (0.000,0.000) \\ 
   & - & - & [0.683] & - & - & [0.943] \\ 
  Number of polio & 2.955 & 2.993 & -0.037 & 2.965 & 2.998 & -0.032 \\ 
     drops to child  & (2.935,2.974) & (2.976,3.010) & (-0.063,-0.010) & (2.953,2.977) & (2.985,3.010) & (-0.049,-0.016) \\ 
   & - & - & [0.013] & - & - & [0.000] \\ 
Fraction of children YN & 0.803 & 0.926 & -0.121 & 0.908 & 0.927 & -0.027 \\ 
    received immunization card & (0.754,0.851) & (0.882,0.969) & (-0.187,-0.054) & (0.881,0.932) & (0.898,0.959) & (-0.059,0.007) \\ 
   & - & - & [0.001] & - & - & [0.217] \\ 
  Fraction of children received  & 0.133 & 0.243 & -0.106 & 0.126 & 0.260 & -0.131 \\ 
     Measles vaccine by 15 months of age   & (0.097,0.169) & (0.209,0.276) & (-0.153,-0.056) & (0.095,0.159) & (0.228,0.291) & (-0.176,-0.085) \\ 
   & - & - & [0.000] & - & - & [0.000] \\ 
 Measles at credible locations  & 0.293 & 0.399 & -0.110 & 0.289 & 0.433 & -0.142 \\ 
   & (0.246,0.338) & (0.358,0.444) & (-0.174,-0.045) & (0.246,0.331) & (0.391,0.475) & (-0.206,-0.084) \\ 
   & - & - & [0.002] & - & - & [0.000] \\ 
   \hline \hline
\end{tabular}
\caption*{\footnotesize Notes: Medians over 250 splits, based upon prediction intervals presented in Section \ref{sec:MP}. Confidence Intervals for Median Parameter $(\alpha =.05, \beta =.225)$ in parenthesis.}
\end{table}